# 1 Deep Learning for Video Classification and Captioning

Zuxuan Wu (University of Maryland, College Park),
Ting Yao (Microsoft Research Asia),
Yanwei Fu (Fudan University),
Yu-Gang Jiang (Fudan University)

## 1.1 Introduction

Today's digital contents are inherently multimedia: text, audio, image, video, and so on. Video, in particular, has become a new way of communication between Internet users with the proliferation of sensor-rich mobile devices. Accelerated by the tremendous increase in Internet bandwidth and storage space, video data has been generated, published, and spread explosively, becoming an indispensable part of today's big data. This has encouraged the development of advanced techniques for a broad range of video understanding applications including online advertising, video retrieval, video surveillance, etc. A fundamental issue that underlies the success of these technological advances is the understanding of video contents. Recent advances in deep learning in image [Krizhevsky et al. 2012, Russakovsky et al. 2015, Girshick 2015, Long et al. 2015] and speech [Graves et al. 2013, Hinton et al. 2012] domains have motivated techniques to learn robust video feature representations to effectively exploit abundant multimodal clues in video data.

In this chapter, we review two lines of research aiming to stimulate the comprehension of videos with deep learning: video classification and video captioning. While video classification concentrates on automatically labeling video clips based on their semantic contents like human actions or complex events, video captioning attempts to generate a complete and natural sentence, enriching video classification's single label to capture the most informative dynamics in videos.

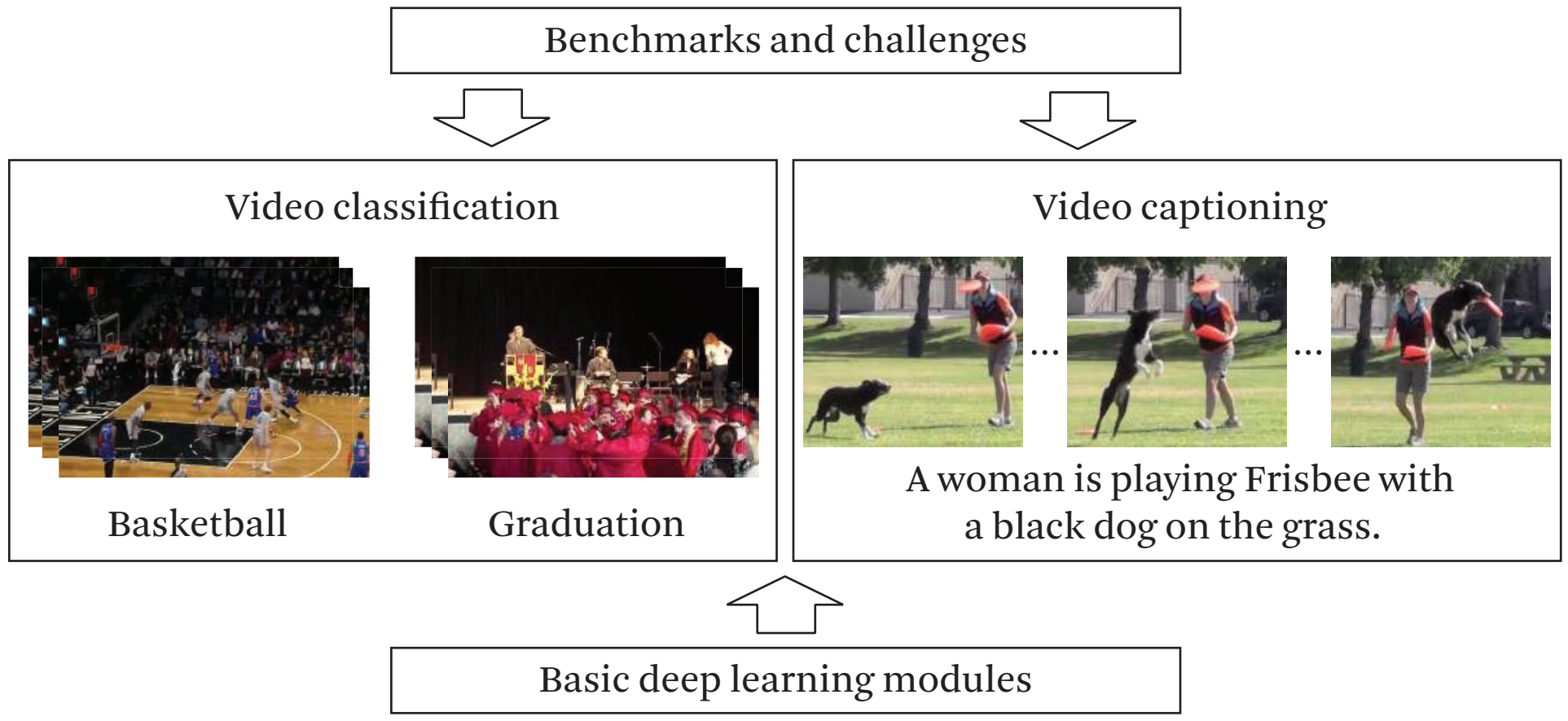

**Figure 1.1** An overview of the organization of this chapter.

There have been several efforts surveying the literature on video content understanding. Most of the approaches surveyed in these works adopted handcrafted features coupled with typical machine learning pipelines for action recognition and event detection [Aggarwal and Ryoo 2011, Turaga et al. 2008, Poppe 2010, Jiang et al. 2013]. In contrast, this chapter focuses on discussing state-of-the-art deep learning techniques not only for video classification but also video captioning. As deep learning for video analysis is an emerging and vibrant field, we hope this chapter could help stimulate future research along the line.

Figure 1.1 shows the organization of this chapter. To make it self-contained, we first introduce the basic modules that are widely adopted in state-of-the-art deep learning pipelines in Section 1.2. After that, we discuss representative works on video classification and video captioning in Section 1.3 and Section 1.4, respectively. Finally, in Section 1.5 we provide a review of popular benchmarks and challenges in that are critical for evaluating the technical progress of this vibrant field.

## 1.2 Basic Deep Learning Modules

In this section, we briefly review basic deep learning modules that have been widely adopted in the literature for video analysis.

### 1.2.1 Convolutional Neural Networks (CNNs)

Inspired by the visual perception mechanisms of animals [Hubel and Wiesel 1968] and the McCulloch-Pitts model [McCulloch and Pitts 1943], Fukushima proposed the “neocognitron” in 1980, which is the first computational model of using local connectivities between neurons of a hierarchically transformed image [Fukushima 1980]. To obtain the translational invariance, Fukushima applied neurons with the same parameters on patches of the previous layer at different locations; thus this can be considered the predecessor of convolutional neural networks (CNN. Further inspired by this idea, LeCun et al. [1990] designed and trained the modern framework of CNNs LeNet-5, and obtained the state-of-the-art performance on several pattern recognition datasets (e.g., handwritten character recognition). LeNet-5 has multiple layers and is trained with the back-propagation algorithm in an end-to-end formulation, that is, classifying visual patterns directly by using raw images. However, limited by the scale of labeled training data and computational power, LeNet-5 and its variants [LeCun et al. 2001] did not perform well on more complex vision tasks until recently.

To better train deep networks, Hinton et al. in 2006 made a breakthrough and introduced deep belief networks (DBNs) to greedily train each layer of the network in an unsupervised manner. And since then, researchers have developed more methods to overcome the difficulties in training CNN architectures. Particularly, AlexNet, as one of the milestones, was proposed by Krizhevsky et al. in 2012 and was successfully applied to large-scale image classification in the well-known ImageNet Challenge. AlexNet contains five convolutional layers followed by three fully connected (*fc*) layers [Krizhevsky et al. 2012]. Compared with LeNet-5, two novel components were introduced in AlexNet:

1. ReLUs (Rectified Linear Units) are utilized to replace the tanh units, which makes the training process several times faster.
2. Dropout is introduced and has proven to be very effective in alleviating overfitting.

Inspired by AlexNet, several variants, including VGGNet [Simonyan and Zisserman 2015], GoogLeNet [Szegedy et al. 2015a], and ResNet [He et al. 2016b], have been proposed to further improve the performance of CNNs on visual recognition tasks:

**VGGNet** has two versions, VGG16 and VGG19, which contain 16 and 19 layers, respectively [Simonyan and Zisserman 2015]. VGGNet pushed the depth of CNN architecture from 8 layers as in AlexNet to 16–19 layers, which greatly improves the discriminative power. In addition, by using very small (3 × 3) convolutional filters, VGGNet is capable of capturing details in the input images.

**GoogLeNet** is inspired by the Hebbian principle with multi-scale processing and it contains 22 layers [Szegedy et al. 2015a]. A novel CNN architecture commonly referred to as Inception is proposed to increase both the depth and the width of CNN while maintaining an affordable computational cost. There are several extensions upon this work, including BN-Inception-V2 [Szegedy et al. 2015b], Inception-V3 [Szegedy et al. 2015b], and Inception-V4 [Szegedy et al. 2017].

**ResNet**, as one of the latest deep architectures, has remarkably increased the depth of CNN to 152 layers using deep residual layers with skip connections [He et al. 2016b]. ResNet won the first place in the 2015 ImageNet Challenge and has recently been extended to more than 1000 layers on the CIFAR-10 dataset [He et al. 2016a].

From AlexNet, VGGNet, and GoogLeNet to the more recent ResNet, one trend in the evolution of these architectures is to deepen the network. The increased depth allows the network to better approximate the target function, generating better feature representations with higher discriminative power. In addition, various methods and strategies have been proposed from different aspects, including but not limited to Maxout [Goodfellow et al. 2013], DropConnect [Wan et al. 2013], and Batch Normalization [Ioffe and Szegedy 2015], to facilitate the training of deep networks. Please refer to Bengio et al. [2013] and Gu et al. [2016] for a more detailed review.

### 1.2.2 Recurrent Neural Networks (RNNs)

The CNN architectures discussed above are all feed-forward neural networks (FFNNs) whose connections do not form cycles, which makes them insufficient for sequence labeling. To better explore the temporal information of sequential data, recurrent connection structures have been introduced, leading to the emergence of recurrent neural networks (RNNs). Unlike FFNNs, RNNs allow cyclical connections to form cycles, which thus enables a "memory" of previous inputs to persist in the network's internal state [Graves 2012]. It has been pointed out that a finite-sized RNN with sigmoid activation functions can simulate a universal Turing machine [Siegelmann and Sontag 1991].

**The basic RNN block**, at a time step $t$, accepts an external input vector $\mathbf{x}^{(t)} \in \mathbb{R}^n$ and generates an output vector $\mathbf{z}^{(t)} \in \mathbb{R}^m$ via a sequence of hidden states

$\mathbf{h}^{(t)} \in \mathbb{R}^{r}$:

$$\mathbf{h}^{(t)} = \sigma\left(W_x \mathbf{x}^{(t)} + W_h \mathbf{h}^{(t-1)} + \mathbf{b}_h\right) \tag{1.1}$$
$$\mathbf{z}^{(t)} = \text{softmax}\left(W_z \mathbf{h}^{(t)} + \mathbf{b}_z\right)$$

where $W_x \in \mathbb{R}^{r\times n}$, $W_h \in \mathbb{R}^{r\times r}$, and $W_z \in \mathbb{R}^{m\times r}$ are weight matrices and $\mathbf{b}_h$ and $\mathbf{b}_z$ are biases. The $\sigma$ is defined as sigmoid function $\sigma(x) = \frac{1}{1+e^{-x}}$ and softmax $(\cdot)$ is the softmax function.

A problem with RNN is that it is not capable of modeling long-range dependencies and is unable to store information about past inputs for a very long period [Bengio et al. 1994], though one large enough RNN should, in principle, be able to approximate the sequences of arbitrary complexity. Specifically, two well-known issues—*vanishing* and *exploding* gradients, exist in training RNNs: the vanishing gradient problem refers to the exponential shrinking of gradients' magnitude as they are propagated back through time; and the exploding gradient problem refers to the explosion of long-term components due to the large increase in the norm of the gradient during training sequences with long-term dependencies. To solve these issues, researchers introduced Long short-term memory models.

**Long short-term memory (LSTM)** is an RNN variant that was designed to store and access information in a long time sequence. Unlike standard RNNs, non-linear multiplicative gates and a memory cell are introduced. These gates, including input, output, and forget gates, govern the information flow into and out of the memory cell. The structure of an LSTM unit is illustrated in Figure 1.2.

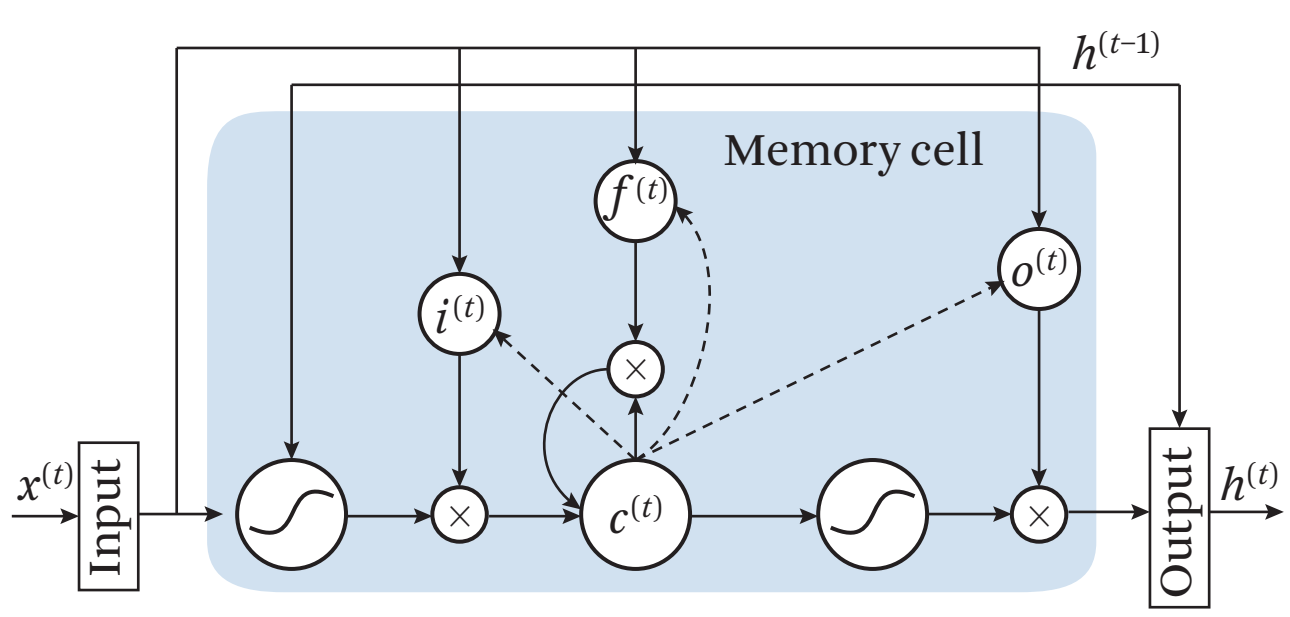

**Figure 1.2** The structure of an LSTM unit. (Modified from Wu et al. [2016])

More specifically, given a sequence of an external input vector $\mathbf{x}^{(t)} \in \mathbb{R}^n$, an LSTM maps the input to an output vector $\mathbf{z}^{(t)} \in \mathbb{R}^m$ by computing activations of the units in the network with the following equations recursively from $t = 1$ to $t = T$:

$$
\begin{aligned}
\mathbf{i}^{(t)} &= \sigma(W_{xi}\mathbf{x}^{(t)} + W_{hi}\mathbf{h}^{(t-1)} + W_{ci}\mathbf{c}^{(t)} + \mathbf{b}_i), \\
\mathbf{f}^{(t)} &= \sigma(W_{xf}\mathbf{x}^{(t)} + W_{hf}\mathbf{h}^{(t)} + W_{cf}\mathbf{c}^{(t)} + \mathbf{b}_f), \\
\mathbf{c}^{(t)} &= \mathbf{f}^{(t)}\mathbf{c}^{(t-1)} + \mathbf{i}_t \tanh(W_{xc}\mathbf{x}^{(t)} + W_{hc}\mathbf{h}^{(t-1)} + \mathbf{b}_c), \\
\mathbf{o}^{(t)} &= \sigma(W_{xo}\mathbf{x}^{(t)} + W_{ho}\mathbf{h}^{(t-1)} + W_{co}\mathbf{c}^{(t)} + \mathbf{b}_o), \\
\mathbf{h}^{(t)} &= \mathbf{o}^{(t)} \tanh(\mathbf{c}^{(t)}),
\end{aligned}
\tag{1.2}
$$

where $\mathbf{x}^{(t)}$, $\mathbf{h}^{(t)}$ are the input and hidden vectors with the subscription $t$ denoting the $t$-th time step, while $\mathbf{i}^{(t)}$, $\mathbf{f}^{(t)}$, $\mathbf{c}^{(t)}$, $\mathbf{o}^{(t)}$ are, respectively, the activation vectors of the input gate, forget gate, memory cell, and output gate. $W_{\alpha\beta}$ denotes the weight matrix between $\alpha$ and $\beta$. For example, the weight matrix from the input $\mathbf{x}^{(t)}$ to the input gate $\mathbf{i}^{(t)}$ is $W_{xi}$.

In Equation 1.2 and Figure 1.2 and at time step $t$, the input $\mathbf{x}^{(t)}$ and the previous states $\mathbf{h}^{(t-1)}$ are used as the input of LSTM. The information of the memory cell is updated/controlled from two sources: (1) the previous cell memory unit $\mathbf{c}^{(t-1)}$ and (2) the input gate's activation $\mathbf{i}_t$. Specifically, $\mathbf{c}^{(t-1)}$ is multiplied by the activation from the forget gate $\mathbf{f}^{(t)}$, which learns to forget the information of the previous states. In contrast, the $\mathbf{i}_t$ is combined with the new input signal to consider new information. LSTM also utilizes the output gate $\mathbf{o}^{(t)}$ to control the information received by hidden state variable $\mathbf{h}^{(t)}$. To sum up, with these explicitly designed memory units and gates, LSTM is able to exploit the long-range temporal memory and avoids the issues of vanishing/exploding gradients. LSTM has recently been popularly used for video analysis, as will be discussed in the following sections.

## 1.3 Video Classification

The sheer volume of video data has motivated approaches to automatically categorizing video contents according to classes such as human activities and complex events. There is a large body of literature focusing on computing effective local feature descriptors (e.g., HoG, HoF, MBH, etc.) from spatio-temporal volumes to account for temporal clues in videos. These features are then quantized into bag-of-words or Fisher Vector representations, which are further fed into classifiers like support vector machines (SVMs). In contrast to hand crafting features, which is usually time-consuming and requires domain knowledge, there is a recent trend to learn robust feature representations with deep learning from raw video data. In the following, we review two categories of deep learning algorithms for video classification, i.e., supervised deep learning and unsupervised feature learning.

## 1.3.1 Supervised Deep Learning for Classification

### 1.3.1.1 Image-Based Video Classification

The great success of CNN features on image analysis tasks [Girshick et al. 2014, Razavian et al. 2014] has stimulated the utilization of deep features for video classification. The general idea is to treat a video clip as a collection of frames, and then for each frame, feature representation could be derived by running a feed-forward pass till a certain fully-connected layer with state-of-the-art deep models pre-trained on ImageNet [Deng et al. 2009], including AlexNet [Krizhevsky et al. 2012], VGGNet [Simonyan and Zisserman 2015], GoogLeNet [Szegedy et al. 2015a], and ResNet [He et al. 2016b], as discussed earlier. Finally, frame-level features are averaged into video-level representations as inputs of standard classifiers for recognition, such as the well-known SVMs.

Among the works on image-based video classification, Zha et al. [2015] systematically studied the performance of image-based video recognition using features from different layers of deep models together with multiple kernels for classification. They demonstrated that off-the-shelf CNN features coupled with kernel SVMs can obtain decent recognition performance. Motivated by the advanced feature encoding strategies in images [Sánchez et al. 2013], Xu et al. [2015c] proposed to obtain video-level representation through vector of locally aggregated descriptors (VLAD) encoding [Jégou et al. 2010b], which can attain performance gain over the trivial averaging pooling approach. Most recently, Qiu et al. [2016] devised a novel Fisher Vector encoding with Variational AutoEncoder (FV-VAE) to quantize the local activations of the convolutional layer, which learns powerful visual representations of better generalization.

### 1.3.1.2 End-to-End CNN Architectures

The effectiveness of CNNs on a variety of tasks lies in their capability to learn features from raw data as an end-to-end pipeline targeting a particular task [Szegedy et al. 2015a, Long et al. 2015, Girshick 2015]. Therefore, in contrast to the image-based classification methods, there are many works focusing on applying CNN models to the video domain with an aim to learn hidden spatio-temporal patterns. Ji et al. [2010] introduced the 3D CNN model that operates on stacked video frames, extending the traditional 2D CNN designed for images to the spatio-temporal space. The 3D CNN utilizes 3D kernels for convolution to learn motion information between adjacent frames in volumes segmented by human detectors. Karpathy et al. [2014] compared several similar architectures on a large scale video dataset in order to explore how to better extend the original CNN architectures to learn spatio-temporal clues in videos. They found that the performance of the CNN model with a single frame as input achieves similar results to models operating on a stack of frames, and they also suggested that a mixed-resolution architecture consisting of a low-resolution context and a high-resolution stream could speed up training effectively. Recently, Tran et al. [2015] also utilized 3D convolutions with modern deep architectures. However, they adopted full frames as the inputs of 3D CNNs instead of the segmented volumes in Ji et al. [2010].

Though the extension of conventional CNN models by stacking frames makes sense, the performance of such models is worse than that of state-of-the-art hand-crafted features [Wang and Schmid 2013]. This may be because the spatio-temporal patterns in videos are too complex to be captured by deep models with insufficient training data. In addition, the training of CNNs with inputs of 3D volumes is usually time-consuming. To effectively handle 3D signals, Sun et al. [2015] introduced factorized spatio-temporal convolutional networks that factorize the original 3D convolution kernel learning as a sequential process of learning 2D spatial kernels in the lower layer. In addition, motivated by the fact that videos can naturally be decomposed into spatial and temporal components, Simonyan and Zisserman [2014] proposed a two-stream approach (see Figure 1.3), which breaks down the learning of video representation into separate feature learning of spatial and temporal clues. More specifically, the authors first adopted a typical spatial CNN to model appearance information with raw RGB frames as inputs. To account for temporal clues among adjacent frames, they explicitly generated multiple-frame dense optical flow, upon which a temporal CNN is trained. The dense optical flow is derived from computing displacement vector fields between adjacent frames (see Figure 1.4), which represent motions in an explicit way, making the training of the network easier. Finally, at test time, each individual CNN generates a prediction by averaging scores from 25 uniformly sampled frames (optical flow frames) for a video clip, and then the final output is produced by the weighted sum of scores from the two streams. The authors reported promising results on two action recognition benchmarks. As the two-stream approach contains many implementation choices that may affect the performance, Ye et al. [2015b] evaluated different options, including dropout ratio and network architecture, and discussed their findings.

Very recently, there have been several extensions of the two-stream approach. Wang et al. utilized the point trajectories from the improved dense trajectories [Wang and Schmid 2013] to pool two-stream convolutional feature maps to generate trajectory-pooled deep-convolutional descriptors (TDD) [Wang et al. 2015]. Feichtenhofer et al. [2016] improved the two-stream approach by exploring a better fusion approach to combine spatial and temporal streams. They found that two streams could be fused using convolutional layers rather than averaging classification scores to better model the correlations of spatial and temporal streams. Wang et al. [2016b] introduced temporal segment networks, where each segment is used as the input of a two-stream network and the final prediction of a video clip is produced by a consensus function combining segment scores. Zhang et al. [2016] proposed to replace the optical flow images with motion vectors with an aim to achieve real-time action recognition. More recently, Wang et al. [2016c] proposed to learn feature representation by modeling an action as a transformation from an initial state (condition) to a new state (effect) with two Siamese CNN networks, operating on RGB frames and optical flow images. Similar to the original two-stream approach, they then fused the classification scores from two streams linearly to obtain final predictions. They reported better results on two challenging benchmarks than Simonyan and Zisserman [2014], possibly because the transformation from precondition to effect could implicitly model the temporal coherence in videos. Zhu et al. [2016] proposed a key volume mining approach that attempts to identify key volumes and perform classification at the same time. Bilen et al. [2016] introduced the dynamic image to represent motions with rank pooling in videos, upon which a CNN model is trained for recognition.

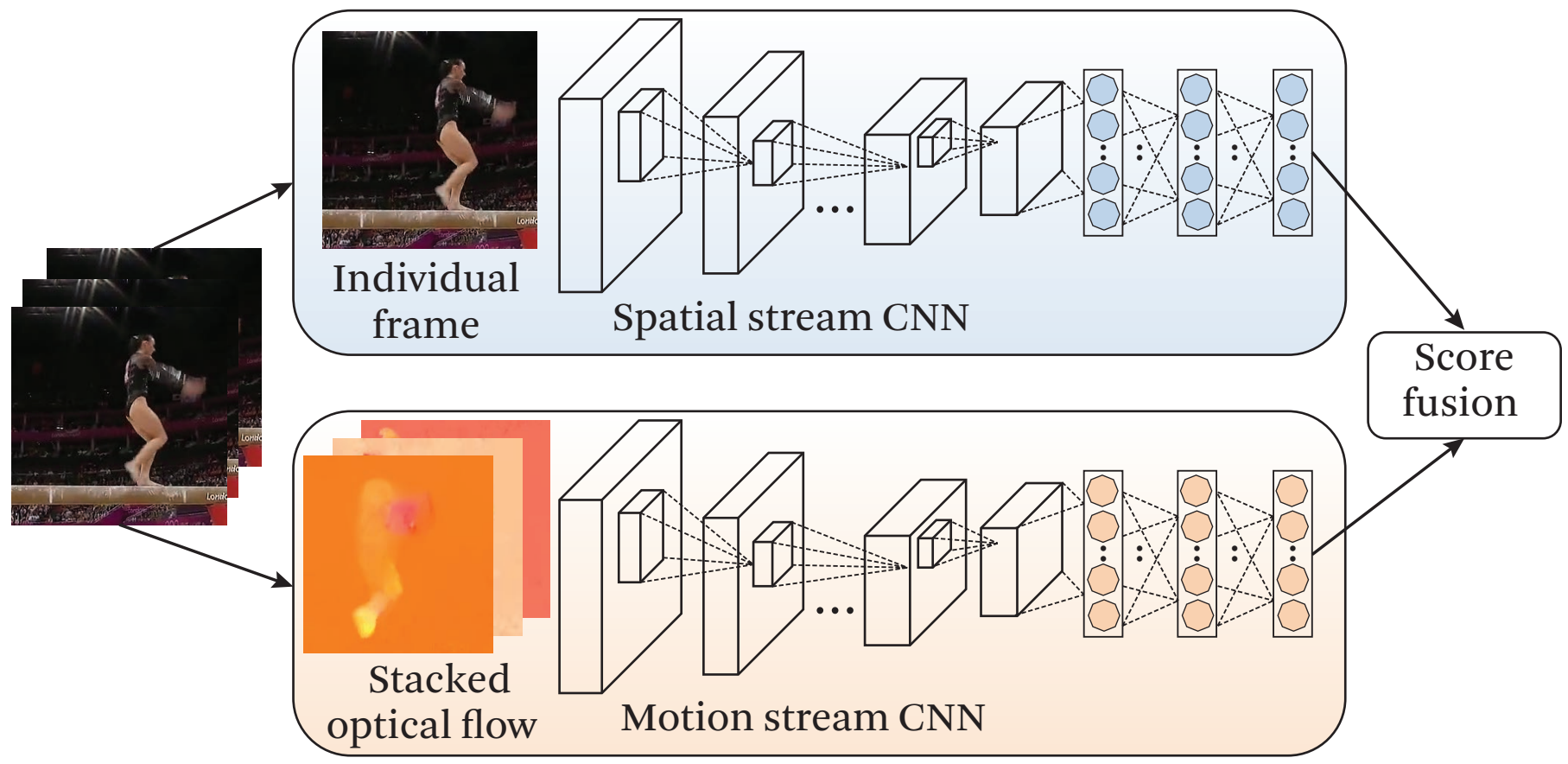

**Figure 1.3** Two-stream CNN framework. (From Wu et al. [2015c])

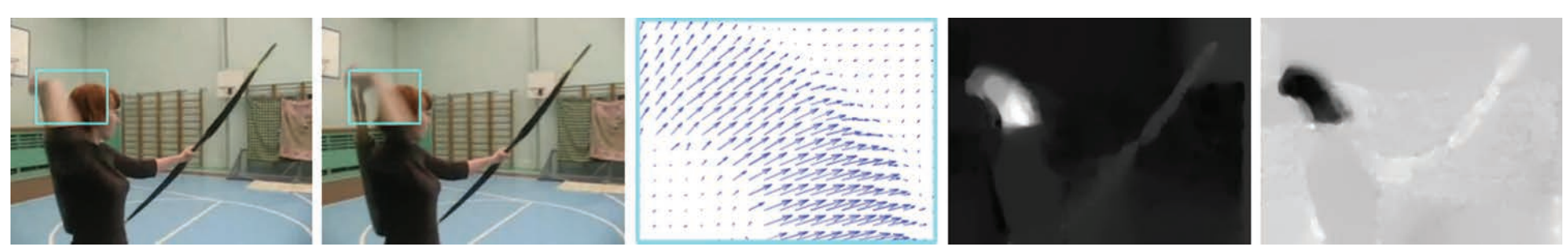

**Figure 1.4** Examples of optical flow images. (From Simonyan and Zisserman [2014])

#### 1.3.1.3 Modeling Long-Term Temporal Dynamics

As discussed earlier, the temporal CNN in the two-stream approach [Simonyan and Zisserman 2014] explicitly captures the motion information among adjacent frames, which, however, only depicts movements within a short time window. In addition, during the training of CNN models, each sweep takes a single frame (or a stacked optical frame image) as the input of the network, failing to take the order of frames into account. This is not sufficient for video analysis, since complicated events/actions in videos usually consist of multiple actions happening over a long time. For instance, a "making pizza" event can be decomposed into several sequential actions, including "making the dough," "topping," and "baking." Therefore, researchers have recently attempted to leverage RNN models to account for the temporal dynamics in videos, among which LSTM is a good fit without suffering from the "vanishing gradient" effect, and has demonstrated its effectiveness in several tasks like image/video captioning [Donahue et al. 2017, Yao et al. 2015a] (to be discussed in detail later) and speech analysis [Graves et al. 2013].

Donahue et al. [2017] trained two two-layer LSTM networks (Figure 1.5) for action recognition with features from the two-stream approach. They also tried to fine-tune the CNN models together with LSTM but did not obtain significant performance gain compared with only training the LSTM model. Wu et al. [2015c] fused the outputs of LSTM models with CNN models to jointly model spatio-temporal clues for video classification and observed that CNNs and LSTMs are highly complementary. Ng et al. [2015] further trained a 5-layer LSTM model and compared several pooling strategies. Interestingly, the deep LSTM model performs on par with single frame CNN on a large YouTube video dataset called Sports-1M, which may be because the videos in this dataset are uploaded by ordinary users without professional editing and contain cluttered backgrounds and severe camera motion. Veeriah et al. [2015] introduced a differential gating scheme for LSTM to emphasize the change in information gain to remove redundancy in videos. Recently, in a multi-granular spatio-temporal architecture [Li et al. 2016a], LSTMs have been utilized to further model the temporal information of frame, motion, and clip streams. Wu et al. [2016] further employed a CNN operating on spectrograms derived from soundtracks of videos to complement visual clues captured by CNN and LSTMs, and demonstrated strong results.

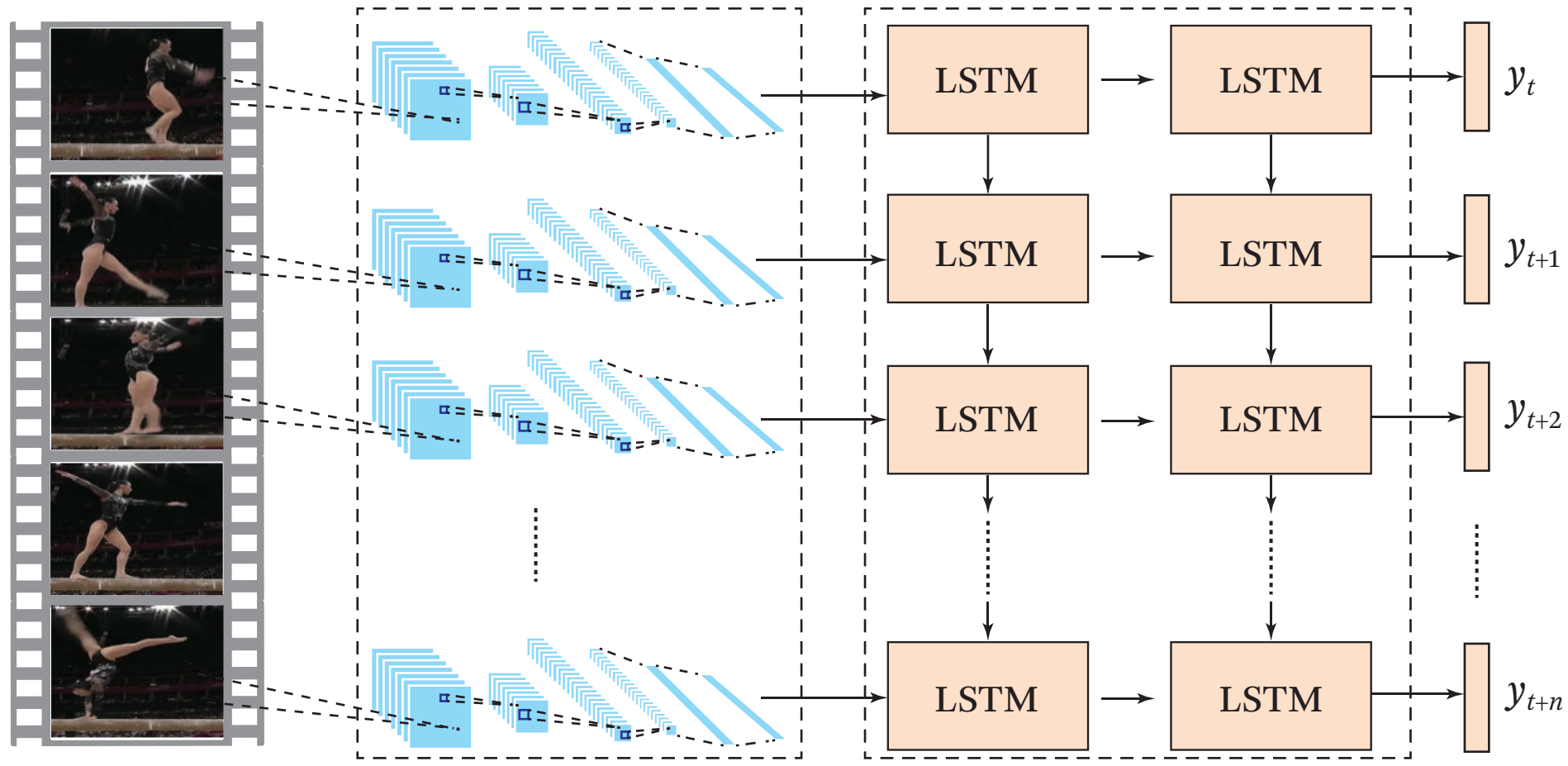

**Figure 1.5** Utilizing LSTMs to explore temporal dynamics in videos with CNN features as inputs.

#### 1.3.1.4 Incorporating Visual Attention

Videos contain many frames. Using all of them is computationally expensive and may degrade the performance of recognizing a class of interest as not all the frames are relevant. This issue has motivated researchers to leverage the attention mechanism to identify the most discriminative spatio-temporal volumes that are directly related to the targeted semantic class. Sharma et al. [2015] proposed the first attention LSTM for action recognition with a soft-attention mechanism to attach higher importance to the learned relevant parts in video frames. More recently, Li et al. [2016c] introduced the VideoLSTM, which applied attention in convolutional LSTM models to discover relevant spatio-temporal volumes. In addition to soft-attention, VideoLSTM also employed motion-based attention derived from optical flow images for better action localization.

### 1.3.2 Unsupervised Video Feature Learning

Current remarkable improvements with deep learning heavily rely on a large amount of labeled data. However, scaling up to thousands of video categories presents significant challenges due to insurmountable annotation efforts even at video level, not to mention frame-level fine-grained labels. Therefore, the utilization of unsupervised learning, integrating spatial and temporal context information, is a promising way to find and represent structures in videos. Taylor et al. [2010]

proposed a convolutional gated boltzmann Machine to learn to represent optical flow and describe motion. Le et al. [2011] utilized two-layer independent subspace analysis (ISA) models to learn spatio-temporal models for action recognition. More recently, Srivastava et al. [2015] adopted an encoder-decoder LSTM to learn feature representations in an unsupervised way. They first mapped an input sequence into a fixed-length representation by an encoder LSTM, which would be further decoded with single or multiple decoder LSTMs to perform different tasks, such as reconstructing the input sequence, or predicting the future sequence. The model was first pre-trained on YouTube data without manual labels, and then fine-tuned on standard benchmarks to recognize actions. Pan et al. [2016a] explored both local temporal coherence and holistic graph structure preservation to learn a deep intrinsic video representation in an end-to-end fashion. Ballas et al. [2016] leveraged convolutional maps from different layers of a pre-trained CNN as the input of a gated recurrent unit (GRU)-RNN to learn video representations.

### Summary

The latest developments discussed above have demonstrated the effectiveness of deep learning for video classification. However, current deep learning approaches for video classification usually resort to popular deep models in image and speech domain. The complicated nature of video data, containing abundant spatial, temporal, and acoustic clues, makes off-the-shelf deep models insufficient for video-related tasks. This highlights the need for a tailored network to effectively capture spatial and acoustic information, and most importantly to model temporal dynamics. In addition, training CNN/LSTM models requires manual labels that are usually expensive and time-consuming to acquire, and hence one promising direction is to make full utilization of the substantial amounts of unlabeled video data and rich contextual clues to derive better video representations.

## 1.4 Video Captioning

Video captioning is a new problem that has received increasing attention from both computer vision and natural language processing communities. Given an input video, the goal is to automatically generate a complete and natural sentence, which could have a great potential impact, for instance, on robotic vision or on helping visually impaired people. Nevertheless, this task is very challenging, as a description generation model should capture not only the objects, scenes, and activities presented in the video, but also be capable of expressing how these objects/scenes/activities relate to each other in a natural sentence. In this section, we elaborate the problem by surveying the state-of-the-art methods. We classify exist-

**Input video:**

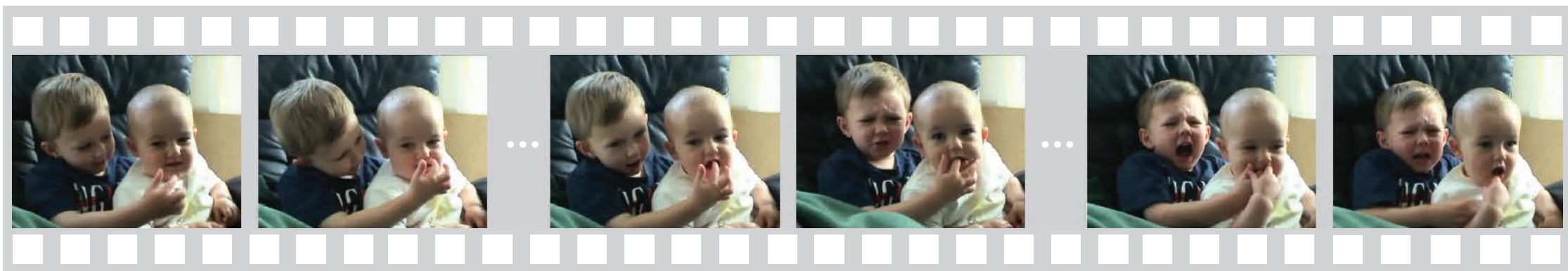

**Video tagging:**
baby, boy, chair

**Video (frame) captioning:**
Two baby boys are in the chair.

**Video captioning:**
A baby boy is biting finger of another baby boy.

**Figure 1.6** Examples of video tagging, image (frame) captioning, and video captioning. The input is a short video, while the output is a text response to this video, in the form of individual words (tags), a natural sentence describing one single image (frame), and dynamic video contents, respectively.

ing methods in terms of different strategies for sentence modeling. In particular, we distill a common architecture of combining convolutional and recurrent neural networks for video captioning. As video captioning is an emerging area, we start by introducing the problem in detail.

### 1.4.1 Problem Introduction

Although there has already been extensive research on video tagging [Siersdorfer et al. 2009, Yao et al. 2013] and image captioning [Vinyals et al. 2015, Donahue et al. 2017], video-level captioning has its own characteristics and thus is different from tagging and image/frame-level captioning. A video tag is usually the name of a specific object, action, or event, which is recognized in the video (e.g., "baby," "boy," and "chair" in Figure 1.6). Image (frame) captioning goes beyond tagging by describing an image (frame) with a natural sentence, where the spatial relationships between objects or object and action are further described (e.g., "Two baby boys are in the chair" generated on one single frame of Figure 1.6). Video captioning has been taken as an even more challenging problem, as a description should not only capture the above-mentioned semantic knowledge in the video but also express the spatio-temporal relationships in between and the dynamics in a natural sentence (e.g., "A baby boy is biting finger of another baby boy" for the video in Figure 1.6).

Despite the difficulty of the problem, there have been several attempts to address video caption generation [Pan et al. 2016b, Yu et al. 2016, Xu et al. 2016], which are mainly inspired by recent advances in machine translation [Sutskever et al. 2014]. The elegant recipes behind this are the promising developments of the CNNs and the RNNs. In general, 2D [Simonyan and Zisserman 2015] and/or 3D CNNs [Tran et al. 2015] are exploited to extract deep visual representations and LSTM [Hochreiter and Schmidhuber 1997] is utilized to generate the sentence word by word. More sophisticated frameworks, additionally integrating internal or external knowledge in the form of high-level semantic attributes or further exploring the relationship between the semantics of sentence and video content, have also been studied for this problem.

In the following subsections we present a comprehensive review of video captioning methods through two main categories based on the strategies for sentence generation (Section 1.4.2) and generalizing a common architecture by leveraging sequence learning for video captioning (Section 1.4.3).

## 1.4.2 Approaches for Video Captioning

There are mainly two directions for video captioning: a template-based language model [Kojima et al. 2002, Rohrbach et al. 2013, Rohrbach et al. 2014, Guadarrama et al. 2013, Xu et al. 2015b] and sequence learning models (e.g., RNNs) [Donahue et al. 2017, Pan et al. 2016b, Xu et al. 2016, Yu et al. 2016, Venugopalan et al. 2015a, Yao et al. 2015a, Venugopalan et al. 2015b, Venugopalan et al. 2015b]. The former predefines the special rule for language grammar and splits the sentence into several parts (e.g., subject, verb, object). With such sentence fragments, many works align each part with detected words from visual content by object recognition and then generate a sentence with language constraints. The latter leverages sequence learning models to directly learn a translatable mapping between video content and sentence. We will review the state-of-the-art research along these two dimensions.

### 1.4.2.1 Template-based Language Model

Most of the approaches in this direction depend greatly on the sentence templates and always generate sentences with syntactical structure. Kojima et al. [2002] is one of the early works that built a concept hierarchy of actions for natural language description of human activities. Tan et al. [2011] proposed using predefined concepts and sentence templates for video event recounting. Rohrbach et al.'s conditional random field (CRF) learned to model the relationships between different components of the input video and generate descriptions for videos [Rohrbach et al. 2013]. Furthermore, by incorporating semantic unaries and hand-centric features, Rohrbach et al. [2014] utilized a CRF-based approach to generate coherent video descriptions. In 2013, Guadarrama et al. used semantic hierarchies to choose an appropriate level of the specificity and accuracy of sentence fragments. Recently, a deep joint video-language embedding model in Xu et al. [2015b] was designed for video sentence generation.

#### 1.4.2.2 Sequence Learning

Unlike the template-based language model, sequence learning-based methods can learn the probability distribution in the common space of visual content and textual sentence and generate novel sentences with more flexible syntactical structure. Donahue et al. [2017] employed a CRF to predict activity, object, and location present in the video input. These representations were concatenated into an input sequence and then translated to a natural sentence with an LSTM model. Later, Venugopalan et al. [2015b] proposed an end-to-end neural network to generate video descriptions by reading only the sequence of video frames. By mean pooling, the features over all the frames can be represented by one single vector, which is the input of the following LSTM model for sentence generation. Venugopalan et al. [2015a] then extended the framework by inputting both frames and optical flow images into an encoder-decoder LSTM. Inspired by the idea of learning visual-semantic embedding space in search [Pan et al. 2014, Yao et al. 2015b], [Pan et al. 2016b] additionally considered the relevance between sentence semantics and video content as a regularizer in LSTM based architecture. In contrast to mean pooling, Yao et al. [2015a] proposed to utilize the temporal attention mechanism to exploit temporal structure as well as a spatio-temporal convolutional neural network to obtain local action features. Then, the resulting video representations were fed into the text-generating RNN. In addition, similar to the knowledge transfer from image domain to video domain [Yao et al. 2012, 2015c], Liu and Shi [2016] leveraged the learned models on image captioning to generate a caption for each video frame and incorporate the obtained captions, regarded as the attributes of each frame, into a sequence-to-sequence architecture to generate video descriptions. Most recently, with the encouraging performance boost reported on the image captioning task by additionally utilizing high-level image attributes in Yao et al. [2016], Pan et al. [2016c] further leveraged semantic attributes learned from both images and videos with a transfer unit for enhancing video sentence generation.

### 1.4.3 A Common Architecture for Video Captioning

To better summarize the frameworks of video captioning by sequence learning, we illustrate a common architecture as shown in Figure 1.7. Given a video, 2D and/or 3D CNNs are utilized to extract visual features on raw video frames, optical flow images, and video clips. The video-level representations are produced by mean pooling or soft attention over these visual features. Then, an LSTM is trained for generating a sentence based on the video-level representations.

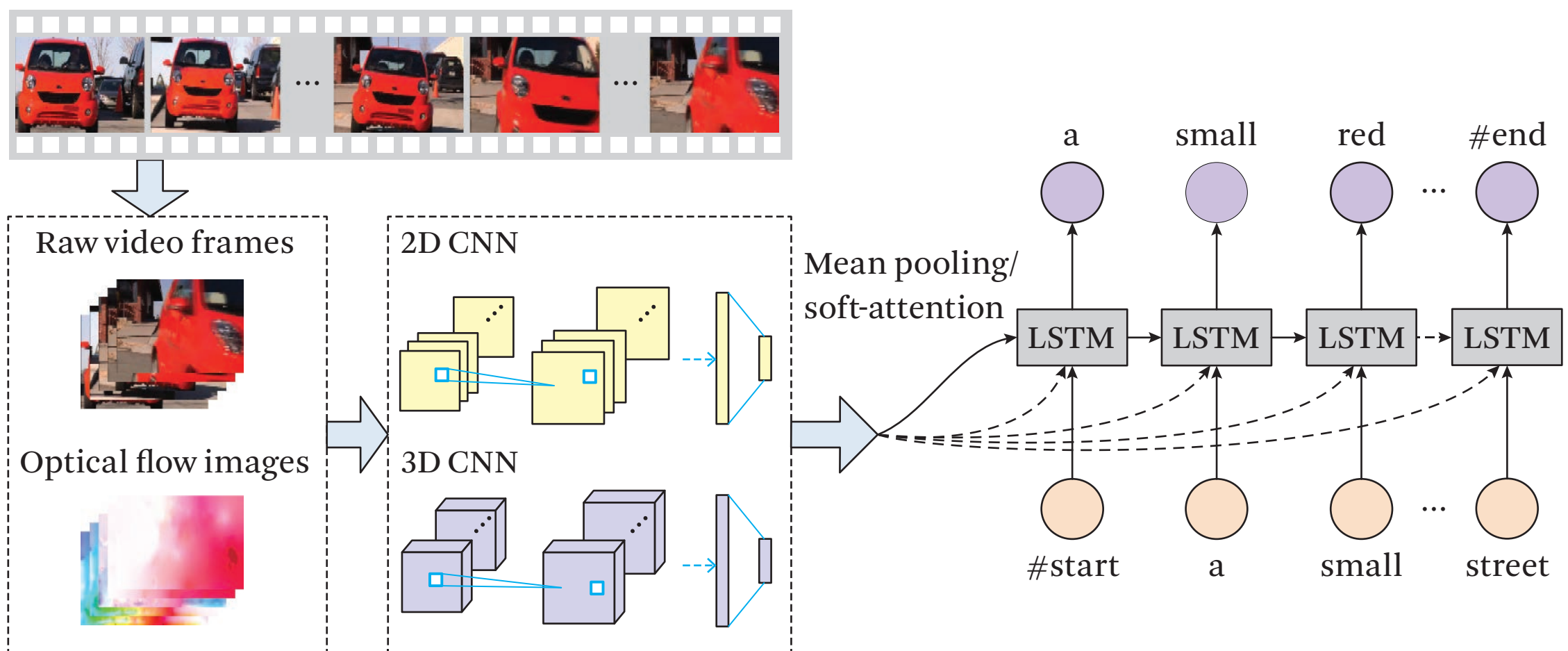

**Figure 1.7** A common architecture for video captioning by sequence learning. The video representations are produced by mean pooling or soft-attention over the visual features of raw video frames/optical flow images/video clips, extracted by 2D/3D CNNs. The sentence is generated word by word in the following LSTM, based on the video representations.

Technically, suppose we have a video $\mathcal{V}$ with $N_v$ sample frames/optical images/clips (uniform sampling) to be described by a textual sentence $\mathcal{S}$, where $\mathcal{S} = \{w_1, w_2, \ldots, w_{N_s}\}$ consisting of $N_s$ words. Let $\mathbf{v} \in \mathbb{R}^{D_v}$ and $\mathbf{w}_t \in \mathbb{R}^{D_w}$ denote the $D_v$-dimensional visual features of a video $\mathcal{V}$ and the $D_w$-dimensional textual features of the $t$-th word in sentence $\mathcal{S}$, respectively. As a sentence consists of a sequence of words, a sentence can be represented by a $D_w \times N_s$ matrix $\mathbf{W} \equiv [\mathbf{w}_1, \mathbf{w}_2, \ldots, \mathbf{w}_{N_s}]$, with each word in the sentence as its column vector. Hence, given the video representations $\mathbf{v}$, we aim to estimate the conditional probability of the output word sequence $\{w_1, w_2, \ldots, w_{N_s}\}$, i.e.,

$$\Pr(\mathbf{w}_1, \mathbf{w}_2, \ldots, \mathbf{w}_{N_s} | \mathbf{v}). \tag{1.3}$$

Since the model produces one word in the sentence at each time step, it is natural to apply the chain rule to model the joint probability over the sequential words. Thus, the log probability of the sentence is given by the sum of the log probabilities over the words and can be expressed as:

$$\log \Pr(\mathbf{W}|\mathbf{v}) = \sum_{t=1}^{N_s} \log \Pr\left(\mathbf{w}_t \middle| \mathbf{v}, \mathbf{w}_1, \ldots, \mathbf{w}_{t-1}\right). \tag{1.4}$$

In the model training, we feed the start sign word *#start* into LSTM, which indicates the start of the sentence generation process. We aim to maximize the log probability of the output video description $\mathcal{S}$ given the video representations, the previous words it has seen, and the model parameters $\theta$, which can be formulated as

$$\theta^* = \arg\max_{\theta} \sum_{t=1}^{N_s} \log \Pr \left( \mathbf{w}_t \middle| \mathbf{v}, \mathbf{w}_1, \ldots, \mathbf{w}_{t-1}; \theta \right). \tag{1.5}$$

This log probability is calculated and optimized over the whole training dataset using stochastic gradient descent. Note that the end sign word *#end* is required to terminate the description generation. During inference, we choose the word with maximum probability at each time step and set it as the LSTM input for the next time step until the end sign word is emitted.

### Summary

The introduction of the video captioning problem is relatively new. Recently, this task has sparked significant interest and may be regarded as the ultimate goal of video understanding. Video captioning is a complex problem and has been initially forwarded by the fundamental technological advances in recognition that can effectively recognize key objects or scenes from video contents. The developments of RNNs in machine translation have further accelerated the growth of this research direction. The recent results, although encouraging, are still indisputably far from practical use, as the forms of the generated sentences are simple and the vocabulary is still limited. How to generate free-form sentences and support open vocabulary are vital issues for the future of this task.

## 1.5 Benchmarks and Challenges

We now discuss popular benchmarks and challenges for video classification (Section 1.5.1) and video captioning (Section 1.5.2).

### 1.5.1 Classification

Research on video classification has been stimulated largely by the release of the large and challenging video datasets such as UCF101 [Soomro et al. 2012], HMDB51 [Kuehne et al. 2011], and FCVID [Jiang et al. 2015], and by the open challenges organized by fellow researchers, including the THUMOS challenge [Jiang et al. 2014b], the ActivityNet Large Scale Activity Recognition Challenge [Heilbron et al. 2015], and the TRECVID multimedia event detection (MED) task [Over et al. 2014].

**Table 1.1** **Popular benchmark datasets for video classification, sorted by the year of construction**

| Dataset | #Video | #Class | Released Year | Background |
|---|---|---|---|---|
| KTH | 600 | 6 | 2004 | Clean Static |
| Weizmann | 81 | 9 | 2005 | Clean Static |
| Kodak | 1,358 | 25 | 2007 | Dynamic |
| Hollywood | 430 | 8 | 2008 | Dynamic |
| Hollywood2 | 1,787 | 12 | 2009 | Dynamic |
| MCG-WEBV | 234,414 | 15 | 2009 | Dynamic |
| Olympic Sports | 800 | 16 | 2010 | Dynamic |
| HMDB51 | 6,766 | 51 | 2011 | Dynamic |
| CCV | 9,317 | 20 | 2011 | Dynamic |
| UCF-101 | 13,320 | 101 | 2012 | Dynamic |
| THUMOS-2014 | 18,394 | 101 | 2014 | Dynamic |
| MED-2014 (Dev. set) | ≈31,000 | 20 | 2014 | Dynamic |
| Sports-1M | 1,133,158 | 487 | 2014 | Dynamic |
| ActivityNet | 27,901 | 203 | 2015 | Dynamic |
| EventNet | 95,321 | 500 | 2015 | Dynamic |
| MPII Human Pose | 20,943 | 410 | 2014 | Dynamic |
| FCVID | 91,223 | 239 | 2015 | Dynamic |

In the following, we first discuss related datasets according to the list shown in Table 1.1, and then summarize the results of existing works.

#### 1.5.1.1 Datasets

**KTH dataset** is one of the earliest benchmarks for human action recognition [Schuldt et al. 2004]. It contains 600 short videos of 6 human actions performed by 25 people in four different scenarios.

**Weizmann dataset** is another very early and simple dataset, consisting of 81 short videos associated with 9 actions performed by 9 actors [Blank et al. 2005].

**Kodak Consumer Videos dataset** was recorded by around 100 customers of the Eastman Kodak Company [Loui et al. 2007]. The dataset collected 1,358 video clips labeled with 25 concepts (including activities, scenes, and single objects) as a part of the Kodak concept ontology.

**Hollywood Human Action dataset** contains 8 action classes collected from 32 Hollywood movies, totaling 430 video clips [Laptev et al. 2008]. It was further extended to the Hollywood2 [Marszalek et al. 2009] dataset, which is composed of 12 actions from 69 Hollywood movies with 1,707 video clips in total. This Hollywood series is challenging due to cluttered background and severe camera motion throughout the datasets.

**MCG-WEBV dataset** is another large set of YouTube videos that has 234,414 web videos with annotations on several topic-level events like “a conflict at Gaza” [Cao et al. 2009].

**Olympic Sports** includes 800 video clips and 16 action classes [Niebles et al. 2010]. It was first introduced in 2010 and, unlike in previous datasets, all the videos were downloaded from the Internet.

**HMDB51 dataset** comprises 6,766 videos annotated into 51 classes [Kuehne et al. 2011]. The videos are from a variety of sources, including movies and YouTube consumer videos.

**Columbia Consumer Videos (CCV) dataset** was constructed in 2011, aiming to stimulate research on Internet consumer video analysis [Jiang et al. 2011]. It contains 9,317 user-generated videos from YouTube, which were annotated into 20 classes, including objects (e.g., “cat” and “dog”), scenes (e.g., “beach” and “playground”), sports events (e.g., “basketball” and “soccer”), and social activities (e.g., “birthday” and “graduation”).

**UCF-101 & THUMOS-2014 dataset** is another popular benchmark for human action recognition in videos, consisting of 13,320 video clips (27 hours in total) with 101 annotated classes such as “diving” and “weight lifting” [Soomro et al. 2012]. More recently, the THUMOS-2014 Action Recognition Challenge [Jiang et al. 2014b] created a benchmark by extending the UCF-101 dataset (used as the training set). Additional videos were collected from the Internet, including 2,500 background videos, 1,000 validation videos, and 1,574 test videos.

**TRECVID MED dataset** was released and annually updated by the task of MED, created by NIST since 2010 [Over et al. 2014]. Each year an extended dataset based on datasets from challenges of previous years is constructed and released for worldwide system comparison. For example, in 2014 the MED dataset contained 20 events, such as "birthday party," "bike trick," etc. According to NIST, in the development set, there are around 8,000 videos for training and 23,000 videos used as dry-run validation samples (1,200 hours in total). The MED dataset is only available to the participants of the task, and the labels of the official test set (200,000 videos) are not available even to the participants.

**Sports-1M dataset** consists of 1 million YouTube videos in 487 classes, such as "bowling," "cycling," "rafting," etc., and has been available since 2014 [Karpathy et al. 2014]. The video annotations were automatically derived by analyzing online textual contexts of the videos. Therefore the labels of this dataset are not clean, but the authors claim that the quality of annotation is fairly good.

**ActivityNet dataset** is another large-scale video dataset for human activity recognition and understanding and was released in 2015 [Heilbron et al. 2015]. It consists of 27,801 video clips annotated into 203 activity classes, totaling 849 hours of video. Compared with existing datasets, ActivityNet contains more fine-grained action categories (e.g., "drinking beer" and "drinking coffee").

**EventNet dataset** consists of 500 events and 4,490 event-specific concepts and was released in 2015 [Ye et al. 2015a]. It includes automatic detection models for its video events and some constituent concepts, with around 95,000 training videos from YouTube. Similarly to Sports-1M, EventNet was labeled by online textual information rather than manually labeled.

**MPII Human Pose dataset** includes around 25,000 images containing over 40,000 people with annotated body joints [Andriluka et al. 2014]. According to an established taxonomy of human activities (410 in total), the collected images (from YouTube videos) were provided with activity labels.

**Fudan-Columbia Video Dataset (FCVID) dataset** contains 91,223 web videos annotated manually into 239 categories [Jiang et al. 2015]. The categories cover a wide range of topics, such as social events (e.g., "tailgate party"), procedural events (e.g., "making cake"), object appearances (e.g., "panda"), and scenes (e.g., "beach").

#### 1.5.1.2 Challenges

To advance the state of the art in video classification, several challenges have been introduced with the aim of exploring and evaluating new approaches in realistic settings. We briefly introduce three representative challenges here.

**THUMOS Challenge** was first introduced in 2013 in the computer vision community, aiming to explore and evaluate new approaches for large-scale action recognition of Internet videos [Idrees et al. 2016]. The three editions of the challenge organized in 2013–2015 made THUMOS a common benchmark for action classification and detection.

**TRECVID Multimedia Event Detection (MED) Task** aims to detect whether a video clip contains an instance of a specific event [Awad et al. 2016, Over et al. 2015, Over et al. 2014]. Specifically, based on the released TRECVID MED dataset each year, each participant is required to provide for each testing video the confidence score of how likely one particular event is to happen in the video. Twenty pre-specified events are used each year, and this task adopts the metrics of average precision (AP) and inferred AP for event detection. Each event was also complemented with an event kit, i.e., the textual description of the event as well as the potentially useful information about related concepts that are likely contained in the event.

**ActivityNet Large Scale Activity Recognition Challenge** was first organized as a workshop in 2016 [Heilbron et al. 2015]. This challenge is based on the ActivityNet dataset [Heilbron et al. 2015], with the aim of recognizing high-level and goal-oriented activities. By using 203 activity categories, there are two tasks in this challenge: (1) Untrimmed Classification Challenge, and (2) Detection Challenge, which is to predict the labels and temporal extents of the activities present in videos.

#### 1.5.1.3 Results of Existing Methods

Some of the datasets introduced above have been popularly adopted in the literature. We summarize the results of several recent approaches on UCF-101 and HMDB51 in Table 1.2, where we can see the fast pace of development in this area. Results on video classification are mostly measured by the AP (for a single class) and mean AP (for multiple classes), which are not introduced in detail as they are well known.

### 1.5.2 Captioning

A number of datasets have been proposed for video captioning; these commonly contain videos that have each been paired with its corresponding sentences annotated by humans. This section summarizes the existing datasets and the adopted evaluation metrics, followed by quantitative results of representative methods.

**Table 1.2** **Comparison of recent video classification methods on UCF-101 and HMDB51 datasets**

| Methods | UCF-101 | HMDB51 |
|---|---|---|
| LRCN [Donahue et al. 2017] | 82.9 | — |
| LSTM-composite [Srivastava et al. 2015] | 84.3 | — |
| $F_{ST}CN$ [Sun et al. 2015] | 88.1 | 59.1 |
| C3D [Tran et al. 2015] | 86.7 | — |
| Two-Stream [Simonyan and Zisserman 2014] | 88.0 | 59.4 |
| LSTM [Ng et al. 2015] | 88.6 | — |
| Image-Based [Zha et al. 2015] | 89.6 | — |
| Transformation CNN [Wang et al. 2016c] | 92.4 | 63.4 |
| Multi-Stream [Wu et al. 2016] | 92.6 | — |
| Key Volume Mining [Zhu et al. 2016] | 92.7 | 67.2 |
| Convolutional Two-Stream [Feichtenhofer et al. 2016] | 93.5 | 69.2 |
| Temporal Segment Networks [Wang et al. 2016b] | 94.2 | 69.4 |

**Table 1.3** **Comparison of video captioning benchmarks**

| Dataset | Context | Sentence Source | #Videos | #Clips | #Sentences | #Words |
|---|---|---|---|---|---|---|
| MSVD | Multi-category | AMT workers | — | 1,970 | 70,028 | 607,339 |
| TV16-VTT | Multi-category | Humans | 2,000 | — | 4,000 | — |
| YouCook | Cooking | AMT workers | 88 | — | 2,668 | 42,457 |
| TACoS-ML | Cooking | AMT workers | 273 | 14,105 | 52,593 | — |
| M-VAD | Movie | DVS | 92 | 48,986 | 55,905 | 519,933 |
| MPII-MD | Movie | Script+DVS | 94 | 68,337 | 68,375 | 653,467 |
| MSR-VTT-10K | 20 categories | AMT workers | 7,180 | 10,000 | 200,000 | 1,856,523 |

#### 1.5.2.1 Datasets

Table 1.3 summarizes key statistics and comparisons of popular datasets for video captioning. Figure 1.8 shows a few examples from some of the datasets.

**Microsoft Research Video Description Corpus (MSVD)** contains 1,970 YouTube snippets collected on Amazon Mechanical Turk (AMT) by requesting workers to pick short clips depicting a single activity [Chen and Dolan 2011]. Annotators then label the video clips with single-sentence descrip-

**(a) MSVD dataset**

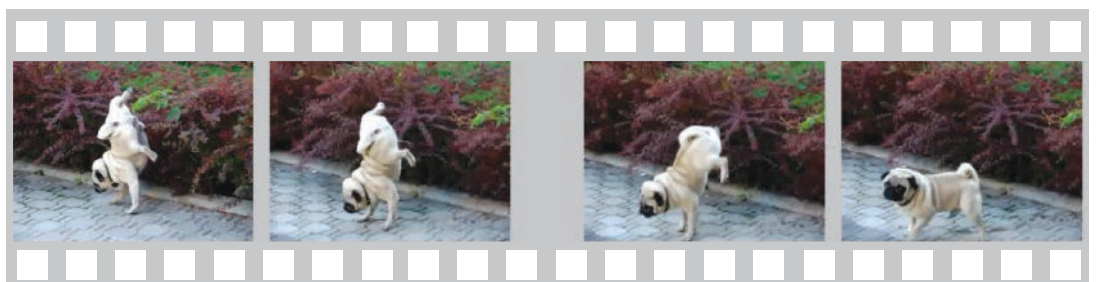

**Sentences:**
- A dog walks around on its front legs.
- The dog is doing a handstand.
- A pug is trying for balance walk on two legs.

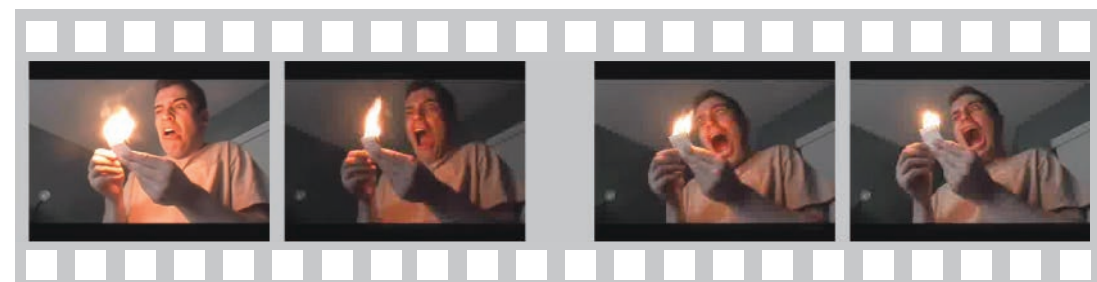

**Sentences:**
- A man lights a match book on fire.
- A man playing with fire sticks.
- A man lights matches and yells.

**(b) M-VAD dataset**

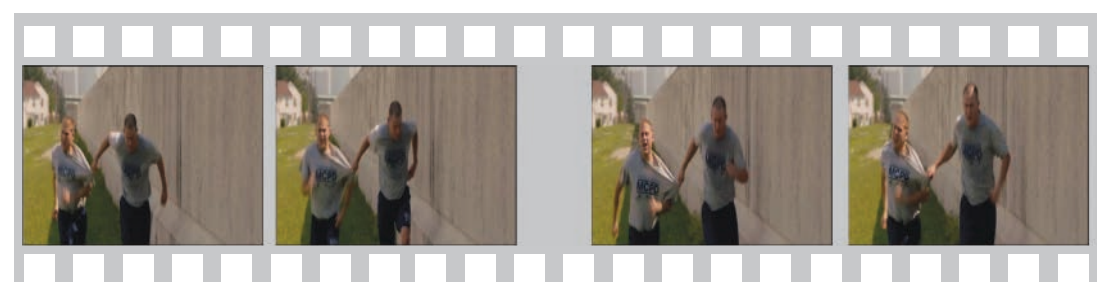

**Sentence:**
- Later he drags someone through a jog.

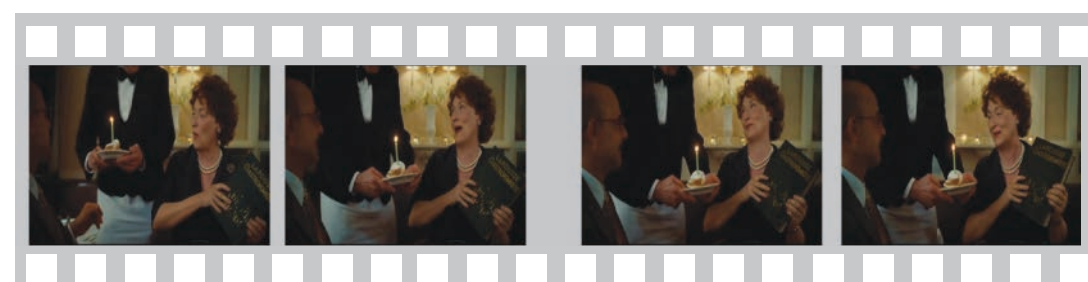

**Sentence:**
- A waiter brings a pastry with a candle.

**(c) MPII-MD dataset**

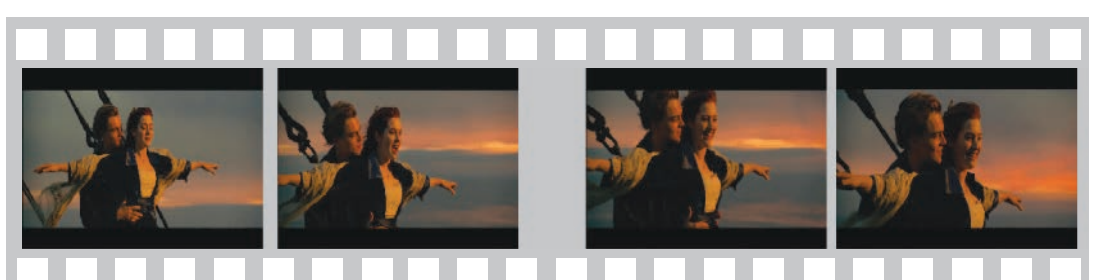

**Sentence:**
- He places his hands around her waist as she opens her eyes.

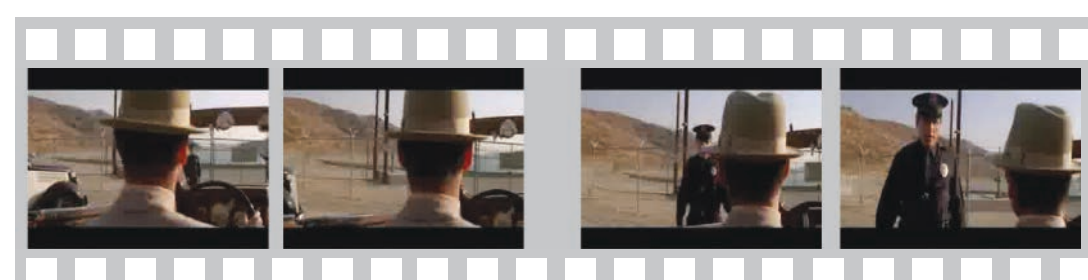

**Sentence:**
- Someone's car is stopped by a couple of uniformed police.

**(d) MSR-VTT-10K dataset**

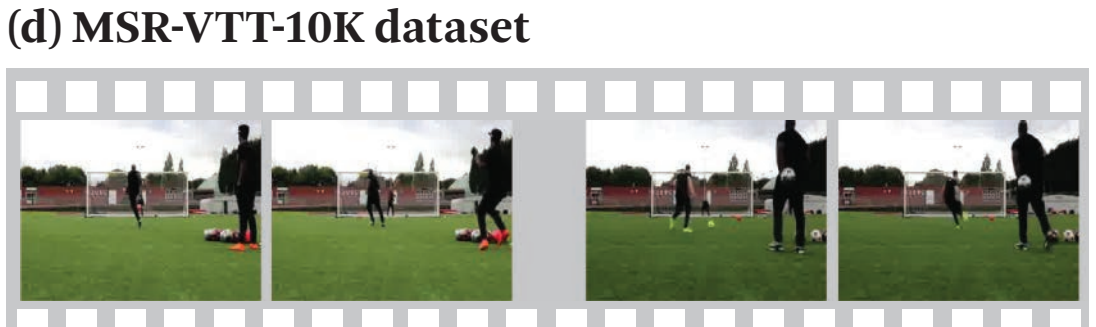

**Sentences:**
- People practising volleyball in the play ground.
- A man is hitting a ball and he falls.
- A man is playing a football game on green land.

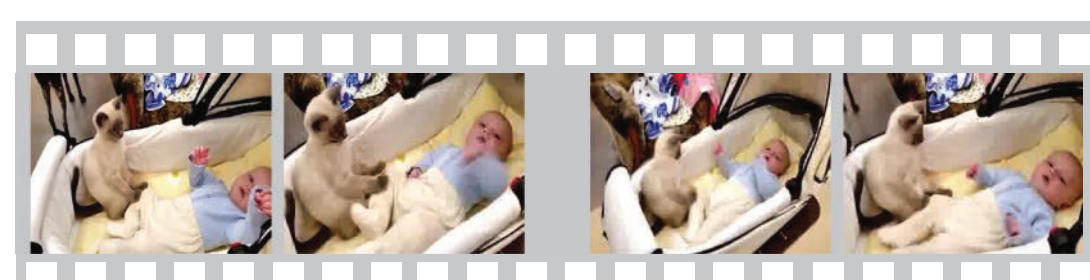

**Sentences:**
- A cat is hanging out in a bassinet with a baby.
- The cat is in the baby bed with the baby.
- A cat plays with a child in a crib.

**Figure 1.8** Examples from (a) MSVD, (b) M-VAD, (c) MPII-MD, and (d) MSR-VTT-10K datasets.

tions. The original corpus has multi-lingual descriptions, but only the English descriptions are commonly exploited on video captioning tasks. Specifically, there are roughly 40 available English descriptions per video and the standard split of MSVD is 1,200 videos for training, 100 for validation, and 670 for testing, as suggested in Guadarrama et al. [2013].

**YouCook dataset** consists of 88 in-house cooking videos crawled from YouTube and is roughly uniformly split into six different cooking styles, such as baking and grilling [Das et al. 2013]. All the videos are in a third-person viewpoint and in different kitchen environments. Each video is annotated with multiple human descriptions by AMT. Each annotator in AMT is instructed to describe the video in at least three sentences totaling a minimum of 15 words, resulting in 2,668 sentences for all the videos.

**TACoS Multi-Level Corpus (TACoS-ML)** is mainly built [Rohrbach et al. 2014] based on MPII Cooking Activities dataset 2.0 [Rohrbach et al. 2015c], which records different activities used when cooking. TACoS-ML consists of 185 long videos with text descriptions collected via AMT workers. Each AMT worker annotates a sequence of temporal intervals across the long video, pairing every interval with a single short sentence. There are 14,105 distinct intervals and 52,593 sentences in total.

**Montreal Video Annotation Dataset (M-VAD)** is composed of about 49,000 DVD movie snippets, which are extracted from 92 DVD movies [Torabi et al. 2015]. Each movie clip is accompanied by one single sentence from semi-automatically transcribed descriptive video service (DVS) narrations. The fact that movies always contain a high diversity of visual and textual content, and that there is only one single reference sentence for each movie clip, has made the video captioning task on the M-VAD dataset very challenging.

**MPII Movie Description Corpus (MPII-MD)** is another collection of movie descriptions dataset that is similar to M-VAD [Rohrbach et al. 2015b]. It contains around 68,000 movie snippets from 94 Hollywood movies and each snippet is labeled with one single sentence from movie scripts and DVS.

**MSR Video to Text (MSR-VTT-10K)** is a recent large-scale benchmark for video captioning that contains 10K Web video clips totalling 41.2 hours, covering the most comprehensive 20 categories obtained from a commercial video search engine, e.g., music, people, gaming, sports, and TV shows [Xu et al. 2016]. Each clip is annotated with about 20 natural sentences by AMT workers. The training/validation/test split is provided by the authors with 6,513 clips for training, 2,990 for validation, and 497 for testing.

**The TRECVID 2016 Video to Text Description (TV16-VTT)** is another recent video captioning dataset that consists of 2,000 videos randomly selected from Twitter Vine videos [Awad et al. 2016]. Each video has a total duration of about 6 seconds and is annotated with 2 sentences by humans. The human annotators are asked to address four facets in the generated sentences: who the video is describing (kinds of persons, animals, things) and what the objects and beings are doing, plus where it is taking place and when.

#### 1.5.2.2 Evaluation Metrics

For quantitative evaluation of the video captioning task, three metrics are commonly adopted: BLEU@$N$ [Papineni et al. 2002], METEOR [Banerjee and Lavie 2005], and CIDEr [Vedantam et al. 2015]. Specifically, BLEU@$N$ is a popular machine translation metric which measures the fraction of N-gram (up to 4-gram) in common between a hypothesis and a reference or set of references. However, as pointed out in Chen et al. [2015], the N-gram matches for a high N (e.g., 4) rarely occur at a sentence level, resulting in poor performance of BLEU@$N$ especially when comparing individual sentences. Hence, a more effective evaluation metric, METEOR, utilized along with BLEU@$N$, is also widely used in natural language processing (NLP) community. Unlike BLEU@$N$, METEOR computes unigram precision and recall, extending exact word matches to include similar words based on WordNet synonyms and stemmed tokens. Another important metric for image/video captioning is CIDEr, which measures consensus in image/video captioning by performing a Term Frequency Inverse Document Frequency (TF-IDF) weighting for each N-gram.

#### 1.5.2.3 Results of Existing Methods

Most popular methods of video captioning have been evaluated on MSVD [Chen and Dolan 2011], M-VAD [Torabi et al. 2015], MPII-MD [Rohrbach et al. 2015b], and TACoS-ML [Rohrbach et al. 2014] datasets. We summarize the results on these four datasets in Tables 1.4, 1.5, and 1.6. As can be seen, most of the works are very recent, indicating that video captioning is an emerging and fast-developing research topic.

**Table 1.4** **Reported results on the MSVD dataset, where B@$N$, M, and C are short for BLEU@$N$, METEOR, and CIDEr-D scores, respectively**

| Methods | B@1 | B@2 | B@3 | B@4 | M | C |
|---|---|---|---|---|---|---|
| FGM [Thomason et al. 2014] | — | — | — | 13.7 | 23.9 | — |
| LSTM-YT [Venugopalan et al. 2015b] | — | — | — | 33.3 | 29.1 | — |
| MM-VDN [Xu et al. 2015a] | — | — | — | 37.6 | 29.0 | — |
| S2VT [Venugopalan et al. 2015a] | — | — | — | — | 29.8 | — |
| S2FT [Liu and Shi 2016] | — | — | — | — | 29.9 | — |
| SA [Yao et al. 2015a] | 80.0 | 64.7 | 52.6 | 41.9 | 29.6 | 51.7 |
| Glove+Deep Fusion [Venugopalan et al. 2016] | — | — | — | 42.1 | 31.4 | — |
| LSTM-E [Pan et al. 2016b] | 78.8 | 66.0 | 55.4 | 45.3 | 31.0 | — |
| GRU-RCN [Ballas et al. 2016] | — | — | — | 43.3 | 31.6 | 68.0 |
| h-RNN [Yu et al. 2016] | 81.5 | 70.4 | 60.4 | 49.9 | 32.6 | 65.8 |

All values are reported as percentages (%).

**Table 1.5** **Reported results on (a) M-VAD and (b) MPII-MD datasets, where M is short for METEOR**

| M-VAD dataset | |
|---|---|
| Methods | M |
| SA [Yao et al. 2015a] | 4.3 |
| Mean Pool [Venugopalan et al. 2015a] | 6.1 |
| Visual-Labels [Rohrbach et al. 2015a] | 6.4 |
| S2VT [Venugopalan et al. 2015a] | 6.7 |
| Glove+Deep Fusion [Venugopalan et al. 2016] | 6.8 |
| LSTM-E [Pan et al. 2016b] | 6.7 |

| MPII-MD dataset | |
|---|---|
| Methods | M |
| SMT [Rohrbach et al. 2015b] | 5.6 |
| Mean Pool [Venugopalan et al. 2015a] | 6.7 |
| Visual-Labels [Rohrbach et al. 2015a] | 7.0 |
| S2VT [Venugopalan et al. 2015a] | 7.1 |
| Glove+Deep Fusion [Venugopalan et al. 2016] | 6.8 |
| LSTM-E [Pan et al. 2016b] | 7.3 |

All values are reported as percentages (%).

**Table 1.6** Reported results on the TACoS-ML dataset, where B@$N$, M, and C are short for BLEU@$N$, METEOR, and CIDEr-D scores, respectively

| Methods | B@1 | B@2 | B@3 | B@4 | M | C |
|---|---|---|---|---|---|---|
| CRF-T [Rohrbach et al. 2013] | 56.4 | 44.7 | 33.2 | 25.3 | 26.0 | 124.8 |
| CRF-M [Rohrbach et al. 2014] | 58.4 | 46.7 | 35.2 | 27.3 | 27.2 | 134.7 |
| LRCN [Donahue et al. 2017] | 59.3 | 48.2 | 37.0 | 29.2 | 28.2 | 153.4 |
| h-RNN [Yu et al. 2016] | 60.8 | 49.6 | 38.5 | 30.5 | 28.7 | 160.2 |

All values are reported as percentages (%).

## 1.6 Conclusion

In this chapter, we have reviewed state-of-the-art deep learning techniques on two key topics related to video analysis, video classification and video captioning, both of which rely on the modeling of the abundant spatial and temporal information in videos. In contrast to hand crafted features that are costly to design and have limited generalization capability, the essence of deep learning for video classification is to derive robust and discriminative feature representations from raw data through exploiting massive videos with an aim to achieve effective and efficient recognition, which could hence serve as a fundamental component in video captioning. Video captioning, on the other hand, focuses on bridging visual understanding and language description by joint modeling. We also provided a review of popular benchmarks and challenges for both video classification and captioning tasks. Though extensive efforts have been made in video classification and captioning with deep learning, we believe we are just beginning to unleash the power of deep learning in the big video data era. Given the substantial amounts of videos generated at an astounding speed every hour and every day, it remains a challenging open problem how to derive better video representations with deep learning modeling the abundant interactions of objects and their evolution over time with limited (or without any) supervisory signals to facilitate video content understanding (i.e., the recognition of human activities and events as well as the generation of free-form and open-vocabulary sentences for describing videos). We hope this chapter sheds light on the nuts and bolts of video classification and captioning for both current and new researchers.

# Bibliography

M. Abadi, R. Subramanian, S. Kia, P. Avesani, I. Patras, and N. Sebe. 2015. DECAF: MEG-based multimodal database for decoding affective physiological responses. *IEEE Transactions on Affective Computing*, 6(3): 209–222. DOI: 10.1109/TAFFC.2015.2392932. 221, 238, 240, 249

A. E. Abdel-Hakim and A. A. Farag. 2006. CSIFT: A SIFT descriptor with color invariant characteristics. In *Proceedings of the IEEE Computer Society Conference on Computer Vision and Pattern Recognition (CVPR)*, vol. 2, pp. 1978–1983. DOI: 10.1109/CVPR.2006.95. 234

E. Abdulin and O. Komogortsev. 2015. User eye fatigue detection via eye movement behavior. In *CHI Extended Abstracts*, pp. 1265–1270. DOI: 10.1145/2702613.2732812. 250

E. Adam. 1993. Fighter cockpits of the future. In *Digital Avionics Systems Conference, 1993. 12th DASC., AIAA/IEEE*, pp. 318–323. IEEE. DOI: 10.1109/DASC.1993.283529. 165, 167

A. Adelsbach and A.-R. Sadeghi. 2001. Zero-knowledge watermark detection and proof of ownership. In *Information Hiding*, volume 2137 of *Lecture Notes in Computer Science*, pp. 273–288. Springer. DOI: 10.1007/3-540-45496-9_20. 81

A. Adi and O. Etzion. 2004. Amit—the situation manager. *The VLDB Journal—The International Journal on Very Large Data Bases*, 13(2): 177–203. DOI: 10.1007/s00778-003-0108-y. 165

C. Aggarwal, J. H. Jiawei, and J. W. P. S. Yu. 2003. A framework for clustering evolving data streams. In *Proceedings of the VLDB Endowment* 29*th International Conference on Very Large Data Bases*, volume 29, pp. 81–92. 97, 98

C. Aggarwal, D. Olshefski, D. Saha, Z.-Y. Shae, and P. Yu. 2005. CSR: Speaker recognition from compressed VoIP packet stream. In *IEEE International Conference on Multimedia and Expo*, pp. 970–973. DOI: 10.1109/ICME.2005.1521586. 97, 98

J. K. Aggarwal and M. S. Ryoo. 2011. Human activity analysis: A review. *ACM Computing Surveys*, 43(3): April 2011. DOI: 10.1145/1922649.1922653. 4

A. Aghasaryan, M. Bouzid, D. Kostadinov, M. Kothari, and A. Nandi. July 2013. On the use of LSH for privacy preserving personalization. In *IEEE International Conference on Trust, Security, and Privacy in Computing and Communications (TrustCom)*, pp. 362–371. DOI: 10.1109/TrustCom.2013.46. 105

E. Agichtein, E. Brill, S. Dumais, and R. Ragno. 2006. Learning user interaction models for predicting web search result preferences. In *Proceedings of the 29th Annual International ACM SIGIR Conference on Research and Development in Information Retrieval*, pp. 3–10. ACM. DOI: 10.1145/1148170.1148175. 147

E. Agrell, T. Eriksson, A. Vardy, and K. Zeger. 2002. Closest point search in lattices. *IEEE Trans. Inform. Theory*, 48(8): 2201–2214. DOI: 10.1109/TIT.2002.800499. 115

M. Aguilar, C. XLIM, S. Fau, C. Fontaine, G. Gogniat, and R. Sirdey. 2013. Recent advances in homomorphic encryption: A possible future for signal processing in the encrypted domain. *IEEE Signal Processing Magazine*, 30(2): 108–117. DOI: 10.1109/MSP.2012.2230219. 77

N. Ahituv, Y. Lapid, and S. Neumann. 1987. Processing encrypted data. *Communications of the ACM*, 30(9): 777–780. DOI: 10.1145/30401.30404. 78, 84

L. Ai, J. Yu, Z. Wu, Y. He, and T. Guan. 2015. Optimized residual vector quantization for efficient approximate nearest neighbor search. *Multimedia Systems*, pp. 1–13. DOI: 10.1007/s00530-015-0470-9. 131

X. Alameda-Pineda and R. Horaud. 2015. Vision-guided robot hearing. *International Journal of Robotics Research*, 34(4–5): 437–456. DOI: 10.1177/0278364914548050. 55

X. Alameda-Pineda, J. Sanchez-Riera, J. Wienke, V. Franc, J. Cech, K. Kulkarni, A. Deleforge, and R. Horaud. 2013. Ravel: An annotated corpus for training robots with audiovisual abilities. *Journal on Multimodal User Interfaces*, 7(1–2): 79–91. DOI: 10.1007/s12193-012-0111-y. 55

X. Alameda-Pineda, Y. Yan, E. Ricci, O. Lanz, and N. Sebe. 2015. Analyzing free-standing conversational groups: A multimodal approach. In *ACM International Conference on Multimedia (ACMMM)*, pp. 4–15. DOI: 10.1145/2733373.2806238. 55, 56, 61

X. Alameda-Pineda, E. Ricci, Y. Yan, and N. Sebe. 2016a. Recognizing emotions from abstract paintings using non-linear matrix completion. In *CVPR*. 58

X. Alameda-Pineda, J. Staiano, R. Subramanian, L. Batrinca, E. Ricci, B. Lepri, O. Lanz, and N. Sebe. 2016b. SALSA: A novel dataset for multimodal group behavior analysis. *IEEE Trans. Pattern Analysis and Machine Intelligence, (IEEE TPAMI)*, 38(8): 1707–1720. DOI: 10.1109/TPAMI.2015.2496269. 56, 66, 238

M. Alhamad, T. Dillon, and E. Chang. 2010. SLA-based trust model for cloud computing. In *Proceedings of IEEE International Conference on Network-Based Information Systems (NBiS)*, pp. 321–324. DOI: 10.1109/NBiS.2010.67. 260

Y. Aloimonos, A. K. Mishra, L. F. Cheong, and A. Kassim. 2012. Active visual segmentation. *IEEE PAMI*, 34: 639–653. DOI: 10.1109/TPAMI.2011.171. 220

D. Anderson. 2004. BOINC: A system for public-resource computing and storage. In *Proceedings of IEEE/ACM Grid Computing (GC)*, pp. 4–10. DOI: 10.1109/GRID.2004.14. 258, 261

D. Anderson, J. Cobb, E. Korpela, M. Lebofsky, and D. Werthimer. 2002. SETI@home: An experiment in public-resource computing. *ACM Transactions on Communications*, 45(11): 56–61. DOI: 10.1145/581571.581573. 258

N. Anderson, W. F. Bischof, K. E. Laidlaw, E. F. Risko, and A. Kingstone. 2013. Recurrence quantification analysis of eye movements. *Behavior Research Methods*, 45(3): 842–856. DOI: 10.3758/s13428-012-0299-5. 229

A. Andoni. Nov. 2009. *Nearest Neighbor Search: The Old, the New, and the Impossible*. PhD thesis, MIT. 108, 113, 115

A. Andoni and P. Indyk. 2006. Near-optimal hashing algorithms for near neighbor problem in high dimensions. In *Proceedings of the Symposium on the Foundations of Computer Science*, pp. 459–468. DOI: 10.1109/FOCS.2006.49. 105, 112, 115, 117

P. André, E. Cutrell, D. S. Tan, and G. Smith. 2009. Designing novel image search interfaces by understanding unique characteristics and usage. In *IFIP Conference on Human-Computer Interaction*, pp. 340–353. Springer. 148

M. Andriluka, L. Pishchulin, P. Gehler, and B. Schiele. 2014. 2D human pose estimation: New benchmark and state of the art analysis. In *CVPR*, pp. 3686–3693. DOI: 10.1109/CVPR.2014.471. 22

S. Argamon, S. Dhawle, M. Koppel, and J. Pennbaker. 2005. Lexical predictors of personality type. In *Interface and the Classification Society of North America*. DOI: 10.1.1.60.6697. 238

V. Athitsos, J. Alon, S. Sclaroff, and G. Kollios. Jan. 2008. BoostMap: An embedding method for efficient nearest neighbor retrieval. *IEEE Trans. PAMI*, 30(1): 89–104. DOI: 10.1109/TPAMI.2007.1140. 133

S. Avidan and M. Butman. 2006. Blind vision. In *Proceedings of the* 9*th European Conference on Computer Vision*, volume 3953 of *Lecture Notes in Computer Science*, pp. 1–13. Springer. DOI: 10.1007/11744078_1. 82, 91, 93

G. Awad, J. Fiscus, M. Michel, D. Joy, W. Kraaij, A. F. Smeaton, G. Quenot, M. Eskevich, R. Aly, G. J. F. Jones, R. Ordelman, B. Huet, and M. Larson. 2016. TRECVID 2016: Evaluating video search, video event detection, localization, and hyperlinking. In *Proceedings of TRECVID 2016*, pp. 407–411. 23, 27

S. Ba, X. Alameda-Pineda, A. Xompero, and R. Horaud. 2016. An on-line variational Bayesian model for multi-person tracking from cluttered scenes. In *Computer Vision and Human Understanding (CVHU)* 153:64–76. DOI: 10.1016/j.cviu.2015.07.006. 52, 55

O. Babaoglu, M. Marzolla, and M. Tamburini. 2012. Design and implementation of a P2P cloud system. In *Proceedings of ACM Symposium on Applied Computing (SAC)*, pp. 412–417. DOI: 10.1145/2245276.2245357. 259

A. Babenko and V. Lempitsky. June 2014. Additive quantization for extreme vector compression. In *CVPR*. DOI: 10.1109/CVPR.2014.124. 131

A. Babenko and V. Lempitsky. June 2015a. The inverted multi-index. In *IEEE TPAMI*, 37(6): 1247–1260. DOI: 10.1109/CVPR.2012.6248038. 132

A. Babenko and V. Lempitsky. June 2015b. Tree quantization for large-scale similarity search and classification. In *CVPR*. DOI: 10.1109/CVPR.2015.7299052. 131

A. Babenko, A. Slesarev, A. Chigorin, and V. Lempitsky. September 2014. Neural codes for image retrieval. In *Proceedings of the European Conference on Computer Vision (ECCV)*, pp. 584–599. DOI: 10.1007/978-3-319-10590-1_38. 107

M. Backes, G. Doychev, M. Dürmuth, and B. Köpf. 2010. Speaker recognition in encrypted voice streams. In *Computer Security—ESORICS*, volume 6345 of *Lecture Notes in Computer Science*, pp. 508–523. Springer. 98, 100

A. Baddeley and R. Turner. 2000. Practical maximum pseudolikelihood for spatial point patterns. *Australian & New Zealand Journal of Statistics*, 42: 283–322. DOI: 10.1111/1467-842X.00128. 208

S. Bahmani and B. Raj. 2013. A unifying analysis of projected gradient descent for $\ell$p-constrained least squares. *Applied and Computational Harmonic Analysis*, 34(3): 366–378. DOI: 10.1016/j.acha.2012.07.004. 39

S. Bahmani, P. Boufounos, and B. Raj. 2011. Greedy sparsity-constrained optimization. In *Proceedings 45th IEEE Asilomar Conference on Signals, Systems, and Computers (ASILOMAR)*, pp. 1148–1152. 39

S. Bahmani, B. Raj, and P. Boufounos. 2016. Learning model-based sparsity via projected gradient descent. *IEEE Transactions on Information Theory*, 62(4): 2092–2099. DOI: 10.1109/TIT.2016.2515078. 39

N. Ballas, L. Yao, C. Pal, and A. Courville. 2016. Delving deeper into convolutional networks for learning video representations. In *Proceedings of the International Conference on Learning Representations (ICLR). arXiv:1511.06432, 2015.* 14, 28

T. Baltrušaitis, P. Robinson, and L.-P. Morency. 2016. OpenFace: an open source facial behavior analysis toolkit. In *IEEE Winter Conference on Applications of Computer Vision*, pp. 1–10. DOI: 10.1109/WACV.2016.7477553. 225

R. Balu, T. Furon, and H. Jégou. April 2014. Beyond project and sign for distance estimation with binary codes. In *Proc IEEE International Conference on Acoustics, Speech, and Signal Processing (ICASSP)*, pp. 6884–6888. DOI: 10.1109/ICASSP.2014.6854934. 123, 127

S. Banerjee and A. Lavie. 2005. METEOR: An automatic metric for MT evaluation with improved correlation with human judgments. In *Proceedings of the ACL Workshop on Intrinsic and Extrinsic Evaluation Measures for Machine Translation and/or Summarization*, pp. 65-72. 27

K. Baokye, B. Trueba-Hornero, O. Vinyals, and G. Friedland. 2008a. Overlapped speech detection for improved diarization in multiparty meetings. In *ICASSP*, pp. 4353–4356. DOI: 10.1109/ICASSP.2008.4518619. 36

K. Baokye, O. Vinyals, and G. Friedland. 2008b. Two's a crowd: Improving speaker diarization by automatically identifying and excluding overlapped speech. In *Proceedings of Interspeech*, pp. 32–35. 36

R. Bardeli, D. Wolff, F. Kurth, M. Koch, K.-H. Tauchert, and K.-H. Frommolt. 2010. Detecting bird sounds in a complex acoustic environment and application to bioacoustic monitoring. *Pattern Recognition Letters*, 31(12): 1524–1534. DOI: 10.1016/j.patrec.2009.09.014. 38

M. Barni and A. Piva. 2008. Co-chairs, special issue on signal processing in the encrypted domain. *Proceedings of EURASIP European Signal Processing Conference*. 77

M. Barni, T. Bianchi, D. Catalano, M. Di Raimondo, R. D. Labati, P. Failla, D. Fiore, R. Lazzeretti, V. Piuri, A. Piva, and F. Scotti. 2010. A privacy-compliant fingerprint recognition system based on homomorphic encryption and fingercode templates. In $4^{th}$ *IEEE International Conference on Biometrics: Theory Applications and Systems*, pp. 1–7. DOI: 10.1109/BTAS.2010.5634527. 82

M. Barni, P. Failla, R. Lazzeretti, A. R. Sadeghi, and T. Schneider. 2011. Privacy-preserving ECG classification with branching programs and neural networks. *IEEE Transactions on Information Forensics and Security*, 6(2): 452–468. DOI: 10.1109/TIFS.2011.2108650. 83

R. Barthes. 1981. *Camera Lucida: Reflections on Photography*. Macmillan. 187

J. Barwise and J. Perry. 1980. The situation underground. *Stanford Working Papers in Semantics*, 1: 1–55. 165, 167

J. Barwise and J. Perry. 1981. Situations and attitudes. *The Journal of Philosophy*, 78(11): 668–691. DOI: 10.2307/2026573. 165

R. Bayer and E. M. McCreight. July 1970. Organization and maintenance of large ordered indices. Mathematical and Information Sciences Report 20, Boeing Scientific Research Laboratories. DOI: 10.1145/1734663.1734671. 107

M. Bebbington and D. S. Harte. 2001. On the statistics of the linked stress release model. *Journal of Applied Probability*, 38(A): 176–187. DOI: 10.1017/S0021900200112768. 208

J. S. Beis and D. G. Lowe. June 1997. Shape indexing using approximate nearest-neighbour search in high-dimensional spaces. In *CVPR*. DOI: 10.1109/CVPR.1997.609451. 105, 107, 118

A. J. Bell and T. J. Sejnowski. 1995. An information-maximization approach to blind separation and blind deconvolution. *Neural Computation*, 7(6): 1129–1159. citeseer.nj.nec.com/bell95informationmaximization.html. 36

A. Beloglazov, J. Abawajy, and R. Buyya. 2012. Energy-aware resource allocation heuristics for efficient management of data centers for cloud computing. *Future Generation Computer Systems*, 28(5): 755–768. DOI: 10.1016/j.future.2011.04.017. 259

J. Benaloh. 1994. Dense probabilistic encryption. In *Proceedings of the Workshop on Selected Areas of Cryptography*, pp. 120–128. 81

B. Benfold and I. Reid. 2011. Unsupervised learning of a scene-specific coarse gaze estimator. In *Proceedings of IEEE Internationa Conference on Computer Vision (ICCV)*, pp. 2344–2351. DOI: 10.1109/ICCV.2011.6126516. 57

Y. Bengio, P. Simard, and P. Frasconi. 1994. Learning long-term dependencies with gradient descent is difficult. *IEEE Transactions on Neural Networks (TNN)*, 5(2): 157–166. DOI: 10.1109/72.279181. 7

Y. Bengio, A. Courville, and P. Vincent. 2013. Representation learning: A review and new perspectives. *IEEE TPAMI*, pp. 1798–1828. DOI: 10.1109/TPAMI.2013.50. 6

C. J. Bennett. 2008. *The Privacy Advocates: Resisting the Spread of Surveillance*. MIT Press. 91

W. Bennett. July 1948. Spectra of quantized signals. *Bell System Technical Journal*, 27(3): 1–43. DOI: 10.1002/j.1538-7305.1948.tb01340.x. 117

A. Berenzweig, B. Logan, D. P. W. Ellis, and B. Whitman. June 2004. A large-scale evaluation of acoustic and subjective music-similarity measures. *Computer Music Journal*, 28(2): 63–76. http://www.ee.columbia.edu/~dpwe/pubs/BerenLEW04-museval.pdf. DOI: 10.1162/014892604323112257. 38

T. Bertin-Mahieux and D. P. W. Ellis. Oct. 2011. Large-scale cover song recognition using hashed chroma landmarks. In *Proceedings IEEE Workshop on Applications of Signal Proceedings to Audio and Acoustics*, pp. 117–120. DOI: 10.7916/D8J67S98. 38

T. Bianchi, A. Piva, and M. Barni. 2008a. Efficient pointwise and blockwise encrypted operations. In *Proceedings of the ACM* $10^{th}$ *Workshop on Multimedia and Security*, pp. 85–90. DOI: 10.1145/1411328.1411344. 79, 80

T. Bianchi, A. Piva, and M. Barni. 2008b. Comparison of different FFT implementations in the encrypted domain. In *Proceedings of EUSIPCO* $16^{th}$ *European Signal Processing Conference*, pp. 1–5. 80

T. Bianchi, A. Piva, and M. Barni. 2008c. Implementing the discrete Fourier transform in the encrypted domain. In *IEEE International Conference on Acoustics*, pp. 1757–1760. 80

T. Bianchi, A. Piva, and M. Barni. 2009a. Efficient linear filtering of encrypted signals via composite representation. In *Proceedings of the IEEE* $16^{th}$ *International Conference on Digital Signal Processing*, pp. 1–6. DOI: 10.1109/ICDSP.2009.5201116. 80

T. Bianchi, T. Veugen, A. Piva, and M. Barni. 2009b. Processing in the encrypted domain using a composite signal representation: PROS and CONS. In *Proceedings of the IEEE* $1^{st}$ *International Workshop on Information Forensics and Security*, pp. 176–180. DOI: 10.1109/WIFS.2009.5386460. 76, 77, 79, 80

T. Bianchi, S. Turchi, A. Piva, R. D. Labati, V. Piuri, and F. Scotti. 2010. Implementing fingercode-based identity matching in the encrypted domain. In *I E E E Workshop on Biometric Measurements and Systems for Security and Medical Applications*, pp. 15–21. DOI: 10.1109/BIOMS.2010.5610445. 77, 82

H. Bilen, B. Fernando, E. Gavves, A. Vedaldi, and S. Gould. 2016. Dynamic image networks for action recognition. In *CVPR*, pp. 3034–3042. 12

D. Bitouk, N. Kumar, S. Dhillon, P. Belhumeur, and S. Nayar. 2008. Face swapping: automatically replacing faces in photographs. In *ACM Transactions on Graphics*, volume 27, p. 39. DOI: 10.1145/1399504.1360638. 92, 93

M. Blank, L. Gorelick, E. Shechtman, M. Irani, and R. Basri. 2005. Actions as space-time shapes. In *ICCV*. DOI: 10.1109/TPAMI.2007.70711. 20

D. Bogdanov, 2007. Foundations and properties of Shamir's secret sharing scheme. Research Seminar in Cryptography. 88

F. Bonomi, R. Milito, J. Zhu, and S. Addepalli. 2012. Fog computing and its role in the Internet of Things. In *Proceedings of ACM Workshop on Mobile Cloud Computing (MCC)*, pp. 13–16. DOI: 10.1145/2342509.2342513. 256

D. Borth, R. Ji, T. Chen, T. Breuel, and S.-F. Chang. 2013. Large-scale visual sentiment ontology and detectors using adjective noun pairs. In *Proceedings of the 21st ACM International Conference on Multimedia*, pp. 223–232. ACM. DOI: 10.1145/2502081.2502282. 150

P. T. Boufounos. March 2012. Universal rate-efficient scalar quantization. *IEEE Trans. Inform. Theory*, 58(3): 1861–1872. DOI: 10.1109/TIT.2011.2173899. 122

P. T. Boufounos and R. G. Baraniuk. March 2008. 1-bit compressive sensing. In *Proceedings Conference on Information Science and Systems (CISS)*, pp. 16–21. DOI: 10.1109/CISS.2008.4558487. 121

T. E. Boult. 2005. Pico: Privacy through invertible cryptographic obscuration. In *Proceedings of IEEE Computer Vision for Interactive and Intelligent Environment*, pp. 27–38. DOI: 10.1109/CVIIE.2005.16. 91

A. Bourrier, F. Perronnin, R. Gribonval, P. Pérez, and H. Jégou. 2015. Nearest neighbor search for arbitrary kernels with explicit embeddings. *Journal of Mathematical Imaging and Vision*, 52(3): 459–468. 133

S. Boyd, N. Parikh, E. Chu, B. Peleato, and J. Eckstein. 2011. Distributed optimization and statistical learning via the alternating direction method of multipliers. *Foundations and Trends in Machine Learning*, 3(1): 1–122. DOI: 10.1561/2200000016. 63, 64

G. Bradski and A. Kaehler. 2008. *Learning OpenCV: Computer Vision with the OpenCV Library*. O'Reilly Media. 179

Z. Brakerski, C. Gentry, and V. Vaikuntanathan. 2012. (Leveled) fully homomorphic encryption without bootstrapping. In *Proceedings of 3rd ACM Innovations in Theoretical Computer Science Conference*, pp. 309–325. DOI: 10.1145/2090236.2090262. 101, 102

J. Brandt. June 2010. Transform coding for fast approximate nearest neighbor search in high dimensions. In *CVPR*, pp. 1815–1822. DOI: 10.1109/CVPR.2010.5539852. 128

O. Brdiczka, P. Yuen, S. Zaidenberg, P. Reignier, and J. Crowley. 2006. Automatic acquisition of context models and its application to video surveillance. In *18th International Conference on Pattern Recognition, 2006 (ICPR)*, volume 1, pp. 1175–1178. IEEE. DOI: 10.1109/ICPR.2006.292. 165

P. Brémaud and L. Massoulié. 1996. Stability of nonlinear Hawkes processes. *The Annals of Probability*, 24(3): 1563–1588. DOI: 10.1214/aop/1065725193. 200

J. Bringer, H. Chabanne, M. Favre, A. Patey, T. Schneider, and M. Zohner. 2014. GSHADE: Faster privacy-preserving distance computation and biometric identification. In *Proceedings of the 2nd ACM Workshop on Information Hiding and Multimedia Security*, pp. 187–198. ACM. DOI: 10.1145/2600918.2600922. 82

A. Z. Broder. June 1997. On the resemblance and containment of documents. In *Proceedings Compression and Complexity of Sequences*, p. 21. DOI: 10.1109/SEQUEN.1997.666900. 120

G. J. Brown and M. P. Cooke. 1994. Computational auditory scene analysis. *Computer Speech and Language*, 8: 297–336. DOI: 10.1006/csla.1994.1016. 36, 37

A. Bujari, M. Massaro, and C. E. Palazzi. Dec. 2015. Vegas over access point: Making room for thin client game systems in a wireless home. *IEEE Transactions on Circuits and Systems for Video Technology*, 25(12): 2002–2012. DOI: 10.1109/TCSVT.2015.2450332. 309

R. Buyya, S. Garg, and R. Calheiros. 2011. SLA-oriented resource provisioning for cloud computing: Challenges. In *Proceedings of IEEE International Conference on Cloud and Service Computing (CSC)*, pp. 1–10. DOI: 10.1109/CSC.2011.6138522. 259

R. Cabral, F. De la Torre, J. Costeira, and A. Bernardino. 2014. Matrix completion for multi-label image classification. *IEEE TPAMI*, 37(1): 121–135. DOI: 10.1109/TPAMI.2014.2343234. 58, 60

W. Cai, M. Chen, and V. C. M. Leung. May 2014. Toward gaming as a service. *IEEE Internet Computing*, 18(3): 12–18. http://ieeexplore.ieee.org/lpdocs/epic03/wrapper.htm?arnumber=6818918. DOI: 10.1109/MIC.2014.22. 290, 291

W. Cai, R. Shea, C.-Y. Huang, K.-T. Chen, J. Liu, V. C. M. Leung, and C.-H. Hsu. April 2016a. The future of cloud gaming. *Proceedings of IEEE*, 104(4): 687–691. DOI: 10.1109/JPROC.2016.2539418. 304

W. Cai, R. Shea, C.-Y. Huang, K.-T. Chen, J. Liu, V. C. M. Leung, and C.-H. Hsu. Jan. 2016b. A survey on cloud gaming: Future of computer games. *IEEE Access*, pp. 1–25. DOI: 10.1109/ACCESS.2016.2590500. 289, 291

A. T. Campbell, S. B. Eisenman, N. D. Lane, E. Miluzzo, and R. A. Peterson. 2006. People-centric urban sensing. In *Int. Work. on Wireless Internet*, article 18. DOI: 10.1145/1234161.1234179. 55

E. J. Candès and T. Tao. 2010. The power of convex relaxation: Near-optimal matrix completion. *IEEE Trans. Inf. Theory*, 56(5): 2053–2080. DOI: 10.1109/TIT.2010.2044061. 61

J. Canny. 2002. Collaborative filtering with privacy. In *Proceedings of IEEE Symposium on Security and Privacy*, pp. 45–57. DOI: 10.1145/564376.564419. 82

C. Canton-Ferrer, C. Segura, J. R. Casas, M. Pardas, and J. Hernando. 2008. Audiovisual head orientation estimation with particle filtering in multisensor scenarios. *EURASIP Journal on Advances in Signal Processing*, article 32. 57

J. Cao, Y.-D. Zhang, Y.-C. Song, Z.-N. Chen, X. Zhang, and J.-T. Li. 2009. MCG-WEBV: A benchmark dataset for web video analysis. *Technical Report, CAS Institute of Computing Technology*, pp. 1–10. DOI: 10.1155/2008/276846. 21

N. Carrier, T. Deutsch, C. Gruber, M. Heid, and L. Jarrett. Aug. 2008. The business case for enterprise mashups. IBM White Paper. 171

M. Casey and M. Slaney. April 2007. Fast recognition of remixed music audio. In *ICASSP*, volume 4, pp. 1425–1428. DOI: 10.1109/ICASSP.2007.367347. 105

C. Castelluccia, A. Chan, E. Mykletun, and G. Tsudik. 2009. Efficient and provably secure aggregation of encrypted data in wireless sensor networks. *ACM Transactions on Sensor Networks*, 5(3): 20. 103

S. Chachada and C.-C. J. Kuo. 2014. Environmental sound recognition: A survey. *APSIPA Transactions on Signal and Information Processing*, 3. ISSN 2048-7703. http://journals.cambridge.org/article_S2048770314000122. DOI: 10.1017/ATSIP.2014.12. 38

S.-F. Chang. 2013. How far we've come: Impact of 20 years of multimedia information retrieval. *ACM Transactions on Multimedia Computing, Communications, and Applications (TOMCCAP)*, 9(1s): 42. DOI: 10.1109/CCGRID.2010.132. 151, 155

V. Chang, D. Bacigalupo, G. Wills, and D. Roure. 2010. Categorisation of cloud computing business models. In *Proceedings of IEEE/ACM International Conference on Cluster, Cloud and Grid Computing (CCGRID)*, pp. 509–512. DOI: 10.1145/2491844. 259

Y.-C. Chang, K.-T. Chen, C.-C. Wu, and C.-L. Lei. 2008. Inferring speech activity from encrypted Skype traffic. In *Proceedings of IEEE Global Telecommunications Conference*, pp. 1–5. 97, 98

Y.-C. Chang, P.-H. Tseng, K.-T. Chen, and C.-L. Lei. May 2011. Understanding the performance of thin-client gaming. In *Proceedings of IEEE Communications Quality and Reliability (CQR) 2011*, pp. 1—6. DOI: 10.1109/CQR.2011.5996092. 291

M. S. Charikar. May 2002. Similarity estimation techniques from rounding algorithms. In (*Proceedings of the 34th Annual ACM Symposium on Theory of Computing (STOC)*, pp. 380–388. DOI: 10.1145/509907.509965. 105, 110, 112, 120, 121, 123, 133

A. Chattopadhyay and T. E. Boult. 2007. PrivacyCam: A privacy preserving camera using uCLinux on the Blackfin DSP. In *Proceeding of the IEEE Conference on Computer Vision and Pattern Recognition*, pp. 1–8. DOI: 10.1109/CVPR.2007.383413. 91, 93

S. Chaudhuri. 2013. *Structured Models for Semantic Analysis of Audio Content*. PhD thesis, Carnegie Mellon University. 46, 47

S. Chaudhuri and B. Raj. 2011. Learning contextual relevance of audio segments using discriminative models over AUD sequences. In *IEEE Workshop on Applications of Signal Processing to Audio and Acoustics (WASPAA)*, pp. 197–200. DOI: 10.1109/ASPAA.2011.6082335. 39

S. Chaudhuri and B. Raj. 2012. Unsupervised structure discovery for semantic analysis of audio. In *Proceedings of Neural Information Processing Systems (NIPS)*, 2:1178–1186. 39, 46, 47

S. Chaudhuri, M. Harvilla, and B. Raj. 2011. Unsupervised learning of acoustic unit descriptors for audio content representation and classification. In *Proceedings of Interspeech*, pp. 2265–2268. 39

S. Chaudhuri, R. Singh, and B. Raj. 2012. Exploiting temporal sequence structure for semantic analysis of multimedia. In *Proceedings of Interspeech*, 2:1726–1729. 39, 46, 47

C. Chen and J.-M. Odobez. 2012. We are not contortionists: Coupled adaptive learning for head and body orientation estimation in surveillance video. In *CVPR*, pp. 1544–1551. DOI: 10.1109/CVPR.2012.6247845. 53, 57, 58, 62, 68, 70, 71

D. L. Chen and W. B. Dolan. 2011. Collecting highly parallel data for paraphrase evaluation. In *Proceedings of the 49th Annual Meeting of the Association for Computational Linguistics (ACL)*, pp. 190–200. 24, 27

K.-T. Chen, Y.-C. Chang, P.-H. Tseng, C.-Y. Huang, and C.-L. Lei. Nov. 2011. Measuring the latency of cloud gaming systems. In *Proceedings of ACM Multimedia 2011*, pp. 1269–1272. DOI: 10.1145/2072298.2071991. 291, 292

K.-T. Chen, C.-Y. Huang, and C.-H. Hsu. 2014. Cloud gaming onward: Research opportunities and outlook. *Proceedings of the 2014 IEEE International Conference on Multimedia and Expo (ICME2014)*, pp. 1–4. DOI: 10.1109/ICMEW.2014.6890683. 290, 291

X. Chen, H. Fang, T.-Y. Lin, R. Vedantam, S. Gupta, P. Dollár, and C. L. Zitnick. 2015. Microsoft COCO captions: Data collection and evaluation server. *arXiv:1504.00325*. 27

Y. Chen, T. Guan, and C. Wang. 2010. Approximate nearest neighbor search by residual vector quantization. *Sensors*, 10(12): 11259–11273. DOI: 10.3390/s101211259. 131

G. Chéron, I. Laptev, and C. Schmid. 2015. P-CNN: Pose-based CNN features for action recognition. In *Proceedings of the IEEE International Conference on Computer Vision*, pp. 3218–3226. *arXiv:1506.03607 [cs.CV]* 53

Y. W. Ching and P. C. Su. 2009. A region of interest rate-control scheme for encoding traffic surveillance videos. *Proceedings of the* $5^{th}$ *International Conference on Intelligent Information Hiding and Multimedia Signal Processing*, pp. 194–197. DOI: 10.1109/IIH-MSP.2009.114. 92

J. Choi, H. Lei, and G. Friedland. September 2011. The 2011 ICSI video location estimation system. *Proceedings of MediaEval 2011*. http://www.icsi.berkeley.edu/pubs/speech/icsivideolocation11.pdf 39

T. Choudhury and A. Pentland. 2003. Sensing and modeling human networks using the sociometer. In *Inter. Sym. on Wearable Comp.*, p. 216. DOI: 10.1109/ISWC.2003.1241414. 53

K.-Y. Chu, Y.-H. Kuo, and W. H. Hsu. 2013. Real-time privacy-preserving moving object detection in the cloud. In *Proceedings of the* $21^{st}$ *ACM International Conference on Multimedia*, pp. 597–600. 94, 96

W.-T. Chu and F.-C. Chang. 2015. A privacy-preserving bipartite graph matching framework for multimedia analysis and retrieval. In *Proceedings of the 5th ACM International Conference on Multimedia Retrieval*, pp. 243–250. ACM. DOI: 10.1145/2671188.2749286. 86, 90

S.-P. Chuah, C. Yuen, and N.-M. Cheung. 2014. Cloud gaming: A green solution to massive multiplayer online games. *IEEE Wireless Communications* (August): 78–87. DOI: 10.1109/MWC.2014.6882299. 290, 291, 308

M. Claypool and K. Claypool. 2006. Latency and player actions in online games. *ACM Communications* 49(11): 40–45. DOI: 10.1145/1167838.1167860. 256, 302

J. Conway and N. Sloane. 1982a. Fast quantizing and decoding algorithms for lattice quantizers and codes. *IEEE Transactions on Information Theory*, 28(2): 227–232. DOI: 10.1109/TIT.1982.1056484. 115

J. Conway and N. Sloane. 1982b. Voronoi regions of lattices, second moments of polytopes, and quantization. *IEEE Transactions on Information Theory*, 28(2): 211–226. DOI: 10.1109/TIT.1982.1056483. 115

J. Conway and N. Sloane. 1990. *Sphere Packings, Lattices and Groups*, third ed. Springer. ISBN:0-387-96617-X 128

M. Cooke, J. Barker, S. Cunningham, and X. Shao. 2006. An audio-visual corpus for speech perception and automatic speech recognition. *The Journal of the Acoustical Society of America*, 120: 2421. DOI: 10.1121/1.2229005. 37

P. T. Costa and R. R. McCrae. 1980. Influence of extraversion and neuroticism on subjective well-being: Happy and unhappy people. *Journal of Personality and Social Psychology*, 38(4): 668. DOI: 10.1037/0022-3514.38.4.668. 244

P. T. J. Costa and R. R. McCrae. 1992. *NEO-PI-R Professional Manual: Revised NEO Personality and NEO Five-Factor Inventory (NEO-FFI)*, volume 4. Psychological Assessment Resources, Odessa, Florida. 237

I. Cox, J. Kilian, T. Leighton, and T. Shamoon. 1996. A secure, robust watermark for multimedia. In *Information Hiding*, volume 1174 of *Lecture Notes in Computer Science*, pp. 185–206. Springer. ISBN:3-540-61996-8. 81

R. Cramer, R. Gennaro, and B. Schoenmakers. 1997. A secure and optimally efficient multi-authority election scheme. *Wiley Online Library, European Transactions on Telecommunications*, 8(5): 481–490. 81

M. Cristani, L. Bazzani, G. Paggetti, A. Fossati, A. Del Blue, G. Menegaz, and V. Murino. 2011. Social interaction discovery by statistical analysis of F-formations. In *British Machine Vision Conference (BMVC)*, 2(4). DOI: 10.5244/C.25.23. 52, 57, 69, 70, 71

P. Cui, F. Wang, S. Liu, M. Ou, S. Yang, and L. Sun. 2011. Who should share what?: Item-level social influence prediction for users and posts ranking. In *Proceedings of the 34th International ACM SIGIR Conference on Research and Development in Information Retrieval*, pp. 185–194. ACM. DOI: 10.1145/2009916.2009945. 148

P. Cui, S.-W. Liu, W.-W. Zhu, H.-B. Luan, T.-S. Chua, and S.-Q. Yang. 2014a. Social-sensed image search. *ACM Transactions on Information Systems (TOIS)*, 32(2): 8. DOI: 10.1145/2590974. 148

P. Cui, Z. Wang, and Z. Su. 2014b. What videos are similar with you?: Learning a common attributed representation for video recommendation. In *Proceedings of the 22nd ACM International Conference on Multimedia*, pp. 597–606. ACM. DOI: 10.1145/2647868.2654946. 140

N. Dalal and B. Triggs. 2005. Histograms of oriented gradients for human detection. In *CVPR*, pp. 886–893. 59

D. J. Daley and D. Vere-Jones. 2003. *An Introduction to the Theory of Point Processes*, 2nd ed. Springer-Verlag, New York. DOI: 10.1007/b97277. 193, 200, 206, 208, 209

I. Damgard and M. Jurik. 2001. A generalisation, a simplification and some applications of Paillier's probabilistic public-key system. In *Proceeedings of 4$^{th}$ International Workshop on Practice and Theory in Public Key Cryptosystems*, volume 1992 of *Lecture Notes in Computer Science*, pp. 119–136. Springer. 81, 96

Q. Danfeng, S. Gammeter, L. Bossard, T. Quack, and L. V. Gool. 2011. Hello neighbor: Accurate object retrieval with k-reciprocal nearest neighbors. In *CVPR*, pp. 777–784. DOI: 10.1109/CVPR.2011.5995373. 106

P. Das, C. Xu, R. F. Doell, and J. J. Corso. 2013. A thousand frames in just a few words: Lingual description of videos through latent topics and sparse object stitching. In *CVPR*, pp. 2634–2641. DOI: 10.1109/CVPR.2013.340. 26

A. Dassios and H. Zhao. 2011. A dynamic contagion process. *Advances in Applied Probability*, pp. 814–846. DOI: 10.1239/aap/1316792671. 200

A. Dassios and H. Zhao. 2013. Exact simulation of Hawkes process with exponentially decaying intensity. *Electronic Communications in Probability*, 18: 1–13. DOI: 10.1214/ECP.v18-2717. 202, 204

M. Datar, N. Immorlica, P. Indyk, and V. Mirrokni. 2004. Locality-sensitive hashing scheme based on p-stable distributions. In *Proceedings of the Symposium on Computational Geometry*, pp. 253–262. DOI: 10.1145/997817.997857. 105, 108, 113, 114, 131

R. Datta, D. Joshi, J. Li, and J. Z. Wang. 2008. Image retrieval: Ideas, influences, and trends of the new age. *ACM Computing Surveys*, 40(2): 5. DOI: 10.1145/1348246.1348248. 84, 86

M. de Berg, M. van Kreveld, M. H. Overmars, and O. Schwarzkopf. March 2008. *Computational Geometry: Algorithms and Applications*, 3rd ed. Springer-Verlag. 109

M. De Choudhury, Y.-R. Lin, H. Sundaram, K. S. Candan, L. Xie, and A. Kelliher. 2010. How does the data sampling strategy impact the discovery of information diffusion in social media? *Proceedings of the 4th International AAAI Conference on Weblogs and Social Media (ICWSM)*, 10: 34–41. 144

F. De la Torre et al. 2009. Guide to the Carnegie Mellon University multimodal activity (CMU-MMAC) database. Technical report. 57

Y.-A. de Montjoye, E. Shmueli, S. S. Wang, and A. S. Pentland. 2014a. openPDS: Protecting the privacy of metadata through SafeAnswers. *PloS ONE*, 9(7): e98790. DOI: 10.1371/journal.pone.0098790. 188

Y.-A. de Montjoye, A. Stopczynski, E. Shmueli, A. Pentland, and S. Lehmann. 2014b. The strength of the strongest ties in collaborative problem solving. *Scientific Reports*, 4:5277. DOI: 10.1038/srep05277. 186

Y.-A. de Montjoye, L. Radaelli, V. K. Singh, and A. S. Pentland. 2015. Unique in the shopping mall: On the reidentifiability of credit card metadata. *Science*, 347(6221): 536–539. DOI: 10.1126/science.1256297. 188

J. Dean and S. Ghemawat. 2008. MapReduce: Simplified data processing on large clusters. *Communications of the ACM*, 51(1): 107–133. DOI: 10.1145/1327452.1327492. 279

A. Delvinioti, H. Jégou, L. Amsaleg, and M. Houle. January 2014. Image retrieval with reciprocal and shared nearest neighbors. In *Proceedings of the International Conference on Computer Vision Theory and Applications (VISAPP)*, pp. 321–328. 106

M. Demirkus, D. Precup, J. J. Clark, and T. Arbel. 2014. Probabilistic temporal head pose estimation using a hierarchical graphical model. In *ECCV*, pp. 328–344. DOI: 10.1007/978-3-319-10590-1_22. 57

J. Deng, W. Dong, R. Socher, L.-J. Li, K. Li, and L. Fei-Fei. 2009. ImageNet: A large-scale hierarchical image database. In *CVPR*. DOI: 10.1109/CVPR.2009.5206848. 9

S. Deterding, D. Dixon, R. Khaled, and L. Nacke. 2011. From game design elements to gamefulness: Defining gamification. In *Proceedings of ACM International Academic MindTrek Conference: Envisioning Future Media Environments (MindTrek)*, pp. 9–15. DOI: 10.1145/2181037.2181040. 261

S. Dey, C. Invited, and P. Paper. 2012. Cloud mobile media: Opportunities, challenges, and directions. In *2012 International Conference on Computing, Networking and Communications, ICNC'12*, pp. 929–933. DOI: 10.1109/ICCNC.2012.6167561. 290, 291

D. Dietrich, W. Kastner, T. Maly, C. Roesener, G. Russ, and H. Schweinzer. 2004. Situation modeling. In *Proceedings of the 2004 IEEE International Workshop on Factory Communication Systems*, pp. 93–102. IEEE. DOI: 10.1109/WFCS.2004.1377687. 165, 167

T. G. Diettereich, R. H. Lathrop, and T. Lozano-Perez. 1998. Solving the multiple instance problem with axis-parallel rectangles. *Artificial Intelligence*, 89:31–71. DOI: 10.1007/3-540-45153-6_20. 48

M. V. Dijk, C. Gentry, S. Halevi, and V. Vaikuntanathan. 2010. Fully homomorphic encryption over the integers. In *Advances in Cryptology—EUROCRYPT*, volume 6110 of *Lecture Notes in Computer Science*, pp. 24–43. Springer. DOI: 10.1007/978-3-642-13190-5_2. 101

T. Dillon, C. Wu, and E. Chang. 2010. Cloud computing: Issues and challenges. In *Proceedings of IEEE International Conference on Advanced Information Networking and Applications (AINA)*, pp. 27–33. DOI: 10.1007/978-3-642-13190-5_2. 259

P. Dimandis. 2012. Abundance is our future. In *TED Conference*. 162

N. Do, C. Hsu, and N. Venkatasubramanian. 2012. CrowdMAC: A crowdsourcing system for mobile access. In *Proceedings of ACM/IFIP/USENIX Middleware*, pp. 1–20. DOI: 10.1007/978-3-642-35170-9_1. 263, 264, 265

N. Do, Y. Zhao, C. Hsu, and N. Venkatasubramanian. 2016. Crowdsourced mobile data transfer with delay bound. *ACM Transactions on Internet Technology (TOIT)*, 16(4): 1–29. DOI: 10.1145/2939376. 263, 264, 265, 267

T. M. T. Do and D. Gatica-Perez. 2013. Human interaction discovery in smartphone proximity networks. *Personal and Ubiquitous Computing*, 17(3): 413–431. DOI: 10.1007/s00779-011-0489-7. 55

C. Dominguez, M. Vidulich, E. Vogel, and G. McMillan. 1994. Situation awareness: Papers and annotated bibliography. Armstrong Laboratory, Human System Center, ref. Technical report, AL/CF-TR-1994-0085. 166, 167

J. Donahue, L. A. Hendricks, M. Rohrbach, S. Venugopalan, S. Guadarrama, K. Saenko, and T. Darrell. 2017. Long-term recurrent convolutional networks for visual recognition and description. In *IEEE Transactions on Pattern Analysis and Machine Intelligence,* 39(4): 677–691 DOI: 10.1109/TPAMI.2016.2599174. 12, 15, 16, 17, 24, 29

W. Dong, M. Charikar, and K. Li. July 2008a. Asymmetric distance estimation with sketches for similarity search in high-dimensional spaces. In *Proceedings of the 31st Annual International ACM SIGIR Conference on Research and Development (SIGIR)*, pp. 123–130. DOI: 10.1145/1390334.1390358. 110, 122, 123

W. Dong, Z. Wang, W. Josephson, M. Charikar, and K. Li. October 2008b. Modeling LSH for performance tuning. In *Proceedings of the 17th Conference on Information and Knowledge Management (CIKM)*, pp. 669–678. DOI: 10.1145/1458082.1458172. 115, 117

W. Dong, M. Charikar, and K. Li. March 2011. Efficient k-nearest neighbor graph construction for generic similarity measures. In *Proceedings of the 20th International Conference on World Wide Web (WWW)*, pp. 577–586. DOI: 10.1145/1963405.1963487. 109, 133

J. J. Dongarra, J. Du Croz, S. Hammarling, and I. S. Duff. 1990. A set of level 3 basic linear algebra subprograms. *ACM Transactions on Mathematical Software (TOMS)*, 16(1): 1–17. DOI: 10.1145/77626.79170. 111

B. Dostal. 2007. Enhancing situational understanding through employment of unmanned aerial vehicle. *Army Transformation Taking Shape: Interim Brigade Combat Team Newsletter*. 166, 167

C. Dousson, P. Gaborit, and M. Ghallab. 1993. Situation recognition: Representation and algorithms. In *International Joint Conference on Artificial Intelligence*, volume 13, pp. 166–166. 166, 167

M. Douze, H. Jégou, H. Singh, L. Amsaleg, and C. Schmid. July 2009. Evaluation of GIST descriptors for web-scale image search. In *Proceedings of the ACM International Conference on Image and Video Retrieval*, pp. 140–147. 132

M. Douze, H. Jégou, and F. Perronnin. October 2016. PCIVRolysemous codes. In *ECCV*, pp. 785–801. DOI: 10.1007/978-3-319-46475-6_48. 109, 134

J. Downie. 2008. The music information retrieval evaluation exchange (2005–2007): A window into music information retrieval research. *Acoustical Science and Technology*, 29(4): 247–255. DOI: 10.1250/ast.29.247. 37

S. Duan, J. Zhang, P. Roe, and M. Towsey. 2014. A survey of tagging techniques for music, speech and environmental sound. *Artificial Intelligence Review*, 42(4): 637–661. DOI: 10.1007/s10462-012-9362-y. 38

L. Dubbeld. 2002. Protecting personal data in camera surveillance practices. *Surveillance & Society*, 2(4). 91

R. Duda and P. Hart. 1996. *Pattern Classification and Scene Analysis*. Wiley. 167

F. Dufaux and T. Ebrahimi. 2004. Video surveillance using JPEG 2000. In *International Society for Optics and Photonics. Optical Science and Technology*, pp. 268–275. DOI: 10.1117/12.564828. 91

F. Dufaux and T. Ebrahimi. 2008. Scrambling for privacy protection in video surveillance systems. *IEEE Transactions on Circuits and Systems for Video Technology*, 18(8): 1168–1174. DOI: 10.1109/TCSVT.2008.928225. 92, 93

N. Eagle and A. Pentland. 2006. Reality mining: Sensing complex social systems. *Personal and Ubiquitous Computing*, 10(4): 255–268. DOI: 10.1007/s00779-005-0046-3. 55, 187

P. Ekman. 1992. An argument for basic emotions. *Cognition and Emotion*, 6(3/4): 169–200. DOI: 10.1080/02699939208411068. 236

A. El Ali, A. Matviienko, Y. Feld, W. Heuten, and S. Boll. 2016. VapeTracker: Tracking vapor consumption to help e-cigarette users quit. In *Proceedings of the 2016 CHI Conference Extended Abstracts on Human Factors in Computing Systems*, pp. 2049–2056. ACM. DOI: 10.1145/2851581.2892318. 188

T. Elgamal. 1985. A public key cryptosystem and a signature scheme based on discrete logarithms. In *Proceedings of the $4^{th}$ Annual International Cryptology Conference*, volume 196 of *Lecture Notes in Computer Science*, pp. 10–18. Springer. DOI: 10.1109/TIT.1985.1057074. 79

T. Elgamal, M. Yabandeh, A. Aboulnaga, W. Mustafa, and M. Hefeeda. 2015. sPCA: Scalable principal component analysis for big data on distributed platforms. In *Proceedings of ACM International Conference on Management of Data (SIGMOD)*, pp. 79–91. DOI: 10.1145/2723372.2751520. 276, 277, 278, 279, 280

P. Elias and L. Roberts. 1963. *Machine Perception of Three-Dimensional Solids*. PhD thesis, Massachusetts Institute of Technology. http://hdl.handle.net/1721.1/11589. 159

B. Elizalde, G. Friedland, H. Lei, and A. Divakaran. October 2012. There is no data like less data: percepts for video concept detection on consumer-produced media. In *Proceedings of ACM Multimedia 2012 (MM'12)*, pp. 27–32. DOI: 10.1145/2390214.2390223. 43

B. Elizalde, A. Kumar, R. Badlani, A. Bhatnagar, A. Shah, R. Singh, B. Raj, and I. Lane, 2016. Never Ending Learning of Sound. http://nels.cs.cmu.edu. 48

D. P. W. Ellis. March 2007. Beat tracking by dynamic programming. *J. New Music Research*, 36(1): 51–60. http://www.ee.columbia.edu/~dpwe/pubs/Ellis07-beattrack.pdf. DOI: 10.1080/09298210701653344. Special Issue on Tempo and Beat Extraction. 38

D. P. W. Ellis and G. Poliner. 2007. Identifying cover songs with chroma features and dynamic programming beat tracking. In *Proceedings of the ICASSP-07*, pp. 1429–1432. Hawai'i. http://www.ee.columbia.edu/~dpwe/pubs/EllisP07-coversongs.pdf. DOI: 10.1109/ICASSP.2007.367348. 38

D. P. W. Ellis, R. Singh, and S. Sivadas. 2001. Tandem acoustic modeling in large-vocabulary recognition. In *(ICASSP*, pp. I–517–520. Salt Lake City. http://www.ee.columbia.edu/~dpwe/pubs/icassp01-spine.pdf. 36

P. Embrechts, T. Liniger, and L. Lin. Aug. 2011. Multivariate Hawkes processes: An application to financial data. *Journal of Applied Probability*, pp. 367–378. DOI: 10.1017/S0021900200099344. 200

M. Endsley. May 1988. Situation awareness global assessment technique (SAGAT). In *Proceedings of the IEEE 1988 National Aerospace and Electronics Conference*, volume 3, pp. 789–795. DOI: 10.1109/NAECON.1988.195097. 165, 167

Z. Erkin, A. Piva, S. Katzenbeisser, R. L. Lagendijk, J. Shokrollahi, G. Neven, and M. Barni, 2006. SPEED project. http://www.speedproject.eu/. 76

Z. Erkin, A. Piva, S. Katzenbeisser, R. L. Lagendijk, J. Shokrollahi, G. Neven, and M. Barni. 2007. Protection and retrieval of encrypted multimedia content: When cryptography meets signal processing. *EURASIP Journal on Information Security*, 2007, article 78943. DOI: 10.1155/2007/78943. 77, 79, 80

Z. Erkin, M. Franz, J. Guajardo, S. Katzenbeisser, I. Lagendijk, and T. Toft. 2009. Privacy-preserving face recognition. In *Proceedings of the* $9^{th}$ *International Symposium, Privacy Enhancing Technologies*, volume 5672 of *Lecture Notes in Computer Science*, pp. 235–253. Springer. DOI: 10.1155/2007/78943. 82, 94, 96

Z. Erkin, M. Beye, T. Veugeri, and R. L. Lagendijk. 2011. Efficiently computing private recommendations. In *(ICASSP)*, pp. 5864–5867. DOI: 10.1504/IJACT.2014.062738. 82

S. Escalera, X. Bar, J. Gonzlez, M. A. Bautista, M. Madadi, M. Reyes, V. Ponce, H. J. Escalante, J. Shotton, and I. Guyon. 2014. ChaLearn Looking at People Challenge 2014: Dataset and results. In *ECCV Workshops*, pp. 459–473. 57

N. Evans, S. Marcel, A. Ross, and A. B. J. Teoh. 2015. Biometrics security and privacy protection [from the guest editors]. *IEEE Signal Processing Magazine*, 32(5): 17–18. 83

M. Everingham, L. Van Gool, C. K. I. Williams, J. Winn, and A. Zisserman. 2012. The PASCAL Visual Object Classes Challenge 2012 (VOC2012) Results. http://www.pascalnetwork.org/challenges/VOC/voc2012/workshop/index.html. 227

M. Everingham, S. M. A. Eslami, L. Gool, C. K. I. Williams, J. Winn, and A. Zisserman. 2014. The Pascal Visual Object Classes Challenge: A retrospective. *International Journal of Computer Vision (IJCV)*, 111(1): 98–136. 220, 226

H. J. Eysenck. 1947. *Dimensions of Personality*. The International Library of Psychology. Transaction Publishers.

R. Fagin. June 1998. Fuzzy queries in multimedia database systems. In *Proceedings of the ACM Symposium on Principles of Database Systems*, pp. 1–10. DOI: 10.1145/275487.275488. 105

R. Fagin, R. Kumar, and D. Sivakumar. 2003. Efficient similarity search and classification via rank aggregation. In *SIGMOD*, pp. 301–312. DOI: 10.1145/872757.872795. 114

C.-I. Fan, S.-Y. Huang, and W.-C. Hsu. 2015. Encrypted data deduplication in cloud storage. In *2015 10th Asia Joint Conference on Information Security (AsiaJCIS)*, pp. 18–25. IEEE. DOI: 10.1109/AsiaJCIS.2015.12. 94

Farach-Colton and P. Indyk. October 1999. Approximate nearest neighbor algorithms for Hausdorff metrics via embeddings. In *Proceedings of the Symposium on the Foundations of Computer Science*, pp. 171–179. DOI: 10.1109/SFFCS.1999.814589. 133

C. Feichtenhofer, A. Pinz, and A. Zisserman. 2016. Convolutional two-stream network fusion for video action recognition. In *CVPR*, pp. 1933–1941. 11, 24

P. Flajolet and G. N. Martin. October 1985. Probabilistic counting algorithms for data base applications. *Journal of Computer and System Sciences*, 31(2): 182–209. DOI: 10.1016/0022-0000(85)90041-8. 120

M. Flickner, S. Harpreet, W. Niblack, J. Ashley, Q. Huang, B. Dom, M. Gorkani, J. Hafner, D. Lee, D. Petkovic, D. Steele, and P. Yanker. September 1995. Query by image and video content: The QBIC system. *Computer*, 28: 23–32. DOI: 10.1109/2.410146. 105

C. Fontaine and F. Galand. 2007. A survey of homomorphic encryption for nonspecialists. *Hindawi Publishing Corporation, EURASIP Journal on Information Security*, 2007, pp. 1–10. DOI: 10.1155/2007/13801. 77

J. Fortmann, E. Root, S. Boll, and W. Heuten. 2016. Tangible apps bracelet: Designing modular wrist-worn digital jewellery for multiple purposes. In *Proceedings of the 2016 ACM Conference on Designing Interactive Systems*, DIS '16, pp. 841–852. ACM, New York. http://doi.acm.org/10.1145/2901790.2901838. DOI: 10.1145/2901790.2901838. 188

T. Foulsham, R. Dewhurst, M. Nyström, H. Jarodzka, R. Johansson, G. Underwood, and K. Holmqvist. 2012. Comparing scanpaths during scene encoding and recognition: A multi-dimensional approach. *Journal of Eye Movement Research*, 5(4:3): 1–14. DOI: 10.16910/jemr.5.4.3. 229

M. Franz and S. Katzenbeisser. 2011. Processing encrypted floating point signals. In *Proceedings of the Thirteenth ACM Multimedia Workshop on Multimedia and Security*, pp. 103–108. ACM. DOI: 10.1145/2037252.2037271. 102

M. Franz, B. Deiseroth, K. Hamacher, S. Katzenbeisser, S. Jha, and H. Schroeder. 2010. Secure computations on real-valued signals. In *Proceedings of the IEEE Workshop on Information Forensics and Security (WIFS)*, pp. 25–27. DOI: 10.1109/WIFS.2010.5711458. 102

G. Friedland and D. van Leeuwen. 2010. Speaker recognition and diarization. In *Semantic Computing*, pp. 115–130. IEEE Press/Wileys. DOI: 10.1002/9780470588222.ch7. 36

J. H. Friedman, J. L. Bentley, and R. A. Finkel. 1977. An algorithm for finding best matches in logarithmic expected time. *ACM Transactions on Mathematical Software*, 3(3): 209–226. DOI: 10.1145/355744.355745. 107

J.-J. Fuchs. November 2011. Spread representations. In *ASILOMAR*. DOI: 10.1109/ACSSC.2011.6190120. 126

T. Fujishima. 1999. Realtime chord recognition of musical sound: A system using Common Lisp music. In *Proceedings International Computer Music Conference*, pp. 464–467. http://www-ccrma.stanford.edu/~jos/mus423h/Real_Time_Chord_Recognition_Musical.html. 37

K. Fukushima. 1980. Neocognitron: A self-organizing neural network model for a mechanism of pattern recognition unaffected by shift in position. *Biological Cybernetics*, 36(4): 193–202. 5

T. Furon and H. Jégou. February 2013. Using extreme value theory for image detection. Research Report RR-8244, INRIA. 106

T. Furon, H. Jégou, L. Amsaleg, and B. Mathon. November 2013. Fast and secure similarity search in high dimensional space. In *WIFS*, pp. 73–78. DOI: 10.1109/WIFS.2013.6707797. 105

GaiKai. January 2015. GaiKai web page. http://www.gaikai.com/. 287, 291

GamingAnywhere Repository. 2013. GamingAnywhere: An open source cloud gaming project. http://gaminganywhere.org. 292, 294, 296

R. Ganti, F. Ye, and H. Lei. 2011. Mobile crowdsensing: Current state and future challenges. *IEEE Communication Magazine*, 49(11): 32–39. DOI: 10.1109/MCOM.2011.6069707. 268

M. Gao. 2012. *EventShop: A scalable framework for analysis of spatio-temporal-thematic data streams*. PhD thesis, University of California, Irvine. 183

M. Gao, V. K. Singh, and R. Jain. 2012. EventShop: From heterogeneous web streams to personalized situation detection and control. In *Proceedings of the 4th Annual ACM Web Science Conference*, pp. 105–108. ACM. DOI: 10.1145/2380718.2380733. 161, 177, 185

L. P. García-Perera, J. A. Nolazco-Flores, B. Raj, and R. M. Stern. 2012. Optimization of the DET curve in speaker verification. In *2012 IEEE Spoken Language Technology Workshop (SLT), Miami, FL, USA, December 2–5, 2012*, pp. 318–323. DOI: 10.1109/SLT.2012.6424243. 36

L. P. García-Perera, B. Raj, and J. A. Nolazco-Flores. 2013a. Ensemble approach in speaker verification. In *INTERSPEECH 2013, 14th Annual Conference of the International Speech Communication Association, Lyon, France, August 25–29, 2013*, pp. 2455–2459. 36

L. P. García-Perera, B. Raj, and J. A. Nolazco-Flores. 2013b. Optimization of the DET curve in speaker verification under noisy conditions. In *IEEE International Conference on Acoustics, Speech and Signal Processing, ICASSP 2013, Vancouver, BC, Canada, May 26–31, 2013*, pp. 7765–7769. 36

D. Gatica-Perez. 2009. Automatic nonverbal analysis of social interaction in small groups: A review. *Image and Vision Computing*, 27(12): 1775–1787. DOI: 10.1016/j.imavis.2009.01.004. 55

T. Ge, K. He, Q. Ke, and J. Sun. June 2013. Optimized product quantization for approximate nearest neighbor search. In *CVPR*, pp. 744–755. DOI: 10.1109/TPAMI.2013.240. 131

I.-D. Gebru, X. Alameda-Pineda, F. Forbes, and R. Horaud. 2016. EM algorithms for weighted-data clustering with application to audio-visual scene analysis. *IEEE TPAMI*, 38(12): 2402–2415. DOI: 10.1109/TPAMI.2016.2522425. 55

H. W. Gellersen, A. Schmidt, and M. Beigl. 2002. Multi-sensor context-awareness in mobile devices and smart artifacts. *Mobile Networks and Applications*, 7(5): 341–351. DOI: 10.1023/A:1016587515822. 187

X. Geng and Y. Xia. 2014. Head pose estimation based on multivariate label distribution. In *CVPR*. DOI: 10.1109/CVPR.2014.237. 57

C. Gentry. 2009. *A fully homomorphic encryption scheme*. PhD thesis, Stanford University. 77, 101

L. Georgiadis, M. Neely, and L. Tassiulas. 2006. Resource allocation and cross-layer control in wireless networks. *Foundations and Trends in Networking*. DOI: 10.1561/1300000001. 265

I. Ghosh and V. Singh. 2016. Predicting privacy attitudes using phone metadata. In *Proceedings of the International Conference on Social Computing, Behavioral-Cultural Modeling & Prediction and Behavior Representation in Modeling and Simulation, Washington DC*, volume 1, pp. 51–60. DOI: 10.1007/978-3-319-39931-7_6. 188

D. Giannoulis, E. Benetos, D. Stowell, M. Rossignol, M. Lagrange, and M. Plumbley, 2013. IEEE AASP challenge: Detection and classification of acoustic scenes and events. Web resource, available: http://c4dm.eecs.qmul.ac.uk/sceneseventschallenge/. DOI: 10.1109/WASPAA.2013.6701819. 38

S. O. Gilani, R. Subramanian, Y. Yan, D. Melcher, N. Sebe, and S. Winkler. 2015. PET: An eye-tracking dataset for animal-centric Pascal object classes. In *International Conference on Multimedia & Expo (ICME)*. DOI: 10.1109/ICME.2015.7177450. 221, 226, 230, 233, 234, 236

A. Gionis, P. Indyk, and R. Motwani. 1999. Similarity search in high dimension via hashing. In *Proceedings of the International Conference on Very Large Databases*, pp. 518–529. 105, 113, 132

M. Girolami, S. Lenzi, F. Furfari, and S. Chessa. 2008. SAIL: A sensor abstraction and integration layer for context awareness. In *2008 34th Euromicro Conference Software Engineering and Advanced Applications*, pp. 374–381. IEEE. DOI: 10.1109/SEAA.2008.30. 187

R. Girshick. 2015. Fast R-CNN. In *ICCV*, pp. 1440–1448. DOI: 10.1109/ICCV.2015.169. 3, 9

R. Girshick, J. Donahue, T. Darrell, and J. Malik. 2014. Rich feature hierarchies for accurate object detection and semantic segmentation. In *CVPR*, pp. 580–587. DOI: 10.1109/CVPR.2014.81. 9

A. Goldberg, B. Recht, J. Xu, R. Nowak, and X. Zhu. 2010. Transduction with matrix completion: Three birds with one stone. In *NIPS*, pp. 757–765. 57, 60, 62, 68

D. Goldman. Apr. 2012. Google unveils 'Project Glass' virtual-reality glasses. In *Money*. DOI: 10.1109/NWeSP.2011.6088206. 313

O. Goldreich, S. Micali, and A. Wigderson. 1987. How to play any mental game. In *Proceedings of the Nineteenth Annual ACM Symposium on Theory of Computing*, pp. 218–229. ACM. DOI: 10.1145/28395.28420. 80

S. Goldwasser and S. Micali. 1984. Probabilistic encryption. *Elsevier Journal of Computer and System Sciences*, 28(2): 270–299. DOI: 10.1016/0022-0000(84)9 0070-9. 80

B. Golub and M. O. Jackson. 2010. Using selection bias to explain the observed structure of internet diffusions. *Proceedings of the National Academy of Sciences*, 107(24): 10833–10836. DOI: 10.1073/pnas.1000814107. 145

Y. Gong and S. Lazebnik. June 2011. Iterative quantization: A procrustean approach to learning binary codes. In *CVPR*, pp. 817–824. DOI: 10.1109/CVPR.2011.5995432. 122, 124

Z. Gong, X. Gu, and J. Wilkes. 2010. PRESS: Predictive elastic resource scaling for cloud systems. In *Proceedings of IEEE International Conference on Network and Service Management (CNSM)*, pp. 9–16. DOI: 10.1109/CNSM.2010.5691343. 259

M. C. Gonzalez, C. A. Hidalgo, and A.-L. Barabasi. 2008. Understanding individual human mobility patterns. *Nature*, 453(7196): 779–782. DOI: 10.1038/nature06958. 187

I. J. Goodfellow, D. Warde-Farley, M. Mirza, A. C. Courville, and Y. Bengio. 2013. Maxout networks. In *Proceedings of the International Conference on Machine Learning (ICML)*, pp. 1319–1327. 6

D. Gorisse, M. Cord, and F. Precioso. February 2012. Locality-sensitive hashing for CHI2 distance. *IEEE Trans. PAMI*, 34(2): 402–409. DOI: 10.1109/TPAMI.2011.193. 133

G. J. Gorn. 1982. The effects of music in advertising on choice behavior: A classical conditioning approach. *The Journal of Marketing*, pp. 94–101. DOI: 10.2307/1251163. 187

M. Goto. 2001. A predominant-F0 estimation method for CD recordings: Map estimation using EM algorithm for adaptive tone models. In *Proceedings ICASSP-2001*, pp. 3365–3368. 37

V. K. Goyal, M. Vetterli, and N. T. Thao. January 1998. Quantized overcomplete expansions in $\mathcal{R}^N$: Analysis, synthesis, and algorithms. *IEEE Trans. Inform. Theory*, 44(1): 16–31. DOI: 10.1109/18.650985. 121, 124

K. Graffi, D. Stingl, C. Gross, H. Nguyen, A. Kovacevic, and R. Steinmetz. 2010. Towards a P2P cloud: Reliable resource reservations in unreliable P2P systems. In *Proceedings of IEEE International Conference on Parallel and Distributed Systems (ICPADS)*, pp. 27–34. DOI: 10.1109/ICPADS.2010.34. 259

A. Graves. 2012. *Supervised Sequence Labelling with Recurrent Neural Networks*. Springer. 6

A. Graves, A. Mohamed, and G. E. Hinton. 2013. Speech recognition with deep recurrent neural networks. In *ICASSP*, pp. 6645–6649. DOI: 10.1109/ICASSP.2013.6638947. 3, 12

R. M. Gray and D. L. Neuhoff. October 1998. Quantization. *IEEE Transactions on Information Theory*, 44: 2325–2384. DOI: 10.1109/18.720541. 117, 127, 128

J. Gu, Z. Wang, J. Kuen, L. Ma, A. Shahroudy, B. Shuai, T. Liu, X. Wang, and G. Wang. 2016. Recent advances in convolutional neural networks. In *arXiv:1512.07108*. 6

S. Guadarrama, N. Krishnamoorthy, G. Malkarnenkar, S. Venugopalan, R. Mooney, T. Darrell, and K. Saenko. 2013. YouTube2Text: Recognizing and describing arbitrary activities using semantic hierarchies and zero-shot recognition. In *ICCV*, pp. 2712–2719. DOI: 10.1109/ICCV.2013.337. 16, 26

J. Guo, C. Gurrin, and S. Lao. 2013. Who produced this video, amateur or professional? In *Proceedings of the 3rd ACM International Conference on Multimedia Retrieval*, pp. 271–278. ACM. DOI: 10.1145/2461466.2461509. 147

S. Gupta and J. Nicholson. 1985. Simple visual reaction time, personality strength of the nervous system: Theory approach. *Personality and Individual Differences*, 6(4): 461–469. DOI: 10.1016/0191-8869(85)90139-4. 244

A. Guttman. 1984. R-trees: A dynamic index structure for spatial searching. In *SIGMOD*, pp. 47–57. DOI: 10.1145/602259.602266. 107

H. Haddadi, H. Howard, A. Chaudhry, J. Crowcroft, A. Madhavapeddy, and R. Mortier. 2015. Personal data: Thinking inside the box. In *arXiv:1501.04737*. 188

A. Hanjalic. 2013. Multimedia retrieval that matters. *ACM Transactions on Multimedia Computing, Communications, and Applications (TOMCCAP)*, 9(1s): 44. DOI: 10.1145/2490827. 155

A. Hanjalic and L.-Q. Xu. 2005. Affective video content representation and modeling. *IEEE Transcations on Multimedia*, 7(1): 143–154. DOI: 10.1109/TMM.2004.840618. 221

A. Hanjalic, C. Kofler, and M. Larson. 2012. Intent and its discontents: The user at the wheel of the online video search engine. In *Proceedings of the 20th ACM International Conference on Multimedia*, pp. 1239–1248. ACM. DOI: 10.1145/2393347.2396424. 140

D. A. Harville. 1998. *Matrix Algebra from a Statistician's Perspective*. Springer-verlag. 185

A. G. Hawkes. 1971. Spectra of some self-exciting and mutually exciting point processes. *Biometrika*, pp. 83–90. DOI: 10.1093/biomet/58.1.83. 193, 197, 198, 200

A. G. Hawkes and D. Oakes. 1974. A cluster process representation of a self-exciting process. *Journal of Applied Probability*, 11: 493–503. DOI: 10.2307/3212693. 200

S. Haynes and R. Jain. 1986. Event detection and correspondence. *Optical Engineering*, 25: 387–393. DOI: 10.1117/12.7973835. 159

K. He, F. Wen, and J. Sun. June 2013. K-means hashing: An affinity-preserving quantization method for learning binary compact codes. In *CVPR*, pp. 2938–2945. DOI: 10.1109/CVPR.2013.378. 121

K. He, X. Zhang, S. Ren, and J. Sun. 2016a. Identity mappings in deep residual networks. In *ECCV*, pp. 630–645. DOI: 10.1007/978-3-319-46493-0_38. 6

K. He, X. Zhang, S. Ren, and J. Sun. 2016b. Deep residual learning for image recognition. In *CVPR*, pp. 770–778. DOI: 10.1109/cvpr.2016.90. 5, 6, 9

L. He, G. Liu, and C. Yuchen. Jul. 2014. Buffer status and content aware scheduling scheme for cloud gaming based on video streaming. In *2014 IEEE International Conference on Multimedia and Expo Workshops (ICME)*, pp. 1–6. DOI: 10.1109/ICMEW.2014.6890629. 309

F. C. Heilbron, V. Escorcia, B. Ghanem, and J. C. Niebles. 2015. ActivityNet: A large-scale video benchmark for human activity understanding. In *CVPR*, pp. 1914–1923. 19, 22, 23

M. Hemmati, A. Javadtalab, A. A. Nazari Shirehjini, S. Shirmohammadi, and T. Arici. 2013. Game as video: Bit rate reduction through adaptive object encoding. In *Proceedings of the 23rd ACM Workshop on Network and Operating Systems Support for Digital Audio and Video*, NOSSDAV '13, pp. 7–12. ACM, New York. DOI: 10.1145/2460782.2460784. 310

J. Hershey, S. Rennie, P. Olsen, and T. Kristjansson. 2010. Super-human multi-talker speech recognition: A graphical modeling approach. *Computer Speech & Language*, 24(1): 45–66. DOI: 10.1016/j.csl.2008.11.001. 37

G. Higgins. December 14, 1993. System for distributing, processing and displaying financial information. US Patent 5,270,922. 165

G. Hinton, L. Deng, D. Yu, G. E. Dahl, A.-R. Mohamed, N. Jaitly, A. Senior, V. Vanhoucke, P. Nguyen, T. N. Sainath, et al. 2012. Deep neural networks for acoustic modeling in speech recognition: The shared views of four research groups. *IEEE Signal Processing Magazine*, pp. 82–97. DOI: 10.1109/MSP.2012.2205597. 3

G. E. Hinton, S. Osindero, and Y.-W. Teh. 2006. A fast learning algorithm for deep belief nets. *Neural Computation*, 18(7): 1527–1554. DOI: 10.1162/neco.2006.18.7.1527.

M. Hirt and K. Sako. 2000. Efficient receipt-free voting based on homomorphic encryption. In *Advances in Cryptology*, volume 1807 of *Lecture Notes in Computer Science*, pp. 539–556. Springer. DOI: 10.1007/3-540-45539-6_38. 81

G. Hjaltason and H. Samet. 1998. Incremental distance join algorithms for spatial databases. In *ACM SIGMOD Record*, volume 27, pp. 237–248. ACM. DOI: 10.1145/276305.276326. 173

S. Hochreiter and J. Schmidhuber. 1997. Long short-term memory. *Neural Computation*, 9(8): 1735–1780. DOI: 10.1162/neco.1997.9.8.1735. 16

S. C. Hoi, W. Liu, and S.-F. Chang. 2008. Semi-supervised distance metric learning for collaborative image retrieval. In *CVPR*, pp. 1–7. DOI: 10.1145/1823746.1823752. 149

H. Hong, D. Chen, C. Huang, K. Chen, and C. Hsu. 2015. Placing virtual machine to optimize cloud gaming experience. *IEEE Transactions on Cloud Computing*, 3(1): 42–53. DOI: 10.1109/TCC.2014.2338295. 259

H. Hong, J. Chuang, and C. Hsu. 2016a. Animation rendering on multimedia fog computing platforms. In *Proceedings of IEEE International Conference on Cloud Computing Technology and Science (CloudCom)*, pp. 336–343. DOI: 10.1109/CloudCom.2016.0060. 274, 275

H. Hong, P. Tsai, and C. Hsu. 2016b. Dynamic module deployment in a fog computing platform. In *Proceedings of the Asia-Pacific Network Operations and Management Symposium (APNOMS)*, pp. 1–6. DOI: 10.1109/APNOMS.2016.7737202. 282, 283, 284

H.-J. Hong, D.-Y. Chen, C.-Y. Huang, K.-T. Chen, and C.-H. Hsu. 2014a. Placing virtual machines to optimize cloud gaming experience. *IEEE Transactions on Cloud Computing*, pp. 1–2. http://ieeexplore.ieee.org/lpdocs/epic03/wrapper.htm?arnumber=6853364. DOI: 10.1109/TCC.2014.2338295. 296

H.-J. Hong, T.-Y. Fan-Chiang, C.-R. Lee, K.-T. Chen, C.-Y. Huang, and C.-H. Hsu. 2014b. GPU consolidation for cloud games: Are we there yet? *2014 13th Annual Workshop on Network and Systems Support for Games (NetGames)*, article 3. DOI: 10.1109/NetGames.2014.7008969. 297, 302

C.-F. Hsu, T.-H. Tsai, C.-Y. Huang, C.-H. Hsu, and K.-T. Chen. Mar 2015. Screencast dissected: Performance measurements and design considerations. In *Proceedings of ACM Multimedia Systems 2015*, pp. 177–188. DOI: 10.1145/2713168.2713176. 297

C. Y. Hsu, C. S. Lu, and S. C. Pei. 2009. Secure and robust SIFT. In *Proceedings of the* $17^{th}$ *ACM International Conference on Multimedia*, pp. 637–640. DOI: 10.1145/1631272.1631376. 86, 87

C. Y. Hsu, C. S. Lu, and S. C. Pei. 2011. Homomorphic encryption-based secure SIFT for privacy-preserving feature extraction. *Proceedings of the International Society for Optics and Photonics, IS&T/SPIE Electronic Imaging*, 7880(788005). DOI: 10.1117/12.873325. 86, 87, 90

G. Hu and D. L. Wang. 2003. Monaural speech separation. In *NIPS*, volume 13. MIT Press, Cambridge MA. 37

X.-S. Hua, L. Yang, J. Wang, J. Wang, M. Ye, K. Wang, Y. Rui, and J. Li. 2013. Clickage: Towards bridging semantic and intent gaps via mining click logs of search engines. In *Proceedings of the 21st ACM International Conference on Multimedia*, pp. 243–252. ACM. DOI: 10.1145/2502081.2502283. 148

C.-Y. Huang, D.-Y. Chen, C.-H. Hsu, and K.-T. Chen. Oct. 2013a. GamingAnywhere: An open-source cloud gaming testbed. In *Proceedings of ACM Multimedia 2013 (Open Source Software Competition Track)*, pp. 36–47. DOI: 10.1145/2502081.2502222. 297

C.-Y. Huang, C.-H. Hsu, D.-Y. Chen, and K.-T. Chen. 2013b. Quantifying user satisfaction in mobile cloud games. In *Proceedings of Workshop on Mobile Video Delivery*, MoViD'14, pp. 4:1–6. ACM, New York. DOI: 10.1145/2579465.2579468. 297

C.-Y. Huang, K.-T. Chen, D.-Y. Chen, H.-J. Hsu, and C.-H. Hsu. Jan. 2014a. GamingAnywhere: The first open source cloud gaming system. *ACM Transactions on Multimedia Computer Communication Applications*, 10(1s): 10:1–25. DOI: 10.1145/2537855. 292, 293

C.-Y. Huang, P.-H. Chen, Y.-L. Huang, K.-T. Chen, and C.-H. Hsu. 2014b. Measuring the client performance and energy consumption in mobile cloud gaming. *2014 13th Annual Workshop on Network and Systems Support for Games (NetGames)*, pp. 4–6. 297

R. W. Hubbard, S. Magotiaux, and M. Sullivan. 2004. The state use of closed circuit TV: Is there a reasonable expectation of privacy in public? *Crim. LQ*, 49: 222. 91

D. H. Hubel and T. N. Wiesel. 1968. Receptive fields and functional architecture of monkey striate cortex. *The Journal of Physiology*, pp. 215–243. DOI: 10.1113/jphysiol.1968.sp008455. 5

H. Hung and B. Kröse. 2011. Detecting F-formations as dominant sets. In *Proceedings of the 13th International Conference on Multimodal Interfaces (ICMI)*, pp. 231–238. DOI: 10.1145/2070481.2070525. 70

ICSI, U. Berkeley, Yahoo, and LLNL. 2015. The Multimedia Commons Project. http://mmcommons.org. 39

H. Idrees, A. R. Zamir, Y.-G. Jiang, A. Gorban, I. Laptev, R. Sukthankar, and M. Shah. 2016. The THUMOS challenge on action recognition for videos "in the wild." In *arXiv:1604.06182*. DOI: 10.1016/j.cviu.2016.10.018. 23

D. Imseng and G. Friedland. 2010. Tuning-robust initialization methods for speaker diarization. *Transactions on Audio, Speech, and Language Processing*, 18(8): 2028–2037. DOI: 10.1109/TASL.2010.2040796. 36

P. Indyk. 2002. Approximate nearest neighbor algorithms for Frechet distance via product metrics. In *Proceedings of the Eighteenth Annual Symposium on Computational Geometry*, pp. 102–106. DOI: 10.1145/513400.513414. 133

P. Indyk and R. Motwani. 1998. Approximate nearest neighbors: Towards removing the curse of dimensionality. In *STOC*, pp. 604–613. DOI: 10.1145/276698.276876. 105, 107, 108, 114

P. Indyk and A. Naor. Aug. 2007. Nearest-neighbor-preserving embeddings. *ACM Transactions on Algorithms (TALG)*, 3: article 31. DOI: 10.1145/1273340.1273347. 133

P. Indyk and N. Thaper. Oct. 2003. Fast image retrieval via embeddings. In *International Workshop on Statistical and Computational Theories of Vision, ICCV Workshop*. 120

Infinity Research. Dec. 2016. Global Cloud Gaming Market 2017–2021. http://www.technavio.com/report/global-gaming-global-cloud-gaming-market-2017-2021. 287

S. Ioffe and C. Szegedy. 2015. Batch normalization: Accelerating deep network training by reducing internal covariate shift. In *ICML*, pp. 448–456. 6

R. Iqbal, S. Shirmohammadi, and A. E. Saddik. 2006. Compressed-domain encryption of adapted H.264 video. *Proceedings of IEEE* $8^{th}$ *International Symposium on Multimedia*, pp. 979–984. DOI: 10.1109/ISM.2006.50. 92, 93

N. Islam, W. Puech, and R. Brouzet. 2009. A homomorphic method for sharing secret images. In *Proceedings of the* $8^{th}$ *International Workshop on Digital Watermarking*, volume 5703 of *Lecture Notes in Computer Science*, pp. 121–135. Springer. DOI: 10.1007/978-3-642-03688-0_13. 88

Y. Isoda, S. Kurakake, and H. Nakano. 2004. Ubiquitous sensors based human behavior modeling and recognition using a spatio-temporal representation of user states. In *18th International Conference on Advanced Information Networking and Applications*, volume 1, pp. 512–517. IEEE. DOI: 10.1109/AINA.2004.1283961. 187

L. Itti and C. Koch. 2000. A saliency-based search mechanism for overt and covert shifts of visual attention. *Vision Research*, 40(10): 1489–1506. DOI: 10.1016/S0042-6989(99)00163-7. 222

L. Itti, C. Koch, and E. Niebur. 1998. A model of saliency-based visual attention for rapid scene analysis. *IEEE Trans. PAMI*, 20(11): 1254–1259. DOI: 10.1109/34.730558. 231

H. Jain, P. Pérez, R. Gribonval, J. Zepeda, and H. Jégou. October 2016. Approximate search with quantized sparse representations. In *ECCV*, pp. 681–696. DOI: 10.1007/978-3-319-46478-7_42. 131

R. Jain and D. Sonnen. 2011. Social life networks. *IT Professional*, 13(5): 8–11. DOI: 10.1109/MITP.2011.86. 137

G. Jakobson, J. Buford, and L. Lewis. Oct. 2006. A framework of cognitive situation modeling and recognition. In *Military Communications Conference, 2006. IEEE*, pp. 1–7. DOI: 10.1109/MILCOM.2006.302076. 166

M. Jamali and M. Ester. 2009. TrustWalker: A random walk model for combining trust-based and item-based recommendation. In *Proceedings of the 15th ACM SIGKDD International Conference on Knowledge Discovery and Data Mining*, pp. 397–406. ACM. DOI: 10.1145/1557019.1557067. 152

B. J. Jansen. 2006. Search log analysis: What it is, what's been done, how to do it. *Library & Information Science Research*, 28(3): 407–432. DOI: 10.1016/j.lisr.2006.06.005. 148

S. Jarvinen, J.-P. Laulajainen, T. Sutinen, and S. Sallinen. 2006. QoS-aware real-time video encoding how to improve the user experience of a gaming-on-demand service. *2006 3rd IEEE Consumer Communications and Networking Conference*, 2: 994–997. http://ieeexplore.ieee.org/lpdocs/epic03/wrapper.htm?arnumber=1593187. DOI 10.1109/CCNC.2006.1593187. 308

E. Jeannot, C. Kelly, and D. Thompson. 2003. The development of situation awareness measures in ATM systems. EATMP Report HRS/HSP-005-REP-01. 165, 167

H. Jégou, H. Harzallah, and C. Schmid. June 2007. A contextual dissimilarity measure for accurate and efficient image search. In *CVPR*, pp. 1–8. DOI: 10.1109/CVPR.2007.382970. 119

H. Jégou, L. Amsaleg, C. Schmid, and P. Gros. April 2008a. Query-adaptive locality sensitive hashing. In *ICASSP*, pp. 825–828. DOI: 10.1109/ICASSP.2008.4517737. 115, 117, 119

H. Jégou, M. Douze, and C. Schmid. October 2008b. Hamming embedding and weak geometric consistency for large scale image search. In *ECCV*, pp. 304–317. DOI: 10.1007/978-3-540-88682-2_24. 120, 123, 132

H. Jégou, M. Douze, and C. Schmid. February 2010a. Improving bag-of-features for large scale image search. *IJCV*, 87(3): 316–336. DOI: 10.1007/s11263-009-0285-2. 116, 119

H. Jégou, M. Douze, C. Schmid, and P. Pérez. June 2010b. Aggregating local descriptors into a compact image representation. In *CVPR*, pp. 3304–3311. DOI: 10.1109/CVPR.2010.5540039. 9, 124, 131, 132

H. Jégou, C. Schmid, H. Harzallah, and J. Verbeek. January 2010. Accurate image search using the contextual dissimilarity measure. *IEEE Trans. PAMI*, 32(1): 2–11. DOI: 10.1109/TPAMI.2008.285. 106

H. Jégou, M. Douze, and C. Schmid. January 2011. Product quantization for nearest neighbor search. *IEEE Trans. PAMI*, 33(1): 117–128. DOI: 10.1109/TPAMI.2010.57. 106, 110, 112, 123, 127, 128, 129, 130, 131

H. Jégou, R. Tavenard, M. Douze, and L. Amsaleg. May 2011. Searching in one billion vectors: Re-rank with source coding. In *ICASSP*, pp. 861–864. DOI: 10.1109/ICASSP.2011.5946540. 131

H. Jégou, T. Furon, and J.-J. Fuchs. January 2012. Anti-sparse coding for approximate nearest neighbor search. In *ICASSP*, pp. 2029–2032. DOI: 10.1109/ICASSP.2012.6288307. 123, 124, 127

J. H. Jensen, M. G. Christensen, D. P. W. Ellis, and S. H. Jensen. May 2009. Quantitative analysis of a common audio similarity measure. *IEEE Transactions on Audio, Speech, and Language Processing*, 17(4): 693–703. http://ieeexplore.ieee.org/stamp/stamp.jsp?tp=&arnumber=4804953&isnumber=4787206. DOI: 10.1109/TASL.2008.2012314. 38

J. Ji, J. Li, S. Yan, B. Zhang, and Q. Tian. Dec. 2012. Super-bit locality-sensitive hashing. In *NIPS*, pp. 108–116. 124

S. Ji, W. Xu, M. Yang, and K. Yu. 2010. 3D convolutional neural networks for human action recognition. In *ICML*, pp. 221–231. DOI: 10.1109/TPAMI.2012.59. 9, 10

M. Jiang, P. Cui, R. Liu, Q. Yang, F. Wang, W. Zhu, and S. Yang. 2012a. Social contextual recommendation. In *Proceedings of the 21st ACM International Conference on Information and Knowledge Management*, pp. 45–54. ACM. DOI: 10.1145/2396761.2396771. 147, 152

M. Jiang, P. Cui, F. Wang, Q. Yang, W. Zhu, and S. Yang. 2012b. Social recommendation across multiple relational domains. In *Proceedings of the 21st ACM International Conference on Information and Knowledge Management*, pp. 1422–1431. ACM. DOI: 10.1145/2396761.2398448. 152

M. Jiang, P. Cui, F. Wang, W. Zhu, and S. Yang. 2014a. Scalable recommendation with social contextual information. *IEEE Transactions on Knowledge and Data Engineering*, 26(11): 2789–2802. DOI: 10.1109/TKDE.2014.2300487. 152

Y.-G. Jiang, G. Ye, S.-F. Chang, D. Ellis, and A. C. Loui. 2011. Consumer video understanding: A benchmark database and an evaluation of human and machine performance. In *ACM International Conference on Multimedia Retrieval (ICMR)*, p. 29. DOI: 10.1145/1991996.1992025. 21

Y.-G. Jiang, S. Bhattacharya, S.-F. Chang, and M. Shah. 2013. High-level event recognition in unconstrained videos. *International Journal of Multimedia Information Retrieval (IJMIR)*, pp. 73–101. DOI: 10.1007/s13735-012-0024-2. 4

Y.-G. Jiang, J. Liu, A. Roshan Zamir, G. Toderici, I. Laptev, M. Shah, and R. Sukthankar. 2014b. THUMOS challenge: Action recognition with a large number of classes. http://crcv.ucf.edu/THUMOS14/. 19, 21

Y.-G. Jiang, Z. Wu, J. Wang, X. Xue, and S.-F. Chang. 2015. Exploiting feature and class relationships in video categorization with regularized deep neural networks. *arXiv:1502.07209*. 19, 22

X. Jin, A. Gallagher, L. Cao, J. Luo, and J. Han. 2010. The wisdom of social multimedia: Using Flickr for prediction and forecast. In *Proceedings of the International Conference on Multimedia*, pp. 1235–1244. ACM. DOI: 10.1145/1873951.1874196. 156

Z. Jin, Y. Hu, Y. Lin, D. Zhang, S. Lin, D. Cai, and X. Li. Dec. 2013. Complementary projection hashing. In *ICCV*, pp. 257–264. DOI: 10.1109/ICCV.2013.39. 118

T. Joachims. 1999. Transductive inference for text classification using SVM. In *ICML*, pp. 200–209. 68, 70, 71

J. Johnson, M. Douze, and H. Jégou. Feb. 2017. Billion-scale similarity search with GPU. *arXiv:1702.08734*. 132

W. B. Johnson and J. Lindenstrauss. 1984. Extensions of Lipschitz mappings into a Hilbert space. *Contemp. Math.*, (26): 189–206. DOI: 10.1090/conm/026/737400. 107

H. Joho, J. Staiano, N. Sebe, and J. M. Jose. 2011. Looking at the viewer: Analysing facial activity to detect personal highlights of multimedia contents. *Multimedia Tools and Applications*, 51(2): 505–523. DOI: 10.1007/s11042-010-0632-x. 221, 242, 243

A. Joly and O. Buisson. 2008. A posteriori multi-probe locality sensitive hashing. In *ACM Multimedia*, pp. 209–218. DOI: 10.1145/1459359.1459388. 118

A. Joly and O. Buisson. Oct. 2009. Logo retrieval with a contrario visual query expansion. In *ACM Multimedia*, pp. 581–584. DOI: 10.1145/1631272.1631361. 106

A. Joly and O. Buisson. 2011. Random maximum margin hashing. In *CVPR*, pp. 873–880. DOI: 10.1109/CVPR.2011.5995709. 133

D. Joshi, R. Datta, E. Fedorovskaya, Q.-T. Luong, J. Z. Wang, J. Li, and J. Luo. 2011. Aesthetics and emotions in images. *Signal Processing Magazine, IEEE*, 28(5): 94–115. DOI: 10.1109/MSP.2011.941851. 150, 156

T. Judd, K. Ehinger, F. Durand, and A. Torralba. 2009. Learning to predict where humans look. In *ICCV*, pp. 2106–2113. DOI: 10.1109/ICCV.2009.5459462. 222, 227

T. Kalker. 2007. A cryptographic method for secure watermark detection. In *Information Hiding*, volume 4437 of *Lecture Notes in Computer Science*, pp. 26–41. Springer. DOI: 10.1155/2007/78943. 81

O. Kallenberg. 2006. *Foundations of Modern Probability*. Springer Science & Business Media. DOI: 10.1007/b98838. 185

V. Kalofolias, X. Bresson, M. Bronstein, and P. Vandergheynst. 2014. Matrix completion on graphs. *arXiv:1408.1717*. 58, 63, 64

Y. Kamarianakis and P. Prastacos. 2003. Forecasting traffic flow conditions in an urban network: Comparison of multivariate and univariate approaches. *Transportation Research Record: Journal of the Transportation Research Board*, (1857): 74–84. DOI: 10.3141/1857-09. 186

M. Kantarcioglu, W. Jiang, Y. Liu, and B. Malin. 2008. A cryptographic approach to securely share and query genomic sequences. *IEEE Transactions on Information Technology in Biomedicine*, 12(5): 606–617. DOI: 10.1109/TITB.2007.908465. 83

A. Kapoor, P. Shenoy, and D. Tan. 2008. Combining brain computer interfaces with vision for object categorization. In *CVPR*, pp. 1–8. DOI: 10.1109/CVPR.2008.4587618. 220

A. Karpathy, G. Toderici, S. Shetty, T. Leung, R. Sukthankar, and L. Fei-Fei. 2014. Large-scale video classification with convolutional neural networks. In *CVPR*, pp. 1725–1732. DOI: 10.1109/CVPR.2014.223. 10, 22

Y. Ke, R. Sukthankar, and L. Huston. 2004. Efficient near-duplicate detection and sub-image retrieval. In *ACM Multimedia*, pp. 869–876. DOI: 10.1145/1027527.1027729. 105

E. G. Kehoe, J. M. Toomey, J. H. Balsters, and A. L. W. Bokde. 2012. Personality modulates the effects of emotional arousal and valence on brain activation. *Social Cognitive & Affective Neuroscience*, 7: 858–870. DOI: 10.1093/scan/nsr059. 244

A. Kendon. 1990. *Conducting Interaction: Patterns of Behavior in Focused Encounters*, volume 7. Cambridge University Press Archive, Cambridge. 51

A. D. Keromytis. 2009. A survey of voice over IP security research. In *Information Systems Security*, volume 5905 of *Lecture Notes in Computer Science*, pp. 1–17. Springer. DOI: 10.1007/978-3-642-10772-6_1. 98

L. A. Khan, M. S. Baig, and A. M. Youssef. 2010. Speaker recognition from encrypted VoIP communications. *Elsevier Digital Investigation*, 7(1): 65–73. DOI: 10.1016/j.diin.2009.10.001. 98, 99, 100

H. Kim, J. Wen, and J. D. Villasenor. 2007. Secure arithmetic coding. *IEEE Transactions on Signal Processing*, 55(5): 2263–2272. DOI: 10.1109/TSP.2007.892710. 86

J. Kim and E. Andre. 2008. Emotion recognition based on physiological changes in music listening. *IEEE Trans. Pattern Analysis and Machine Intelligence*, 30(12): 2067–2083. DOI: 10.1109/TPAMI.2008.26. 242, 243

S.-S. Kim, K.-I. Kim, and J. Won. 2011. Multi-view rendering approach for cloud-based gaming services. In *AFIN 2011, The Third International Conference on Advances in Future Internet*, pp. 102–107. ISBN 9781612081489. 300

S. Kingsbury. 1987. Wisdom for the masses. *American Speech*, pp. 358–360. DOI: 10.2307/455412. 162

H. Kiya and M. Fujiyoshi. 2012. Signal and image processing in the encrypted domain. *ECIT Transactions on Computer and Information Technology*, 6(1): 11–18. 87, 88, 90

A. Klapuri. 2003. Multiple fundamental frequency estimation by harmonicity and spectral smoothness. *IEEE Trans. Speech and Audio Processing*, 11(6): 804–816. http://www.cs.tut.fi/sgn/arg/klap/multiplef0.pdf. DOI: 10.1109/TSA.2003.815516. 37

S. Koelstra and I. Patras. 2013. Fusion of facial expressions and EEG for implicit affective tagging. *Image and Vision Computing*, 31(2): 164–174. DOI: 10.1016/j.imavis.2012.10.002. 247

S. Koelstra, C. Mühl, M. Soleymani, J.-S. Lee, A. Yazdan, T. Ebrahimi, T. Pun, A. Nijholt, and I. Patras. 2012. DEAP: A database for emotion analysis using physiological signals. *IEEE Trans. Affective Computing*, 3(1): 18–31. DOI: 10.1109/T-AFFC.2011.15. 221, 240, 242, 247, 249

C. Kofler, M. Larson, and A. Hanjalic. 2014. Intent-aware video search result optimization. *IEEE Transactions on Multimedia*, 16(5): 1421–1433. DOI: 10.1109/TMM.2014.2315777. 140

A. Kojima, T. Tamura, and K. Fukunaga. 2002. Natural language description of human activities from video images based on concept hierarchy of actions. *IJCV*, pp. 171–184. DOI: 10.1023/A:1020346032608. 16

D. Kounades-Bastian, L. Girin, X. Alameda-Pineda, S. Gannot, and R. Horaud. 2016. A variational EM algorithm for the separation of time-varying convolutive audio mixtures. *IEEE/ACM Transactions on Audio, Speech and Language Processing*, 24(8): 1408–1423. DOI: 10.1109/TASLP.2016.2554286. 53

D. Kounades-Bastian, L. Girin, X. Alameda-Pineda, S. Gannot, and R. Horaud. 2017. An EM algorithm for joint source separation and diarisation of multichannel convolutive mixtures. In *IEEE International Conference on Audio, Speech and Signal Processing*. New Orleans, USA. DOI: 10.1109/ICASSP.2017.7951789. 53

N. Krahnstoever, M.-C. Chang, and W. Ge. 2011. Gaze and body pose estimation from a distance. In *AVSS*, pp. 11–16. DOI: 10.1109/AVSS.2011.6027285. 57

G. Krieger, I. Rentschler, G. Hauske, K. Schill, and C. Zetzsche. 2000. Object and scene analysis by saccadic eye-movements: An investigation with higher-order statistics. *Spatial Vision*, 13(2–3): 201–214. DOI: 10.1163/156856800741216. 230

A. Krizhevsky, I. Sutskever, and G. E. Hinton. 2012. ImageNet classification with deep convolutional neural networks. In *NIPS*, pp. 1097–1105. DOI: 10.1145/3065386. 3, 5, 9, 220

H. Kuehne, H. Jhuang, E. Garrote, T. Poggio, and T. Serre. 2011. HMDB: A large video database for human motion recognition. In *ICCV*, pp. 2556–2563. DOI: 10.1109/ICCV.2011.6126543. 19, 21

G. Kuhn, B. Tatler, and G. Cole. 2009. You look where I look! Effect of gaze cues on overt and covert attention in misdirection. *Visual Cognition*, 17(6–7): 925–944. DOI: 10.1080/13506280902826775. 223

B. Kulis and T. Darrell. December 2009. Learning to hash with binary reconstructive embeddings. In *NIPS*, pp. 1042–1050. 124

B. Kulis and K. Grauman. October 2009. Kernelized locality-sensitive hashing for scalable image search. In *ICCV*, pp. 2130–2137. DOI: 10.1109/TPAMI.2011.219. 133

A. Kumar, P. Dighe, S. Chaudhuri, R. Singh, and B. Raj. 2012. Audio event detection from acoustic unit occurrence patterns. In *Proceedings of IEEE International Conference on Acoustics Speech and Signal Processing (ICASSP)*, pp. 489–492. DOI: 10.1109/ICASSP.2012.6287923. 39

A. Kumar, R. Singh, and B. Raj. 2014. Detecting sound objects in audio recordings. In *22nd European Signal Processing Conference, EUSIPCO 2014, Lisbon, Portugal, September 1–5, 2014*, pp. 905–909. 47

H. Kwak, C. Lee, H. Park, and S. Moon. 2010. What is Twitter, a social network or a news media? In *WWW '10*, pp. 591–600. DOI: 10.1145/1772690.1772751. 213

R. L. Lagendijk, Z. Erkin, and M. Barni. 2013. Encrypted signal processing for privacy protection: Conveying the utility of homomorphic encryption and multiparty computation. *IEEE Signal Processing Magazine*, 30(1): 82–105. DOI: 10.1109/MSP.2012.2219653. 77, 104

O. Lanz. 2006. Approximate Bayesian multibody tracking. *IEEE TPAMI*, 28: 1436–1449. DOI: 10.1109/TPAMI.2006.177. 59, 71

B. M. Lapham. 2014. Hawkes processes and some financial applications. Thesis, University of Cape Town. 208

I. Laptev, M. Marszalek, C. Schmid, and B. Rozenfeld. 2008. Learning realistic human actions from movies. In *CVPR*, pp. 1–8. DOI: 10.1109/CVPR.2008.4587756. 21

A. Lathey and P. K. Atrey. 2015. Image enhancement in encrypted domain over cloud. *ACM Transactions on Multimedia Computing, Communications, and Applications*, 11(3): 38. DOI: 10.1145/2656205. 89, 90

A. Lathey, P. Atrey, and N. Joshi. 2013. Homomorphic low pass filtering over cloud. In *IEEE International Conference on International Conference on Semantic Computing*, pp. 310–313. DOI: 10.1109/ICSC.2013.60. 89, 90

P. J. Laub, T. Taimre, and P. K. Pollett. 2015. Hawkes processes. *arXiv:1507.02822*. 206

J.-P. Laulajainen, T. Sutinen, S. Järvinen, and S. Jarvinen. Apr. 2006. Experiments with QoS-aware gaming-on-demand service. In *20th International Conference on Advanced Information Networking and Applications*, volume 1, pp. 805–810. DOI: 10.1109/AINA.2006.175. 308

R. Lazzeretti, J. Guajardo, and M. Barni. 2012. Privacy preserving ECG quality evaluation. In *Proceedings of the 14th ACM Workshop on Multimedia and Security*, pp. 165–174. ACM. DOI: 10.1145/2361407.2361435. 83

Q. V. Le, W. Y. Zou, S. Y. Yeung, and A. Y. Ng. 2011. Learning hierarchical invariant spatio-temporal features for action recognition with independent subspace analysis. In *CVPR*, pp. 3361–3368. DOI: 10.1145/2361407.2361435. 14

B. B. LeCun, J. S. Denker, D. Henderson, R. E. Howard, W. Hubbard, and L. D. Jackel. 1990. Handwritten digit recognition with a back-propagation network. In *NIPS*, vol. 2. 5

Y. LeCun, L. Bottou, Y. Bengio, and P. Haffner. 2001. Gradient-based learning applied to document recognition. In *Intelligent Signal Processing*, pp. 306–351. DOI: 10.1109/5.726791. 5

C. M. Lee and S. S. Narayanan. 2005. Toward detecting emotions in spoken dialogs. *Speech and Audio Processing*, 13(2): 293–303. DOI: 10.1109/TSA.2004.838534. 221

H. Lei, J. Choi, A. Janin, and G. Friedland. May 2011. Persona linking: Matching uploaders of videos across accounts. In *Proceedings of ICASSP*, pp. 2404–2407. 39

H. Lei, J. Choi, and G. Friedland. 2012. Multimodal city-verification on Flickr videos using acoustic and textual features. In *Proceedings of ICASSP*, pp. 2273–2276. DOI: 10.1109/ICASSP.2012.6288367. 39

H. Lejsek, F. H. Asmundsson, B. P. Jónsson, and L. Amsaleg. May 2009. NV-tree: An efficient disk-based index for approximate search in very large high-dimensional collections. *IEEE Trans. PAMI*, 31(5): 869–883. DOI: 10.1109/TPAMI.2008.130. 112, 114

B. Lepri, R. Subramanian, K. Kalimeri, J. Staiano, F. Pianesi, and N. Sebe. 2012. Connecting meeting behavior with Extraversion—A systematic study. *IEEE Trans. Affective Computing*, 3(4): 443–455. DOI: 10.1109/T-AFFC.2012.17. 238, 240

J. Leskovec, A. Singh, and J. Kleinberg. 2006. Patterns of influence in a recommendation network. In *Pacific-Asia Conference on Knowledge Discovery and Data Mining*, pp. 380–389. Springer. DOI: 10.1007/11731139_44. 152

H. J. Levesque, R. Reiter, Y. Lespérance, F. Lin, and R. B. Scherl. 1997. GOLOG: A logic programming language for dynamic domains. *Journal of Logic Programming*, P1(1–3): 59–83. DOI: 10.1016/S0743-1066(96)00121-5. 165

M. S. Lew, N. Sebe, C. Djeraba, and R. Jain. 2006. Content-based multimedia information retrieval: State of the art and challenges. *ACM Transactions on Multimedia Computing, Communications, and Applications*, 2(1): 1–19. DOI: 10.1145/1126004.1126005. 85

J. Li, Q. Wang, C. Wang, N. Cao, K. Ren, and W. Lou. 2010. Fuzzy keyword search over encrypted data in cloud computing. In *Proceedings of IEEE INFOCOM*, pp. 441–445. DOI: 10.1007/978-3-642-38562-9_74. 260

P. Li, X. Yu, Y. Liu, and T. Zhang. 2014. Crowdsourcing fraud detection algorithm based on Ebbinghaus forgetting curve. *International Journal of Security and Its Applications*, 8(1): 283–290. DOI: 10.14257/ijsia.2014.8.1.26. 262

Q. Li, Z. Qiu, T. Yao, T. Mei, Y. Rui, and J. Luo. 2016a. Action recognition by learning deep multi-granular spatio-temporal video representation. In *ICMR*, 159–166. DOI: 10.1145/2911996.2912001. 12

W. Li, Y. Zhang, Y. Sun, W. Wang, W. Zhang, and X. Lin. 2016b. Approximate nearest neighbor search on high dimensional data—Experiments, analyses, and improvement (v1.0). *arXiv:1610.02455*. 134

X. Li, C. G. Snoek, and M. Worring. 2009. Learning social tag relevance by neighbor voting. *IEEE Transactions on Multimedia*, 11(7): 1310–1322. DOI: 10.1109/TMM.2009.2030598. 148

Z. Li, E. Gavves, M. Jain, and C. G. Snoek. 2016c. VideoLSTM convolves, attends and flows for action recognition. *arXiv:1607.01794*. 13

C. Liao, T. Hou, T. Lin, Y. Cheng, C. H. A. Erbad, and N. Venkatasubramania. 2014. SAIS: Smartphone augmented infrastructure sensing for public safety and sustainability in smart cities. In *Proceedings of ACM International Workshop on Emerging Multimedia Applications and Services for Smart Cities (EMASC)*, pp. 3–8. DOI: 10.1145/2661704.2661706. 268, 269, 270, 271, 272

L. Lin, X. Liao, G. Tan, H. Jin, X. Yang, W. Zhang, and B. Li. 2014. LiveRender: A Cloud Gaming System Based on Compressed Graphics Streaming. In *Proceedings of the ACM International Conference on Multimedia*, MM '14, pp. 347–356. ACM, New York. DOI: 10.1145/2647868.2654943. 307

T. Lin, T. Lin, C. Hsu, and C. King. 2013. Context-aware decision engine for mobile cloud offloading. In *Proceedings of IEEE Wireless Communications and Networking Conference Workshops (WCNCW)*, pp. 111–116. DOI: 10.1109/WCNCW.2013.6533324. 258

F. Lingenfelser, J. Wagner, E. André, G. McKeown, and W. Curran. 2014. An event driven fusion approach for enjoyment recognition in real-time. In *ACMMM*, pp. 377–386. DOI: 10.1145/2647868.2654924. 52

M. Lipczak, M. Trevisiol, and A. Jaimes. 2013. Analyzing favorite behavior in Flickr. In *International Conference on Multimedia Modeling*, pp. 535–545. Springer. DOI: 10.1007/978-3-642-35725-1_49. 148

S. Liu, P. Cui, H. Luan, W. Zhu, S. Yang, and Q. Tian. 2013a. Social visual image ranking for web image search. In *International Conference on Multimedia Modeling*, pp. 239–249. Springer. DOI: 10.1007/978-3-642-35725-1_22. 149

S. Liu, P. Cui, W. Zhu, S. Yang, and Q. Tian. 2014. Social embedding image distance learning. In *Proceedings of the 22nd ACM International Conference on Multimedia*, pp. 617–626. ACM. DOI: 10.1145/2647868.2654905. 148, 149

S. Liu, P. Cui, W. Zhu, and S. Yang. 2015. Learning socially embedded visual representation from scratch. In *Proceedings of the 23rd ACM International conference on Multimedia*, pp. 109–118. ACM. DOI: 10.1145/2733373.2806247. 149

X. Liu, J. He, and B. Lang. July 2013b. Reciprocal hash tables for nearest neighbor search. In *Proceedings of the 27th Association for the Advancement of Artificial Intelligence (AAAI) Conference on Artificial Intelligence*, pp. 626–632. 118

Y. Liu and Z. Shi. 2016. Boosting video description generation by explicitly translating from frame-level captions. In *ACM Multimedia*, pp. 631–634. DOI: 10.1145/2964284.2967298. 17, 28

Y. Liu, Y. Guo, and C. Liang. 2008. A survey on peer-to-peer video streaming systems. *Peer-to-peer Networking and Applications*, 1(1): 18–29. DOI: 10.1007/s12083-007-0006-y. 259

B. Logan et al. 2000. Mel frequency cepstral coefficients for music modeling. In *International Symposium on Music Information Retrieval*, volume 28, p. 5. 37

LogMeIn. July 2012. LogMeIn web page. http://secure.logmein.com/. 291

J. Long, E. Shelhamer, and T. Darrell. 2015. Fully convolutional networks for semantic segmentation. In *CVPR*, p. 3431–3440. DOI: 10.1109/TPAMI.2016.2572683. 3, 9

A. Loui, J. Luo, S.-F. Chang, D. Ellis, W. Jiang, L. Kennedy, K. Lee, and A. Yanagawa. 2007. Kodak's consumer video benchmark data set: Concept definition and annotation. In *ACM Multimedia Information Retrieval (MIR) Workshop*, pp. 245–254. DOI: 10.1145/1290082.1290117. 20

D. Lowe. 2004. Distinctive image features from scale-invariant keypoints. *International Journal of Computer Vision*, 60(2): 91–110. 106, 107, 275

W. Lu, A. Swaminathan, A. L. Varna, and M. Wu. 2009a. Enabling search over encrypted multimedia databases. *Proceedings of International Society for Optics and Photonics, SPIE*, Media Forensics and Security, pp. 7254–7318. DOI: 10.1117/12.806980. 82, 85, 86, 90

W. Lu, A. L. Varna, A. Swaminathan, and M. Wu. 2009b. Secure image retrieval through feature protection. In *IEEE International Conference on Acoustics, Speech and Signal Processing*, pp. 1533–1536. DOI: 10.1109/ICASSP.2009.4959888. 85, 86, 90

W. Lu, A. L. Varna, and M. Wu. 2010. Security analysis for privacy preserving search of multimedia. In *IEEE International Conference on Image Processing*, pp. 2093–2096. DOI: 10.1109/ICIP.2010.5653399. 86

W. Lu, A. Varna, and M. Wu. 2011. Secure video processing: Problems and challenges. In *Proceedings of IEEE International Conference on Acoustics, Speech and Signal Processing*, pp. 5856–5859. DOI: 10.1109/ICASSP.2011.5947693. 77

P. Lucey, J. F. Cohn, T. Kanade, J. Saragih, Z. Ambadar, and I. Matthews. 2010. The extended Cohn-Kanade dataset (CK+): A complete dataset for action unit and emotion-specified expression. In *CVPR Workshops*, pp. 94–101. DOI: 10.1109/CVPRW.2010.5543262. 221

Y. Luo, S. Ye, and S. Cheung. 2010. Anonymous subject identification in privacy-aware video surveillance. In *Proceedings of IEEE International Conference on Multimedia and Expo*, pp. 83–88. DOI: 10.1109/ICME.2010.5583561. 92

Q. Lv, M. Charikar, and K. Li. Nov. 2004. Image similarity search with compact data structures. In *CIKM*, pp. 208–217. DOI: 10.1145/1031171.1031213. 120

Q. Lv, W. Josephson, Z. Wang, M. Charikar, and K. Li. 2007. Multi-probe LSH: Efficient indexing for high-dimensional similarity search. In *Proceedings of the International Conference on Very Large DataBases*, pp. 950–961. 118

M. S. Magnusson. 2000. Discovering hidden time patterns in behavior: T-patterns and their detection. *Behavior Research Methods, Instruments, & Computers*, 32(1): 93–110. DOI: 10.3758/BF03200792. 187

R. Maher and J. Beauchamp. 1994. Fundamental frequency estimation of musical signals using a two-way mismatch procedure. *The Journal of the Acoustical Society of America*, 95(4): 2254–2263. DOI: 10.1121/1.408685. 37

Y. Malkov, A. Ponomarenko, A. Logvinov, and V. Krylov. 2014. Approximate nearest neighbor algorithm based on navigable small world graphs. *Information Systems*, 45: 61–68. DOI: 10.1016/j.is.2013.10.006. 133

Y. A. Malkov and D. A. Yashunin. March 2016. Efficient and robust approximate nearest neighbor search using hierarchical navigable small world graphs. *arXiv:1603.09320*. 133

M. Mandel and D. P. W. Ellis. Sept. 2008. Multiple-instance learning for music information retrieval. In *Proceedings ISMIR*, pp. 577–582. Philadelphia, PA. http://www.ee.columbia.edu/~dpwe/pubs/MandelE08-MImusic.pdf. 38

M. I. Mandel and D. P. W. Ellis. Sept. 2005. Song-level features and support vector machines for music classification. In *Proceedings International Conference on Music Information Retrieval ISMIR*, pp. 594–599. London. http://www.ee.columbia.edu/~dpwe/pubs/ismir05-svm.pdf. 38

C. D. Manning, P. Raghavan, and H. Schütze. 2008. *Introduction to Information Retrieval*. Cambridge University Press. 116

Y. Mao and M. Wu. 2006. A joint signal processing and cryptographic approach to multimedia encryption. *Proceedings of IEEE Transactions on Image Processing*, 15(7): 2061–2075. DOI: 10.1109/TIP.2006.873426. 92, 93, 96

O. Maron and T. Lozano-Perez. 1998. A framework for multiple-instance learning. In *Advances in Neural Information Processing Systems*, pp. 570–576. 48

M. Marszalek, I. Laptev, and C. Schmid. 2009. Actions in context. In *CVPR*. pp. 2929–2936. DOI: 10.1109/CVPRW.2009.5206557. 21

J. Martinez, H. H. Hoos, and J. J. Little. 2014. Stacked quantizers for compositional vector compression. *arXiv:1411.2173*. 131

I. Martínez-Ponte, X. Desurmont, J. Meessen, and J.-F. Delaigle. 2005. Robust human face hiding ensuring privacy. In *Proceedings of International Workshop on Image Analysis for Multimedia Interactive Services*, p. 4. DOI: 10.1.1.72.4758. 91

B. Matei, Y. Shan, H. Sawhney, Y. Tan, R. Kumar, D. Huber, and M. Hebert. July 2006. Rapid object indexing using locality sensitive hashing and joint 3D-signature space estimation. *IEEE Trans. PAMI*, 28(7): 1111–1126. DOI: 10.1109/TPAMI.2006.148. 105

A. Matic, V. Osmani, A. Maxhuni, and O. Mayora. 2012. Multi-modal mobile sensing of social interactions. In *PervasiveHealth*. (Pervasive Computing Technologies for Healthcare), IEEE, May 2012. DOI: 10.4108/icst.pervasivehealth.2012.248689. 55

A. Matviienko, A. Löcken, A. El Ali, W. Heuten, and S. Boll. 2016. NaviLight: Investigating ambient light displays for turn-by-turn navigation in cars. In *Proceedings of the 18th International Conference on Human-Computer Interaction with Mobile Devices and Services*, pp. 283–294. ACM. DOI: 10.1145/2935334.2935359. 188

J. McCarthy, and P. Hayes. 1968. *Some philosophical problems from the standpoint of artificial intelligence*. Stanford University: Stanford Artificial Intelligence Project. 165, 167

W. McCulloch and W. Pitts. 1943. A logical calculus of the ideas immanent in nervous activity. *Bulletin of Mathematical Biophysics*, 5(4): 115–133. DOI: 10.1007/BF02478259. 5

MediaEval. 2010. MediaEval's video CLEF-Placing task. http://www.multimediaeval.org/placing/placing.html. 39

D. Meilander, F. Glinka, S. Gorlatch, L. Lin, W. Zhang, X. Liao, and D. Meil. Apr. 2014. Bringing mobile online games to clouds. In *2014 IEEE Conference on Computer Communications Workshops*, pp. 340–345. DOI: 10.1109/INFCOMW.2014.6849255. 308

D. K. Mellinger. 1991. *Event formation and separation in musical sound*. PhD thesis, CCRMA, Stanford University. 37

Merriam-Webster. 2003. *Merriam-Webster's Collegiate Dictionary*. Merriam-Webster. 166, 167

R. Mertens, P. S. Huang, L. Gottlieb, G. Friedland, and A. Divakarian. 2011a. On the applicability of speaker diarization to audio concept detection for multimedia retrieval. In *Proceedings of the IEEE International Symposium on Multimedia*, pp. 446–451. DOI: 10.1109/ISM.2011.79. 43

R. Mertens, H. Lei, L. Gottlieb, G. Friedland, and A. Divakarian. 2011b. Acoustic super models for large scale video event detection. In *Proceedings of the Joint ACM Workshop on Modeling and Representing Events*, pp. 19–24. DOI: 10.1145/2072508.2072513. 36

I. Mervielde, F. De Fruyt, and S. Jarmuz. 1998. Linking openness and intellect in childhood and adulthood. In *Parental Descriptions of Child Personality: Developmental Antecedents of the Big Five*, pp. 105–126. 246

P. Messaris. 1996. *Visual Persuasion: The Role of Images in Advertising*. Sage Publications. 187

P. B. Miltersen, N. Nisan, S. Safra, and A. Wigderson. August 1998. On data structures and asymmetric communication complexity. *Journal of Computer and System Sciences*, 57: 37–49. DOI: 10.1006/jcss.1998.1577. 108

D. Mishra, M. E. Zarki, A. Erbad, C.-H. Hsu, N. Venkatasubramanian, and M. El Zarki. May 2014. Clouds + games: A multifaceted approach. *Internet Computing, IEEE*, 18(3): 20–27. DOI: 10.1109/MIC.2014.20. 289, 290, 291

M. Mishra, A. Das, P. Kulkarni, and A. Sahoo. 2012. Dynamic resource management using virtual machine migrations. *IEEE Communications Magazine*, 50(9): 34–40. DOI: 10.1109/MCOM.2012.6295709. 259

S. Mishra, M.-A. Rizoiu, and L. Xie. 2016. Feature driven and point process approaches for popularity prediction. In *Proceedings of the 25th ACM International Conference on Information and Knowledge Management*, pp. 1069–1078. DOI: 10.1145/2983323.2983812. 209, 211, 213

M. Mohanty, P. K. Atrey, and W.-T. Tsang. 2012. Secure cloud-based medical data visualization over cloud. In *Proceedings of ACM International Conference on Multimedia*, pp. 1105–1108. DOI: 10.1145/2393347.2396394. 89, 90

M. Mohanty, W.-T. Tsang, and P. K. Atrey. 2013. Scale me, crop me, know me not: Supporting scaling and cropping in secret image sharing. In *IEEE International Conference on Multimedia and Expo*, pp. 1–6. DOI: 10.1109/ICME.2013.6607567. 89, 90

D. Moise, D. Shestakov, G. P. Gudmundsson, and L. Amsaleg. April 2013. Indexing and searching 100M images with Map-Reduce. In *ICMR*, pp. 17–24. DOI: 10.1145/2461466.2461470. 113

J. Møller and J. G. Rasmussen. 2005. Perfect simulation of Hawkes processes. *Advances in Applied Probability*, 37(3): 629–646. DOI: 10.1017/S0001867800000392. 208

M. Montanari, S. Mehrotra, and N. Venkatasubramanian. 2007. Architecture for an automatic customized warning system. In *2007 IEEE Intelligence and Security Informatics*, pp. 32–39. IEEE. DOI: 10.1109/ISI.2007.379530. 177, 182

J. A. Moorer. 1975. *On the Segmentation and Analysis of Continuous Musical Sound by Digital Computer*. PhD thesis, Department of Music, Stanford University. 37

N. Moray and T. Sheridan. 2004. Oil sont les neiges d'antan? In *Human Performance, Situation Awareness and Automation II*. D. A. Vincenzi, H. Mouloua, and P. Hancock, eds. Lawrence Erlbaum Associates, Marwah NJ. 165, 167

C. Moreno, N. Tizon, and M. Preda. 2012. Mobile cloud convergence in GaaS: A business model proposition. In *2012 45th Hawaii International Conference on System Science (HICSS)*, pp. 1344–1352. DOI: 10.1109/HICSS.2012.433. 290

C. P. Moreno, A. G. Antolín, and F. D. Fernando. 2001. Recognizing voice over IP: A robust front-end for speech recognition on the world wide web. *Proceedings of IEEE Transactions on Multimedia*, 3(2): 209–218. DOI: 10.1109/6046.923820. 97

P. J. Moreno, B. Raj, and R. M. Stern. 1996. A vector taylor series approach for environment-independent speech recognition. In *1996 IEEE International Conference on Acoustics, Speech, and Signal Processing*, volume 2, pp. 733–736. IEEE. DOI: 10.1109/ICASSP.1996.543225. 36

F. Morstatter, J. Pfeffer, H. Liu, and K. M. Carley. 2013. Is the sample good enough? Comparing data from Twitter's streaming API with Twitter's firehose. http://scholar.google.de/scholar.bib?q=info:NkS2afIrqyQJ:scholar.google.com/&output=citation&hl=de&as_sdt=0,5&ct=citation&cd=0. 144

M. Muja and D. G. Lowe. February 2009. Fast approximate nearest neighbors with automatic algorithm configuration. In *VISAPP*, pp. 331–340. DOI: 10.5220/0001787803310340. 116, 117

M. Muja and D. G. Lowe. 2014. Scalable nearest neighbor algorithms for high dimensional data. *IEEE Trans. PAMI*, p. 36. DOI: 10.1109/TPAMI.2014.2321376. 117

M. Müller, D. P. W. Ellis, A. Klapuri, and G. Richard. Oct. 2011. Signal processing for music analysis. *IEEE Journal of Selected Topics in Signal Processing*, 5(6): 1088–1110. http://ieeexplore.ieee.org/xpls/abs_all.jsp?arnumber=5709966. DOI: 10.1109/JSTSP.2011.2112333. 38

E. Murphy-Chutorian and M. M. Trivedi. 2009. Head pose estimation in computer vision: A survey. *IEEE Trans. PAMI*, 31(4): 607–626. DOI: 10.1109/TPAMI.2008.106. 219

B. Myers, S. Hudson, and R. Pausch. 2000. Past, present, and future of user interface software tools. *ACM Transactions on Computer-Human Interaction (TOCHI)*, 7(1): 3–28. DOI: 10.1145/344949.344959. 170

G. Mysore, P. Smaragdis, and B. Raj. 2010. Non-negative hidden-Markov modeling of audio with application to source separation. In *Proceedings of 9th International Conference on Latent Variable Analysis and Source Separation (LVA/ICA)*, pp. 140–148. 37

M. Naaman. 2012. Social multimedia: Highlighting opportunities for search and mining of multimedia data in social media applications. *Multimedia Tools and Applications*, 56(1): 9–34. DOI: 10.1007/s11042-010-0538-7. 145

X. Naturel and P. Gros. June 2008. Detecting repeats for video structuring. *Multimedia Tools and Applications*, 38: 233–252. DOI: 10.1007/s11042-007-0180-1. 120

A. Nazari Shirehjini. 2006. Situation modelling: A domain analysis and user study. In *2nd IET International Conference on Intelligent Environments, 2006*, volume 2, pp. 193–199. IET. DOI: 10.1049/cp:20060695. 165

R.-A. Negoescu and D. Gatica-Perez. 2010. Modeling Flickr communities through probabilistic topic-based analysis. *IEEE Transactions on Multimedia*, 12(5): 399–416. DOI: 10.1109/TMM.2010.2050649. 148

E. M. Newton, L. Sweeney, and B. Malin. 2005. Preserving privacy by de-identifying face images. *IEEE Transactions on Knowledge and Data Engineering*, 17(2): 232–243. DOI: 10.1109/TKDE.2005.32. 92, 93

B. Neyshabur and N. Srebro. 2015. On symmetric and asymmetric LSHs for inner product search. In *ICML*, pp. 1926–1934. 133

J. Y.-H. Ng, M. Hausknecht, S. Vijayanarasimhan, O. Vinyals, R. Monga, and G. Toderici. 2015. Beyond short snippets: Deep networks for video classification. In *CVPR*, pp. 4694–4702. DOI: 10.1109/CVPR.2015.7299101. 12, 24

W. Ng. 2009. Clarifying the relation between neuroticism and positive emotions. *Personality and Individual Differences*, 47(1): 69–72. DOI: 10.1016/j.paid.2009.01.049. 244

J. C. Niebles, H. Wang, and L. Fei-Fei. 2008. Unsupervised learning of human action categories using spatial-temporal words. *International Journal of Computer Vision*, 79(3): 299–318. DOI: 10.1007/s11263-007-0122-4. 47

J. C. Niebles, C.-W. Chen, and L. Fei-Fei. 2010. Modeling temporal structure of decomposable motion segments for activity classification. In *ECCV*, pp. 392–405. DOI: 10.1007/978-3-642-15552-9_29. 21

D. Nistér and H. Stewénius. June 2006. Scalable recognition with a vocabulary tree. In *CVPR*, pp. 2161–2168. DOI: 10.1109/CVPR.2006.264. 116

M. Norouzi and D. Fleet. June 2013. Cartesian k-means. In *CVPR*, pp. 3017–3024. DOI: 10.1109/CVPR.2013.388. 131

C. Norris, J. Moran, and G. Armstrong (eds.). 1998. CCTV: A new battleground for privacy. *Surveillance, closed-circuit television and social control*, pp. 243–254. 91

M. Nouh, A. Almaatouq, A. Alabdulkareem, V. K. Singh, E. Shmueli, M. Alsaleh, A. Alarifi, A. Alfaris, and A. S. Pentland. 2014. Social information leakage: Effects of awareness and peer pressure on user behavior. In *Human Aspects of Information Security, Privacy, and Trust*, pp. 352–360. Springer. DOI: 10.1007/978-3-319-07620-1_31. 188

E. Nowak, F. Jurie, and B. Triggs. 2006. Sampling strategies for bag-of-features image classification. *Computer Vision–ECCV 2006*, pp. 490–503. DOI: 10.1007/11744085_38. 167

Y. Ogata. 1981. On Lewis' simulation method for point processes. *IEEE Transactions on Information Theory*, 27(1): 23–31. DOI: 10.1109/TIT.1981.1056305. 202, 204, 209

Y. Ogata. 1988. Statistical models for earthquake occurrences and residual analysis for point processes. *Journal of the American Statistical Association*, 83(401). DOI: 10.1080/01621459.1988.10478560. 200, 209

Y. Ogata, R. S. Matsuúra, and K. Katsura. 1993. Fast likelihood computation of epidemic type aftershock-sequence model. *Geophysical Research Letters*, 20(19): 2143–2146. DOI: 10.1029/93GL02142. 209

A. Ojala and P. Tyrvainen. Jul. 2011. Developing cloud business models: A case study on cloud gaming. *IEEE Software*, 28(4): 42–47. http://ieeexplore.ieee.org/lpdocs/epic03/wrapper.htm?arnumber=5741005. DOI: 10.1109/MS.2011.51. 290

A. Oliva and A. Torralba. 2001. Modeling the shape of the scene: A holistic representation of the spatial envelope. *IJCV*, 42(3): 145–175. DOI: 10.1023/A:1011139631724. 132

D. Omercevic, O. Drbohlav, and A. Leonardis. October 2007. High-dimensional feature matching: employing the concept of meaningful nearest neighbors. In *ICCV*, pp. 1–8 DOI: 10.1109/ICCV.2007.4408880. 106

OnLive. January 2015. OnLive web page. http://www.onlive.com/. 287, 291

M. Osadchy, B. Pinkas, A. Jarrous, and B. Moskovich. 2010. SCiFI—A system for secure face identification. In *Proceedings of IEEE Symposium on Security and Privacy*, pp. 239–254. DOI: 10.1109/SP.2010.39. 94, 96

P. Over, G. Awad, M. Michel, J. Fiscus, G. Sanders, W. Kraaij, A. F. Smeaton, and G. Quéenot. 2014. TRECVID 2014—An overview of the goals, tasks, data, evaluation mechanisms and metrics. In *Proceedings of TRECVID 2014*, National Institute of Standards and Technology (NIST). 19, 21, 23

P. Over, G. Awad, M. Michel, J. Fiscus, W. Kraaij, A. F. Smeaton, G. Quéenot, and R. Ordelman. 2015. TRECVID 2015—An overview of the goals, tasks, data, evaluation mechanisms and metrics. In *Proceedings of TRECVID 2015*, NIST. 23

T. Ozaki. 1979. Maximum likelihood estimation of Hawkes' self-exciting point processes. *Annals of the Institute of Statistical Mathematics*, 31(1): 145–155. DOI: 10.1007/BF02480272. 198

P. Paillier. 1999. Public-key cryptosystems based on composite degree residuosity classes. In *Proceedings of the International Conference on the Theory and Application of Cryptographic Techniques*, volume 1592 of *Lecture Notes in Computer Science*, pp. 223–238. Springer. DOI: 10.1007/3-540-48910-X_16. 79, 87

Y. Pan, T. Yao, T. Mei, H. Li, C.-W. Ngo, and Y. Rui. 2014. Click-through-based cross-view learning for image search. In *Proceedings of the 37th International ACM SIGIR Conference on Research & Development in Information Retrieval*, pp. 717–726. ACM. DOI: 10.1145/2600428.2609568. 17, 148

Y. Pan, Y. Li, T. Yao, T. Mei, H. Li, and Y. Rui. 2016a. Learning deep intrinsic video representation by exploring temporal coherence and graph structure. In *International Joint Conference on Artificial Intelligence (IJCAI)*, pp. 3832–3838. 14

Y. Pan, T. Mei, T. Yao, H. Li, and Y. Rui. 2016b. Jointly modeling embedding and translation to bridge video and language. In *CVPR*, pp. 4594–4602. 15, 16, 17, 28

Y. Pan, T. Yao, H. Li, and T. Mei. 2016c. Video captioning with transferred semantic attributes. *arXiv:1611.07675*. 17

S. Pancoast and M. Akbacak. 2011. Bag-of-audio-words approach for multimedia event classification. In *Proceedings of Interspeech*, pp. 2105–2108. 46

P. F. Panter and W. Dite. Jan. 1951. Quantizing distortion in pulse-count modulation with nonuniform spacing of levels. *Proceedings IRE*, 39: 44–48. DOI: 10.1109/JRPROC.1951.230419. 117

M. Pantic, N. Sebe, J. F. Cohn, and T. Huang. 2005. Affective multimodal human-computer interaction. In *ACMMM*, pp. 669–676. 52

D. P. Papadopoulos, A. D. F. Clarke, F. Keller, and V. Ferrari. 2014. Training object class detectors from eye tracking data. In *ECCV*, pp. 361–376. DOI: 10.1007/978-3-319-10602-1_24. 220, 226

K. Papineni, S. Roukos, T. Ward, and W.-J. Zhu. 2002. Bleu: A method for automatic evaluation of machine translation. In *ACL*, pp. 311–318. DOI: 10.3115/1073083.1073135. 27

O. M. Parkhi, A. Vedaldi, C. V. Jawahar, and A. Zisserman. 2011. The truth about cats and dogs. *2011 International Conference on Computer Vision,* pp. 1427–1434. DOI: 10.1109/ICCV.2011.6126398. 227

D. Parkhurst and E. Niebur. 2003. Scene content selected by active vision. *Spatial Vision*, 16(2): 125–54. DOI: 10.1163/15685680360511645. 230

P. Patel, A. Ranabahu, and A. Sheth. 2009. Service level agreement in cloud computing. In *Proceedings of International Conference on Object-Oriented Programming, Systems, Languages, and Applications (OOPSLA)*, 1(1–10). 260

M. A. Pathak and B. Raj. March 2012. Privacy-preserving speaker verification as password matching. In *ICASSP*, pp. 1849–1852. 105

M. Pathak and B. Raj. 2013. Privacy-preserving speaker verification and identification using gaussian mixture models. *IEEE Transactions on Audio, Speech, and Language Processing*, 21(2): 397–406. DOI: 10.1109/TASL.2012.2215602. 99, 100

L. Paulevé, H. Jégou, and L. Amsaleg. Aug. 2010. Locality sensitive hashing: A comparison of hash function types and querying mechanisms. *Pattern Recognition Letters*, 31(11): 1348–1358. DOI: 10.1016/j.patrec.2010.04.004. 110, 116, 118, 119, 132

E. M. M. Peck, B. F. Yuksel, A. Ottley, R. J. Jacob, and R. Chang. 2013. Using fNIRS brain sensing to evaluate information visualization interfaces. In *ACM Conference on Human Factors in Computing Systems*, pp. 473–482. DOI: 10.1016/j.patrec.2010.04.004. 251

M. Perugini and L. Di Blas. 2002. Analyzing personality-related adjectives from an eticemic perspective: The big five marker scale (BFMS) and the Italian AB5C taxonomy. *Big Five Assessment*, pp. 281–304. 222, 238, 240

R. J. Peters, A. Iyer, L. Itti, and C. Koch. 2005. Components of bottom-up gaze allocation in natural images. *Vision Research*, 45(8): 2397–2416. DOI: 10.1016/j.visres.2005.03.019. 231

S. Petridis, B. Martinez, and M. Pantic. 2013. The MAHNOB laughter database. *Image and Vision Computing*, 31(2): 186–202. DOI: 10.1016/j.imavis.2012.08.014. 55

T. T. Pham, S. Hamid Rezatofighi, I. Reid, and T.-J. Chin. 2016. Efficient point process inference for large-scale object detection. In *Proceedings of the IEEE Conference on Computer Vision and Pattern Recognition*, pp. 2837–2845. DOI: 10.1109/CVPR.2016.310. 193

J. Philbin, O. Chum, M. Isard, J. Sivic, and A. Zisserman. June 2008. Lost in quantization: Improving particular object retrieval in large scale image databases. In *CVPR*, pp. 1–8. DOI: 10.1109/CVPR.2008.4587635. 116, 119

A. Pikrakis, T. Giannakopoulos, and S. Theodoridis. 2008. Gunshot detection in audio streams from movies by means of dynamic programming and Bayesian networks. In *IEEE International Conference on Acoustics, Speech and Signal Processing*, pp. 21–24. IEEE. DOI: 10.1109/ICASSP.2008.4517536. 38

J. K. Pillai, V. M. Patel, R. Chellappa, and N. K. Ratha. 2011. Secure and robust iris recognition using random projections and sparse representations. *IEEE Transactions on Pattern Analysis and Machine Intelligence*, 33(9): 1877–1893. 82

A. Piva and S. Katzenbeisser. 2008. Special issue on signal processing in the encrypted domain. *Hindawi Publishing Corporation, EURASIP Journal on Information Security (eds.)*, 2007. DOI: 10.1155/2007/82790. 77

G. Poliner and D. P. W. Ellis. 2007. A discriminative model for polyphonic piano transcription. *EURASIP Journal on Advances in Signal Processing*, 2007(2007): 9 pages. http://www.ee.columbia.edu/~dpwe/pubs/PoliE06-piano.pdf. DOI: 10.1155/2007/48317. Special Issue on Music Signal Processing. 38

S. Pongpaichet, V. K. Singh, R. Jain, and A. P. Pentland. 2013. Situation fencing: making geo-fencing personal and dynamic. In *2013 ACM International Workshop on Personal Data Meets Distributed Multimedia*. Association for Computing Machinery, pp. 3–10. DOI: 10.1145/2509352.2509401. 161, 177

R. Poppe. 2010. A survey on vision-based human action recognition. *Image and Vision Computing*, pp. 976–990. DOI: 10.1016/j.imavis.2009.11.014. 4

D. A. Pospelov. 1986. *Situational Control: Theory and Practice* (in Russian). Nauka. 165

I. Potamitis, S. Ntalampiras, O. Jahn, and K. Riede. 2014. Automatic bird sound detection in long real-field recordings: Applications and tools. *Applied Acoustics*, 80: 1–9. DOI: 10.1016/j.apacoust.2014.01.001. 38

J. Pouwelse, P. Garbacki, D. Epema, and H. Sips. 2005. The Bittorrent P2P file-sharing system: Measurements and analysis. In *Proceedings of International Workshop on Peer-to-Peer Systems (IPTPS)*, pp. 205–216. DOI: 10.1007/11558989_19. 259

J. Prins, Z. Erkin, and R. L. Lagendijk. 2006. Literature study: Signal processing in the encrypted domain. Technical report, Information and Communication Theory Group, Delft University of Technology. 77

W. Puech, Z. Erkin, M. Barni, S. Rane, and R. L. Lagendijk. 2012. Emerging cryptographic challenges in image and video processing. *Proceedings of* $19^{th}$ *IEEE International Conference on Image Processing*, pp. 2629–2632. DOI: 10.1109/ICIP.2012.6467438. 77, 82, 103

Z. Qin, J. Yan, K. Ren, C. W. Chen, and C. Wang. 2014. Towards efficient privacy-preserving image feature extraction in cloud computing. In *Proceedings of the ACM International Conference on Multimedia*, pp. 497–506. ACM. DOI: 10.1145/2647868.2654941. 87, 90

F. Qiu and J. Cho. 2006. Automatic identification of user interest for personalized search. In *Proceedings of the 15th International Conference on World Wide Web*, pp. 727–736. ACM. DOI: 10.1145/1135777.1135883. 147

Z. Qiu, T. Yao, and T. Mei. 2016. Deep quantization: Encoding convolutional activations with deep generative model. *arXiv:1611.09502*. 9

J. Rabin, J. Delon, and Y. Gousseau. Dec. 2008. A contrario matching of SIFT-like descriptors. In *ICPR*, pp. 1–4. DOI: 10.1109/ICPR.2008.4761371. 106

J. Rabin, J. Delon, and Y. Gousseau. Sep. 2009. A statistical approach to the matching of local features. *SIAM Journal on Imaging Sciences*, 2(3): 931–958. DOI: 10.1137/090751359. 106

M. Radovanović, A. Nanopoulos, and M. Ivanović. Dec. 2010. Hubs in space: Popular nearest neighbors in high-dimensional data. *Journal of Machine Learning Research*, 11: 2487–2531. 106

M. Raginsky and S. Lazebnik. 2010. Locality-sensitive binary codes from shift-invariant kernels. In *NIPS*, pp. 1509–1517. 122

A. Rahimi and B. Recht. 2007. Random features for large-scale kernel machines. In *NIPS*, pp. 1177–1184. 122

B. Raj and R. M. Stern. 2005. Missing-feature approaches in speech recognition. *Signal Processing Magazine, IEEE*, 22(5): 101–116. DOI: 10.1109/MSP.2005.1511828. 36

B. Raj, T. Virtanen, S. Chaudhuri, and R. Singh. 2010a. Non-negative matrix factorization based compensation of music for automatic speech recognition. In *Proceedings of Interspeech*, pp. 717–720. 37

B. Raj, T. Virtanen, S. Chaudhuri, and R. Singh. 2010b. Ungrounded non-negative independent factor analysis. In *Proceedings of Interspeech*, pp. 330–333. 37

B. Raj, R. Singh, and T. Virtanen. 2011. Phoneme-dependent NMF for speech enhancement in monaural mixtures. In *Proceedings of Interspeech*, pp. 1217–1220. 37

A. K. Rajagopal, R. Subramanian, R. L. Vieriu, E. Ricci, O. Lanz, K. Ramakrishnan, and N. Sebe. 2012. An adaptation framework for head-pose classification in dynamic multi-view scenarios. In *Asian Conference on Computer Vision (ACCV)*, pp. 652–666. Springer Berlin Heidelberg. DOI: 10.1007/978-3-642-37444-9_51. 57

A. K. Rajagopal, R. Subramanian, E. Ricci, R. L. Vieriu, O. Lanz, and N. Sebe. 2014. Exploring transfer learning approaches for head pose classification from multi-view surveillance images. *IJCV*, 109(1–2): 146–167. DOI: 10.1007/s11263-013-0692-2. 57

U. Rajashekar, I. van der Linde, A. C. Bovik, and L. K. Cormack. 2008. GAFFE: A gaze-attentive fixation finding engine. *IEEE Transactions on Image Processing*, 17(4): 564–573. DOI: 10.1109/TIP.2008.917218. 231

S. Rane and M. Barni. 2011. Special session on secure signal processing. *Proceedings of IEEE International Conference on Acoustics, Speech and Signal Processing (co-chaired)*, pp. 5848–5871. 77

J. Rasmussen, 2011. Temporal point processes: The conditional intensity function. http://people.math.aau.dk/~jgr/teaching/punktproc11/tpp.pdf. 206

J. G. Rasmussen. 2013. Bayesian inference for Hawkes processes. *Methodology and Computing in Applied Probability*, 15(3): 623–642. DOI: 10.1007/s11009-011-9272-5. 206, 208

A. S. Razavian, H. Azizpour, J. Sullivan, and S. Carlsson. 2014. CNN features off-the-shelf: An astounding baseline for recognition. In *CVPR Workshop*, pp. 512–519. DOI: 10.1109/CVPRW.2014.131. 9

R. Reiter. 2001. *Knowledge in Action: Logical Foundations for Specifying and Implementing Dynamical Systems.* MIT Press. 165

E. Ricci, G. Zen, N. Sebe, and S. Messelodi. 2013. A prototype learning framework using EMD: Application to complex scenes analysis. *IEEE TPAMI*, 35(3): 513–526. DOI: 10.1109/TPAMI.2012.131. 55

E. Ricci, J. Varadarajan, R. Subramanian, S. Rota Bulo, N. Ahuja, and O. Lanz. 2015. Uncovering interactions and interactors: Joint estimation of head, body orientation and f-formations from surveillance videos. In *IEEE ICCV*, pp. 4660–4668. DOI: 10.1109/ICCV.2015.529. 57, 70, 71, 72, 73

L. Riungu-Kalliosaari, J. Kasurinen, and K. Smolander. 2011. Cloud services and cloud gaming in game development. *IADIS International Conference Game and Entertainment Technologies 2013 (GET 2013)*, (1). 290

R. L. Rivest, A. Shamir, and L. Adleman. 1978. A method for obtaining digital signatures and public-key cryptosystems. *Communications of the ACM*, 21(2): 120–126. DOI: 10.1145/359340.359342. 79

M.-A. Rizoiu, L. Xie, S. Sanner, M. Cebrian, H. Yu, and P. Van Hentenryck. 2017. Expecting to be HIP: Hawkes intensity processes for social media popularity. In *International Conference on World Wide Web 2017*, pp. 1–9. Perth, Australia. http://arxiv.org/abs/1602.06033. DOI: 10.1145/3038912.3052650. 198

N. Robertson and I. Reid. 2006. Estimating gaze direction from low-resolution faces in video. In *ECCV*, pp. 402–415. DOI: 10.1007/11744047_31. 57

A. Rohrbach, M. Rohrbach, W. Qiu, A. Friedrich, M. Pinkal, and B. Schiele. 2014. Coherent multi-sentence video description with variable level of detail. In *GCPR*, pp. 184–195. DOI: 10.1007/978-3-319-11752-2_15. 16, 26, 27, 29

A. Rohrbach, M. Rohrbach, and B. Schiele. 2015a. The long-short story of movie description. In *GCPR*, pp. 209–221. DOI: 10.1007/978-3-319-24947-6_17. 28

A. Rohrbach, M. Rohrbach, N. Tandon, and B. Schiele. 2015b. A dataset for movie description. In *CVPR*, pp. 3202–3212. 26, 27, 28

M. Rohrbach, W. Qiu, I. Titov, S. Thater, M. Pinkal, and B. Schiele. 2013. Translating video content to natural language descriptions. In *ICCV*, pp. 433–440. DOI: 10.1109/ICCV.2013.61. 16, 29

M. Rohrbach, A. Rohrbach, M. Regneri, S. Amin, M. Andriluka, M. Pinkal, and B. Schiele. 2015c. Recognizing fine-grained and composite activities using hand-centric features and script data. *IJCV*, 119(3): 346–373. DOI: 10.1007/s11263-015-0851-8. 26

R. C. Rose and D. B. Paul. 1990. A hidden Markov model based keyword recognition system. In *Proceedings of IEEE International Conference on Acoustics, Speech, and Signal Processing*, pp. 129–132. DOI: 10.1109/ICASSP.1990.115555. 95, 98

P. Ross. Mar. 2009. Cloud Computing's Killer App: Gaming. *IEEE Spectrum*, 46(3): 14. http://ieeexplore.ieee.org/lpdocs/epic03/wrapper.htm?arnumber=4795441. DOI: 10.1109/MSPEC.2009.4795441. 289

H. A. Rowley, S. Baluja, and T. Kanade. 1998. Neural network-based face detection. *IEEE TPAMI*, 20(1): 23–38. DOI: 10.1109/34.655647. 225

Y. Rui, T. S. Huang, M. Ortega, and S. Mehrotra. 1998. Relevance feedback: A power tool for interactive content-based image retrieval. *IEEE Transactions on Circuits and Systems for Video Technology*, 8(5): 644–655. DOI: 10.1109/76.718510. 148

O. Russakovsky, J. Deng, H. Su, J. Krause, S. Satheesh, S. Ma, Z. Huang, A. Karpathy, A. Khosla, M. Bernstein, et al. 2015. Image Net large scale visual recognition challenge. *IJCV*, 115(3): 211–252. DOI: 10.1007/s11263-015-0816-y. 3

B. C. Russell, A. Torralba, K. P. Murphy, and W. T. Freeman. 2008. Labelme: A database and web-based tool for image annotation. *IJCV*, 77(1–3): 157–173. DOI: 10.1007/s11263-007-0090-8. 220

J. Russell. 1980. A circumplex model of affect. *Journal of Personality and Social Psychology*, 39: 1161–1178. DOI: 10.1037/h0077714. 236

M. Ryynanen and A. Klapuri. April 2008. Query by humming of midi and audio using locality sensitive hashing. In *ICASSP*, pp. 2249–2252. DOI: 10.1109/ICASSP.2008.4518093. 105

A. Sablayrolles, M. Douze, N. Usunier, and H. Jégou. 2016. How should we evaluate supervised hashing? *arXiv:1609.06753*. 121

A. R. Sadeghi, T. Schneider, and I. Wehrenberg. 2010. Efficient privacy-preserving face recognition. In *Proceedings of the* $12^{th}$ *International Conference on Information, Security and Cryptology*, volume 5984 of *Lecture Notes in Computer Science*, pp. 229–244. Springer. DOI: 10.1007/978-3-642-14423-3_16. 94, 96

S. Saeger, B. Elizalde, C. Schulze, D. Borth, B. Raj, and I. Lane. 2016, under review. Audio content descriptors. *IEEE Transactions on Audio Speech and Language Processing*. 48

M. S. SaghaianNejadEsfahani, Y. Luo, and S.-C. Sen. 2012. Privacy protected image denoising with secret shares. In *Proceedings of the* $19^{th}$ *IEEE International Conference on Image Processing*, pp. 253–256. DOI: 10.1109/ICIP.2012.6466843. 89, 90

M. Saini, P. K. Atrey, S. Mehrotra, and M. S. Kankanhalli. 2012. $W^3$-Privacy: Understanding what, when, and where inference channels in multi-camera surveillance video. *Springer International Journal of Multimedia Tools and Applications*, 68(17): 135–158. DOI: 10.1007/s11042-012-1207-9. 75

M. Saini, P. K. Atrey, S. Mehrotra, and M. S. Kankanhalli. 2013. Privacy aware publication of surveillance video. *Inderscience International Journal of Trust Management in Computing and Communications*, 1(1): 23–51. DOI: 10.1504/IJTMCC.2013.052523. 92, 93

H. Samet. 2007. *Foundations of Multidimensional and Metric Data Structures*. Elsevier. 107

J. Sánchez, F. Perronnin, T. Mensink, and J. Verbeek. 2013. Image classification with the Fisher vector: Theory and practice. *IJCV*, 105(3): 222–245. DOI: 10.1007/s11263-013-0636-x. 9

T. Sander and C. Tschudin. 1998. On software protection via function hiding. In *Information Hiding*, volume 1525 of *Lecture Notes in Computer Science*, pp. 111–123. Springer. DOI: 10.1007/3-540-49380-8_9. 81

H. Sandhawalia and H. Jégou. March 2010. Searching with expectations. In *ICASSP, Signal Processing*, pp. 1242–1245. DOI: 10.1109/ICASSP.2010.5495403. 123, 128

B. Sankaran. 2010. A survey of unsupervised grammar induction. Manuscript, Simon Fraser University. 47

A. Santella and D. DeCarlo. 2004. Robust clustering of eye movement recordings for quantification of visual interest. In *Symposium on Eye Tracking Research & Applications*, pp. 27–34. DOI: 10.1145/968363.968368. 223

T. S. Saponas, J. Lester, C. Hartung, S. Agarwal, and T. Kohno. 2007. Devices that tell on you: Privacy trends in consumer ubiquitous computing. In *Proceedings of the 16th Annual USENIX Security Symposium*, volume 3, pp. 55–70. 97

F. Sarmenta. 2001. Volunteer computing. Technical report, Massachusetts Institute of Technology. 258

N. Sarter and D. Woods. 1991. Situation awareness: A critical but ill-defined phenomenon. *The International Journal of Aviation Psychology*, 1(1): 45–57. DOI: 10.1207/s15327108ijap0101_4. 166, 167

M. Satyanarayanan, P. Bahl, R. Caceres, and N. Davies. 2009. The case for VM-based cloudlets in mobile computing. *IEEE Transactions on Pervasive Computing*, 8(4): 14–24. DOI: 10.1109/MPRV.2009.82. 259, 261

M. Schneider and T. Schneider. 2014. Notes on non-interactive secure comparison in image feature extraction in the encrypted domain with privacy-preserving SIFT. In *Proceedings of the 2nd ACM Workshop on Information Hiding and Multimedia Security*, pp. 135–140. ACM. DOI: 10.1145/2600918.2600927. 87

C. Schuldt, I. Laptev, and B. Caputo. 2004. Recognizing human actions: A local SVM approach. In *ICPR*, pp. 32–36. DOI: 10.1109/ICPR.2004.747. 20

N. Sebe, I. Cohen, T. Gevers, and T. S. Huang. 2006. Emotion recognition based on joint visual and audio cues. In *International Conference on Pattern Recognition*, volume 1, pp. 1136–1139. DOI: 10.1109/ICPR.2006.489. 221

M. L. Seltzer, B. Raj, and R. M. Stern. 2004. Likelihood-maximizing beamforming for robust hands-free speech recognition. *IEEE Transactions on Speech and Audio Processing*, 12(5): 489–498. DOI: 10.1109/TSA.2004.832988. 36

A. Senior, S. Pankanti, A. Hampapur, L. Brown, Y.-L. Tian, and A. Ekin. 2003. Blinkering surveillance: Enabling video privacy through computer vision. Technical report, IBM. 92, 93

F. Setti, O. Lanz, R. Ferrario, V. Murino, and M. Cristani. 2013. Multi-scale F-formation discovery for group detection. In *Proceedings of the International Conference on Image Processing (ICIP)*, pp. 3547–3551. DOI: 10.1109/ICIP.2013.6738732. 52, 69, 71

F. Setti, C. Russell, C. Bassetti, and M. Cristani. 2015. F-formation detection: Individuating free-standing conversational groups in images. *PloS ONE*, 10(5). DOI: 10.1371/journal.pone.0123783. 69, 70, 71

H. Shacham and B. Waters. 2008. Compact proofs of retrievability. *Journal of Cryptology*, 26(3): 442–483. DOI: 10.1007/s00145-012-9129-2. 260

G. Shakhnarovich, T. Darrell, and P. Indyk. March 2006. *Nearest-Neighbor Methods in Learning and Vision: Theory and Practice*, chapter 3. MIT Press. 105

S. Sharma, R. Kiros, and R. Salakhutdinov. 2015. Action recognition using visual attention. *arXiv:1511.04119*. 13

J. Shashank, P. Kowshik, K. Srinathan, and C. Jawahar. 2008. Private content based image retrieval. In *IEEE Conference on Computer Vision and Pattern Recognition*, pp. 1–8. DOI: 10.1109/CVPR.2008.4587388. 84, 86, 91, 93

M. V. Shashanka and P. Smaragdis. 2006. Secure sound classification: Gaussian mixture models. In *2006 IEEE International Conference on Acoustics, Speech and Signal Processing*, volume 3, pp. 1088–1091. IEEE. DOI: 10.1109/ICASSP.2006.1660847. 99, 100

M. Shashanka, B. Raj, and P. Smaragdis. May 2008. Probabilistic latent variable models as non-negative factorizations. *Computational Intelligence and Neuroscience*, article 947438. DOI: 10.1155/2008/947438. 37

R. Shea, J. Liu, E. C. Ngai, and Y. Cui. 2013. Cloud gaming: Architecture and performance. *IEEE Network*, 27(August): 16–21. DOI: 10.1109/MNET.2013.6574660. 301

R. Shea, S. Member, D. Fu, S. Member, J. Liu, and S. Member. 2015. Cloud gaming: Understanding the support from advanced virtualization and hardware. *IEEE Transactions on Circuits and Systems for Video Technology*, 25(12): 2026–2037. DOI: 10.1109/TCSVT.2015.2450172. 299, 301, 302, 305

A. Sheh and D. P. Ellis. 2003. Chord segmentation and recognition using EM-trained hidden Markov models. In *Proceedings of the International Conference on Music Information Retrieval ISMIR-03*, http://doi.org/10.7916/D8TB1H83. DOI: 10.7916/D8TB1H83. 37, 38

S. Shekhar, P. R. Schrater, R. R. Vatsavai, W. Wu, and S. Chawla. 2002. Spatial contextual classification and prediction models for mining geospatial data. *IEEE Transactions on Multimedia*, 4(2): 174–188. DOI: 10.1109/TMM.2002.1017732. 186

P. Shenoy and D. S. Tan. 2008. Human-aided computing: Utilizing implicit human processing to classify images. In *ACM Conference on Human Factors in Computing Systems*, pp. 845–854. DOI: 10.1145/1357054.1357188. 220

B. Sheridan. March 9, 2009. A trillion points of data. *Newsweek*, 34–37. 162

S. Shi, C.-H. Hsu, K. Nahrstedt, and R. Campbell. 2011. Using graphics rendering contexts to enhance the real-time video coding for mobile cloud gaming. *Proceedings of the 19th ACM International Conference on Multimedia*, p. 103. http://dl.acm.org/citation.cfm?doid=2072298.2072313. DOI: 10.1145/2072298.2072313. 307

E. Shmueli, V. K. Singh, B. Lepri, and A. Pentland. 2014. Sensing, understanding, and shaping social behavior. *IEEE Transactions on Computational Social Systems*, 1(1): 22–34. DOI: 10.1109/TCSS.2014.2307438. 186

T. Shortell and A. Shokoufandeh. 2015. Secure brightness/contrast filter using fully homomorphic encryption. In *Proceedings of the 14th International Conference on*

*Information Processing in Sensor Networks*, pp. 346–347. ACM. DOI: 10.1145/2737095.2742922. 89, 90

J. Shotton, A. Fitzgibbon, A. Blake, A. Kipman, M. Finocchio, B. Moore, and T. Sharp. 2013. Real-time human pose recognition in parts from single depth images. *Communications of the ACM*, 56(1): 116–124. DOI: 10.1109/CVPR.2011.5995316. 56

A. Shrivastava and P. Li. 2014. Asymmetric LSH for sublinear time maximum inner product search. In *NIPS*, pp. 2321–2329. 133

H. T. Siegelmann and E. D. Sontag. 1991. Turing computability with neural nets. *Applied Mathematics Letters*, 4(6): 77–80. DOI: 10.1.1.47.8383. 6

S. Siersdorfer, J. S. Pedro, and M. Sanderson. 2009. Automatic video tagging using content redundancy. In *Proceedings of the International ACM SIGIR Conference on Research and Development in Information Retrieval*, pp. 395–402. DOI: 10.1145/1571941.1572010. 15

K. Simonyan and A. Zisserman. 2014. Two-stream convolutional networks for action recognition in videos. In *NIPS*, 568–576. 10, 11, 12, 24

K. Simonyan and A. Zisserman. 2015. Very deep convolutional networks for large-scale image recognition. In *ICLR*, pp. 1–14. 5, 9, 16

K. Simonyan, A. Vedaldi, and A. Zisserman. 2013. Learning local feature descriptors using convex optimisation. Technical report, Department of Engineering Science, University of Oxford. 124

R. Singh, B. Raj, and P. Smaragdis. 2010a. Latent-variable decomposition based dereverberation of monaural and multi-channel signals. In *Proceedings of IEEE International Conference on Acoustics Speech and Signal Processing (ICASSP)*, pp. 1914–1917. DOI: 10.1109/ICASSP.2010.5495326. 37

V. Singh and R. Agarwal. 2016. Cooperative phoneotypes: Exploring phone-based behavioral markers of cooperation. In *Proceedings of the ACM International Conference on Ubiquitous Computing*, pp. 646–657. DOI: 10.1145/2971648.2971755. 186

V. Singh and R. Jain. 2009b. Situation based control for cyber-physical environments. In *Military Communications Conference (MILCOM) 2009. IEEE*, pp. 1–7. DOI: 10.1109/MILCOM.2009.5380000. 166, 167, 185

V. Singh, H. Pirsiavash, I. Rishabh, and R. Jain. 2009a. Towards environment-to-environment (E2E) multimedia communication systems. *Multimedia Tools and Applications*, 44(3): 361–388. DOI: 10.1007/s11042-009-0281-0. 185

V. Singh, M. Gao, and R. Jain. 2010b. Event analytics on microblogs. In *Proceedings of the ACM Web Science Conference*, pp. 1–4. ACM. 185

V. Singh, M. Gao, and R. Jain. 2010c. From microblogs to social images: Event analytics for situation assessment. In *Proceedings of the International Conference on Multimedia Information Retrieval*, pp. 433–436. ACM. DOI: 10.1145/1743384.1743460. 185

V. K. Singh and R. Jain. 2016. *Situation Recognition Using Eventshop*. Springer. 161, 162, 170, 172, 175, 177, 183, 185

V. K. Singh, R. Jain, and M. S. Kankanhalli. 2009b. Motivating contributors in social media networks. In *Proceedings of the First SIGMM Workshop on Social Media*, pp. 11–18. ACM. DOI: 10.1145/1631144.1631149. 162

V. K. Singh, M. Gao, and R. Jain. 2010d. Situation detection and control using spatio-temporal analysis of microblogs. In *Proceedings of the 19th International Conference on World Wide Web*, pp. 1181–1182. ACM. DOI: 10.1145/1772690.1772864. 161, 185

V. K. Singh, M. Gao, and R. Jain. 2010e. Social pixels: Genesis and evaluation. In *Proceedings of the 18th ACM International Conference on Multimedia*, pp. 481–490. ACM. DOI: 10.1145/1873951.1874030. 174, 183, 185

V. K. Singh, M. Gao, and R. Jain. 2012. Situation recognition: An evolving problem for heterogeneous dynamic big multimedia data. In *Proceedings of the 20th ACM International Conference on Multimedia*, pp. 1209–1218. ACM. DOI: 10.1145/2393347.2396421. 161, 176, 183, 185

V. K. Singh, T.-S. Chua, R. Jain, and A. S. Pentland. 2013. Summary abstract for the 1st ACM international workshop on personal data meets distributed multimedia. In *Proceedings of the 21st ACM International Conference on Multimedia*, pp. 1105–1106. ACM. DOI: 10.1145/2502081.2503836. 189

V. K. Singh, A. Mani, and A. Pentland. 2014. Social persuasion in online and physical networks. *Proceedings of the IEEE*, 102(12): 1903–1910. DOI: 10.1109/JPROC.2014.2363986. 161, 187

V. K. Singh, S. Pongpaichet, and R. Jain. 2016. Situation recognition from multimodal data. In *Proceedings of the 2016 ACM International Conference on Multimedia Retrieval*, pp. 1–2. ACM. DOI: 10.1145/2911996.2930061. 189

J. Sivic and A. Zisserman. Oct. 2003. Video Google: A text retrieval approach to object matching in videos. In *ICCV*, pp. 1470–1477. DOI: 10.1109/ICCV.2003.1238663. 114, 116

M. Slaney, Y. Lifshits, and J. He. September 2012. Optimal parameters for locality-sensitive hashing. *Proceedings of the IEEE*, 100(9): 2604–2623. DOI: 10.1109/JPROC.2012.2193849. 115, 117

I. Slivar, M. Suznjevic, L. Skorin-Kapov, and M. Matijasevic. Dec. 2014. Empirical QoE study of in-home streaming of online games. In *2014 13th Annual Workshop on Network and Systems Support for Games (NetGames)*, pp. 1–6. DOI: 10.1109/NetGames.2014.7010133. 296

C. Slobogin. 2002. Public privacy: Camera surveillance of public places and the right to anonymity. *Mississippi Law Journal*, 72: 213–301. DOI: 10.2139/ssrn.364600. 91

P. Smaragdis. 1998. Blind separation of convolved mixtures in the frequency domain. *Neurocomputing*, 22(1): 21–34. DOI: 10.1109/ASPAA.1997.625609. 36

P. Smaragdis and B. Raj. 2010. The Markov selection model for concurrent speech recognition. In *IEEE Workshop on Machine Learning for Signal Processing (MLSP)*, pp. 214–219. DOI: 10.1016/j.neucom.2011.09.014. 37

P. Smaragdis and B. Raj. 2011. Missing data imputation for time-frequency representation of audio signals. *Journal of Signal Processing Systems*, 65(3): 361–370. DOI: 10.1007/s11265-010-0512-7. 37

P. Smaragdis and B. Raj. March 2012. The Markov selection model for concurrent speech recognition. *Neurocomputing*, 80: 64–72. DOI: 10.1109/MLSP.2010.5588124. 37

P. Smaragdis, B. Raj, and M. Shashanka. 2009a. Missing data imputation for spectral audio signals. In *IEEE International Workshop for Machine Learning in Signal Processing*, pp. 1–6. DOI: 10.1109/MLSP.2009.5306194. 37

P. Smaragdis, M. Shashanka, and B. Raj. 2009b. A sparse non-parametric approach for single channel separation of known sounds. In *Proceedings of Neural Information Processing Systems (NIPS)*, pp. 1705–1713. 37

A. W. Smeulders, M. Worring, S. Santini, A. Gupta, and R. Jain. 2000. Content-based image retrieval at the end of the early years. *IEEE Transactions on Pattern Analysis and Machine Intelligence*, 22(12): 1349–1380. DOI: 10.1109/34.895972. 85, 220

K. Smith and P. Hancock. 1995. Situation awareness is adaptive, externally directed consciousness. *Human Factors: The Journal of the Human Factors and Ergonomics Society*, 37(1): 137–148. DOI: 10.1518/001872095779049444. 166, 167

B. Smyth. 2007. A community-based approach to personalizing web search. *Computer*, 40(8): 42–50. DOI: 10.1109/MC.2007.259. 151

H. Sohn, K. Plataniotis, and Y. Ro. 2010. Privacy-preserving watch list screening in video surveillance system. In *Proceedings of the* $11^{th}$ *Pacific Rim Conference on Multimedia, Advances in Multimedia Information Processing*, volume 6297 of *Lecture Notes in Computer Science*, pp. 622–632. Springer. DOI: 10.1007/978-3-642-15702-8_57. 94, 96

M. Soleymani, J. Lichtenauer, T. Pun, and M. Pantic. 2012. A multimodal database for affect recognition and implicit tagging. *IEEE Trans. Affective Computing*, 3: 42–55. DOI: 10.1109/T-AFFC.2011.25. 221, 242, 243

O. Soliman, A. Rezgui, H. Soliman, and N. Manea. 2013. Mobile cloud gaming: Issues and challenges. In F. Daniel, G. Papadopoulos, and P. Thiran, eds., *Mobile Web Information Systems SE - 10*, volume 8093 of *Lecture Notes in Computer Science*, pp. 121–128. Springer Berlin Heidelberg. ISBN 978-3-642-40275-3. DOI: 10.1007/978-3-642-40276-0_10. 290, 291

E. Solovey, P. Schermerhorn, M. Scheutz, A. Sassaroli, S. Fantini, and R. Jacob. 2012. Brainput: Enhancing interactive systems with streaming fNIRS brain input. In *ACM Conference on Human Factors in Computing Systems*, pp. 2193–2202. DOI: 10.1145/2207676.2208372. 251

D. Song, D. Wagner, and A. Perig. 2000. Practical techniques for searches on encrypted data. In *Proceedings of IEEE Symposium on Security and Privacy*, pp. 44–55. DOI: 10.1109/SECPRI.2000.848445. 82

Y. Song, L.-P. Morency, and R. Davis. 2012. Multimodal human behavior analysis: Learning correlation and interaction across modalities. In *ICMI*, pp. 27–30. DOI: 10.1145/2388676.2388684. 55

Sony-Gaikai. July 2012. Cloud gaming adoption is accelerating . . . and fast! http://www.nttcom.tv/2012/07/09/cloud-gaming-adoption-is-acceleratingand-fast/. 287

K. Soomro, A. R. Zamir, and M. Shah. 2012. UCF101: A dataset of 101 human actions classes from videos in the wild. Computing Research Repository. http://arxiv.org/corr/home. 19, 21

T. Spindler, C. Wartmann, L. Hovestadt, D. Roth, L. V. Gool, and A. Steffen. 2008. Privacy in video surveilled spaces. *Journal of Computer Security, IOS Press*, 16(2): 199–222. DOI: 10.1145/1501434.1501469. 91, 93

N. Srivastava, E. Mansimov, and R. Salakhutdinov. 2015. Unsupervised learning of video representations using LSTMs. In *ICML*, pp. 843–852. 14, 24

J. Staiano, M. Menéndez, A. Battocchi, A. De Angeli, and N. Sebe. 2012. UX_Mate: From facial expressions to UX evaluation. In *Designing Interactive Systems*, pp. 741–750. DOI: 10.1145/2317956.2318068. 251

A. Steinberg, C. Bowman, and F. White. 1999. Revisions to the JDL data fusion model. Technical report, DTIC Document. DOI: 10.1117/12.341367. 165, 166, 167

G. Stenberg. 1992. Personality and the EEG: Arousal and emotional arousability. *Personality and Individual Differences*, 13: 1097–1113. DOI: 10.1016/0191-8869(92)90025-K. 244

Strategy Analytics. November 2014. Cloud gaming to reach inflection point in 2015. http://tinyurl.com/p3z9hs2.

StreamMyGame. July 2012. StreamMyGame web page. http://streammygame.com/. 291

C. Strecha, A. M. Bronstein, M. M. Bronstein, and P. Fua. January 2012. LDAHash: Improved matching with smaller descriptors. *IEEE Trans. PAMI*, 34(1): 66–78. DOI: 10.1109/TPAMI.2011.103. 120

R. Subramanian, H. Katti, N. Sebe, M. Kankanhalli, and T.-S. Chua. 2010. An eye fixation database for saliency detection in images. In *ECCV*, pp. 30–43. DOI: 10.1007/978-3-642-15561-1_3. 222

R. Subramanian, V. Yanulevskaya, and N. Sebe. 2011. Can computers learn from humans to see better?: Inferring scene semantics from viewers' eye movements. In *ACM Multimedia*, pp. 33–42. DOI: 10.1145/2072298.2072305. 221, 222, 223, 224, 225

R. Subramanian, Y. Yan, J. Staiano, O. Lanz, and N. Sebe. 2013. On the relationship between head pose, social attention and personality prediction for unstructured and dynamic group interactions. In *Int'l Conference on Multimodal Interaction*, pp. 3–10. DOI: 10.1145/2522848.2522862. 219, 238, 240

R. Subramanian, D. Shankar, N. Sebe, and D. Melcher. 2014. Emotion modulates eye movement patterns and subsequent memory for the gist and details of movie scenes. *Journal of Vision*, 14(3): 31. DOI: 10.1167/14.3.31. 237

R. Subramanian, J. Wache, M. Abadi, R. Vieriu, S. Winkler, and N. Sebe. 2016. ASCERTAIN: Emotion and personality recognition using commercial sensors. *IEEE Transactions on Affective Computing*, issue 99. DOI: 10.1109/TAFFC.2016.2625250. 237

Y. Sugano, Y. Matsushita, and Y. Sato. 2013. Graph-based joint clustering of fixations and visual entities. *ACM Transactions on Applied Perception*, 10(2): 1–16. DOI: 10.1145/2465780.2465784. 224

L. Sun, K. Jia, D.-Y. Yeung, and B. E. Shi. 2015. Human action recognition using factorized spatio-temporal convolutional networks. In *CVPR*, pp. 4597–4605. DOI: 10.1109/ICCV.2015.522. 10, 24

Y. Sutcu, Q. Li, and N. Memon. 2007. Protecting biometric templates with sketch: Theory and practice. *IEEE Transactions on Information Forensics and Security*, 2(3): 503–512. 83

I. Sutskever, O. Vinyals, and Q. V. Le. 2014. Sequence to sequence learning with neural networks. In *NIPS*, pp. 3104–3112. 16

A. Swaminathan, Y. Mao, G.-M. Su, H. Gou, A. L. Varna, S. He, M. Wu, and D. W. Oard. 2007. Confidentiality-preserving rank-ordered search. In *Proceedings of the ACM Workshop on Storage Security and Survivability*, pp. 7–12. DOI: 10.1145/1314313.1314316. 82, 85

D. Szajda, M. Pohl, J. Owen, B. G. Lawson, and V. Richmond. 2006. Toward a practical data privacy scheme for a distributed implementation of the Smith-Waterman genome sequence comparison algorithm. In *Network and Distributed System Security Symposium (NDSS)*, pp. 253–265. 83

C. Szegedy, W. Liu, Y. Jia, P. Sermanet, S. Reed, D. Anguelov, D. Erhan, V. Vanhoucke, and A. Rabinovich. 2015a. Going deeper with convolutions. In *CVPR*, pp. 1–9. DOI: 10.1109/CVPR.2015.7298594. 5, 6, 9

C. Szegedy, V. Vanhoucke, S. Ioffe, J. Shlens, and Z. Wojna. 2015b. Rethinking the inception architecture for computer vision. *arXiv:1512.00567*. DOI: 10.1109/CVPR.2016.308. 6

C. Szegedy, S. Ioffe, and V. Vanhoucke. 2017. Inception-v4, Inception-ResNet and the impact of residual connections on learning. In *AAAI*, pp. 4278–4284. 6

K. Takata, J. Ma, B. Apduhan, R. Huang, and N. Shiratori. 2008. Lifelog image analysis based on activity situation models using contexts from wearable multi sensors. In *International Conference on Multimedia and Ubiquitous Engineering*, pp. 160–163. IEEE. DOI: 10.1109/MUE.2008.69. 165

C.-C. Tan, Y.-G. Jiang, and C.-W. Ngo. 2011. Towards textually describing complex video contents with audio-visual concept classifiers. In *ACM Multimedia*, 655–658. DOI: 10.1145/2072298.2072411. 16

J. Tang, S. Yan, R. Hong, G.-J. Qi, and T.-S. Chua. 2009. Inferring semantic concepts from community-contributed images and noisy tags. In *Proceedings of ACM Multimedia*, pp. 223–232. DOI: 10.1145/1631272.1631305. 47

M. Tang, P. Agrawal, S. Pongpaichet, and R. Jain. 2015. Geospatial interpolation analytics for data streams in EventShop. In *2015 IEEE International Conference on Multimedia and Expo (ICME)*, pp. 1–6. IEEE. DOI: 10.1109/ICME.2015.7177513. 186

H. R. Tavakoli, A. Atyabi, A. Rantanen, S. J. Laukka, S. Nefti-Meziani, and J. Heikkilä. 2015. Predicting the valence of a scene from observers' eye movements. *PLoS ONE*, 10(9): 1–19. DOI: 10.1371/journal.pone.0138198. 237

G. W. Taylor, R. Fergus, Y. LeCun, and C. Bregler. 2010. Convolutional learning of spatio-temporal features. In *ECCV*, pp. 140–153. 13

TeamViewer. July 2012. TeamViewer web page. http://www.teamviewer.com. DOI: 10.1007/978-3-642-15567-3_11. 291

J. Teevan, S. T. Dumais, and E. Horvitz. 2005. Personalizing search via automated analysis of interests and activities. In *Proceedings of the 28th Annual International ACM SIGIR Conference on Research and Development in Information Retrieval*, pp. 449–456. ACM. DOI: 10.1145/1076034.1076111. 147

J. Teevan, S. T. Dumais, and D. J. Liebling. 2008. To personalize or not to personalize: modeling queries with variation in user intent. In *Proceedings of the 31st Annual International ACM SIGIR Conference on Research and Development in Information Retrieval*, pp. 163–170. ACM. DOI: 10.1145/1390334.1390364. 140

A. Temko, R. Malkin, C. Zieger, D. Macho, C. Nadeu, and M. Omologo. 2006. Acoustic event detection and classification in smart-room environments: Evaluation of CHIL project systems. *IV Jornadas en Tecnologia del Habla*, 65(48): 5. 38

J. Thomason, S. Venugopalan, S. Guadarrama, K. Saenko, and R. Mooney. 2014. Integrating language and vision to generate natural language descriptions of videos in the wild. In *Proceedings of the 25th International Conference on Computational Linguistics*, pp. 1218–1227. 28

B. Thomee, D. A. Shamma, B. Elizalde, G. Friedland, K. Ni, D. Poland, D. Borth, and L. Li. 2016. YFCC100M: The new data in multimedia research. *Communications of the ACM*, 59(2): 64–73. 39

Y. Tian, J. Srivastava, T. Huang, and N. Contractor. 2010. Social multimedia computing. *Computer*, 43(8): 27–36. DOI: 10.1109/MC.2010.188. 155

S. Tok, M. Koyuncu, S. Dural, and F. Catikkas. 2010. Evaluation of International Affective Picture System (IAPS) ratings in an athlete population and its relations to personality. *Personality and Individual Differences*, 49(5): 461–466. DOI: 10.1016/j.paid.2010.04.020. 244, 246

I. M. Toke. 2011. An introduction to Hawkes processes with applications to finance. Lectures Notes from Ecole Centrale Paris, BNP Paribas Chair of Quantitative Finance. 193

A. Torabi, C. Pal, H. Larochelle, and A. Courville. 2015. Using descriptive video services to create a large data source for video annotation research. *arXiv:1503.01070*. 26, 27

A. Torralba, R. Fergus, and Y. Weiss. June 2008. Small codes and large databases for recognition. In *2008 IEEE Conference on Computer Vision and Pattern Recognition,* pp. 1–8. DOI: 10.1109/CVPR.2008.4587633. 120

W. A. A. Torres, N. Bhattacharjee, and B. Srinivasan. 2014. Effectiveness of fully homomorphic encryption to preserve the privacy of biometric data. In *Proceedings of the 16th International Conference on Information Integration and Web-based Applications & Services*, pp. 152–158. ACM. DOI: 10.1145/2684200.2684296. 82

D. Tran, L. D. Bourdev, R. Fergus, L. Torresani, and M. Paluri. 2015. C3D: Generic features for video analysis. In *ICCV*, 2(7): 8. 10, 16, 24

J. R. Troncoso-Pastoriza, S. Katzenbeisser, and M. Celik. 2007. Privacy preserving error resilient DNA searching through oblivious automata. In *Proceedings of the 14th ACM Conference on Computer and Communications Security*, pp. 519–528. ACM. DOI: 10.1145/1315245.1315309. 83

B. L. Tseng, C.-Y. Lin, and J. R. Smith. 2004. Using MPEG-7 and MPEG-21 for personalizing video. *MultiMedia, IEEE*, 11(1): 42–52. DOI: 10.1109/MMUL.2004.1261105. 157

V. Tudor, M. Almgren, and M. Papatriantafilou. 2015. Harnessing the unknown in advanced metering infrastructure traffic. In *Proceedings of the 30th Annual ACM Symposium on Applied Computing*, pp. 2204–2211. ACM. DOI: 10.1145/2695664.2695725. 103

S. Tulyakov, X. Alameda-Pineda, E. Ricci, L. Yin, J. F. Cohn, and N. Sebe. 2016. Self-adaptive matrix completion for heart rate estimation from face videos under realistic conditions. In *CVPR*, pp. 2396–2404. DOI: 10.1109/CVPR.2016.263. 58

P. Turaga, R. Chellappa, V. S. Subrahmanian, and O. Udrea. 2008. Machine recognition of human activities: A survey. *IEEE TPAMI*, 18(11): 1473–1488. DOI: 10.1109/TCSVT.2008.2005594. 4

T. Tuytelaars and C. Schmid. Oct. 2007. Vector quantizing feature space with a regular lattice. In *ICCV*, pp. 1–8. DOI: 10.1109/ICCV.2007.4408924. 116

G. Tzanetakis and P. Cook. 2002. Musical genre classification of audio signals. *IEEE transactions on Speech and Audio Processing*, 10(5): 293–302. DOI: 10.1109/TSA.2002.800560. 37

Ubitus. January 2015. Ubitus web page. http://www.ubitus.net. 287

A. Ulges, M. Koch, and D. Borth. 2012. Linking visual concept detection with viewer demographics. In *Proceedings of the 2nd ACM International Conference on Multimedia Retrieval*, p. 24. ACM. DOI: 10.1145/2324796.2324827. 147

J. Ullman. 1983. *Principles of database systems*. W.H. Freeman & Co. New York. 175

UltraVNC. July 2012. UltraVNC web page. http://www.uvnc.com/. 291

M. Upmanyu. 2010. *Efficient Privacy Preserving Protocols for Visual Computation*. Master's thesis, IIIT Hyderabad, India. 82, 93, 96, 102

M. Upmanyu, A. M. Namboodiri, K. Srinathan, and C. V. Jawahar. 2009. Efficient privacy preserving video surveillance. In *Proceedings of IEEE* $12^{th}$ *International Conference on Computer Vision*, pp. 1639–1646. DOI: 10.1109/ICCV.2009.5459370. 82, 91, 93, 96

R. Valenti, N. Sebe, and T. Gevers. 2009. Image saliency by isocentric curvedness and color. In *ICCV*, pp. 2185–2192. DOI: 10.1109/ICCV.2009.5459240. 222

G. Valenzise, L. Gerosa, M. Tagliasacchi, F. Antonacci, and A. Sarti. 2007. Scream and gunshot detection and localization for audio-surveillance systems. In *IEEE Conference on Advanced Video and Signal Based Surveillance*, pp. 21–26. IEEE. DOI: 10.1109/AVSS.2007.4425280. 38

H. Van, F. Tran, and J. Menaud. 2009. SLA-aware virtual resource management for cloud infrastructures. In *Proceedings of IEEE International Conference on Computer and Information Technology (CIT)*, pp. 357–362. DOI: 10.1109/CIT.2009.109. 260

L. Vaquero and L. Merino. 2014. Finding your way in the fog: Towards a comprehensive definition of fog computing. *ACM SIGCOMM Computer Communication Review*, 44(5): 27–32. DOI: 10.1145/2677046.2677052. 257

A. Vardy and Y. Be'ery. July 1993. Maximum likelihood decoding of the leech lattice. *IEEE Trans. Inform. Theory*, 39(4): 1435–1444. DOI: 10.1109/18.243466. 115

A. Vedaldi and A. Zisserman. March 2012. Efficient additive kernels via explicit feature maps. *IEEE Trans. PAMI*, 34: 480–492. DOI: 10.1109/CVPR.2010.5539949. 133

R. Vedantam, C. Lawrence Zitnick, and D. Parikh. 2015. CIDEr: Consensus-based image description evaluation. In *CVPR*, pp. 4566–4575. DOI: 10.1109/CVPR.2015.7299087. 27

A. Veen and F. P. Schoenberg. 2008. Estimation of space-time branching process models in seismology using an EM-type algorithm. *Journal of the American Statistical Association*, 103(482): 614–624. DOI: 10.1198/016214508000000148. 200

V. Veeriah, N. Zhuang, and G.-J. Qi. 2015. Differential recurrent neural networks for action recognition. In *ICCV*, pp. 4041–4049. 12

S. Venugopalan, M. Rohrbach, J. Donahue, R. Mooney, T. Darrell, and K. Saenko. 2015a. Sequence to sequence—video to text. In *ICCV*, pp. 4534–4542. DOI: 10.1109/ICCV.2015.515. 16, 17, 28

S. Venugopalan, H. Xu, J. Donahue, M. Rohrbach, R. Mooney, and K. Saenko. 2015b. Translating videos to natural language using deep recurrent neural networks. In *Proceedings of the 2015 Conference of the North American Chapter of the Association for Computational Linguistics–Human Language Technologies (NAACL HLT)*, pp. 1494–1504. 16, 17, 28

S. Venugopalan, L. A. Hendricks, R. Mooney, and K. Saenko. 2016. Improving LSTM-based video description with linguistic knowledge mined from text. *arXiv:1604.01729*. DOI: 10.18653/v1/D16-1204. 28

T. Verbelen, S. Pieter, T. Filip, and D. Bart. 2012. Cloudlets: Bringing the cloud to the mobile user. In *Proceedings of ACM Workshop on Mobile Cloud Computing and Services (MCS)*, pp. 29–36. DOI: 10.1145/2307849.2307858. 259

O. Verscheure, M. Vlachos, A. Anagnostopoulos, P. Frossard, E. Bouillet, and P. S. Yu. 2006. Finding "who is talking to whom" in VoiP networks via progressive stream clustering.

In *Proceedings of IEEE Sixth International Conference on Data Mining*, pp. 667–677. DOI: 10.1109/ICDM.2006.72. 97, 98

VideoLAN. VLC media player. Official page for VLC media player, the Open Source video framework. http://www.videolan.org/vlc/. 293

C. Viedma. 2010. Mobile web mashups. www.mobilemashups.com 171

A. Vinciarelli, A. Dielmann, S. Favre, and H. Salamin. 2009. Canal9: A database of political debates for analysis of social interactions. In *Proceedings of the International Conference on Affective Computing and Intelligent Interaction (ICACII)*, September 2009, pp. 1–4. DOI: 10.1109/ACII.2009.5349466. 55

O. Vinyals, A. Toshev, S. Bengio, and D. Erhan. 2015. Show and tell: A neural image caption generator. In *CVPR*, pp. 3156–3164. DOI: 10.1109/TPAMI.2016.2587640. 15

P. Viola and M. J. Jones. 2004. Robust real-time face detection. *IJCV*, 57(2): 137–154. DOI: 10.1023/B:VISI.0000013087.49260.fb. 225

T. Virtanen, J. F. Gemmeke, B. Raj, and P. Smaragdis. 2015. Compositional models for audio processing: Uncovering the structure of sound mixtures. *Signal Processing Magazine, IEEE*, 32(2): 125–144. DOI: 10.1109/MSP.2013.2288990. 42

M. Voit and R. Stiefelhagen. 2010. 3D user-perspective, voxel-based estimation of visual focus of attention in dynamic meeting scenarios. In *ICMI*, DOI: 10.1145/1891903.1891966. 57

L. von Ahn and L. Dabbish. 2004. Labeling images with a computer game. In *ACM Conference on Human Factors in Computing Systems*, pp. 319–326. DOI: 10.1145/985692.985733. 220

J. Vuurens and A. Vries. 2012. Obtaining high-quality relevance judgments using crowdsourcing. *IEEE Transactions on Internet Computing*, 16(5): 20–27. DOI: 10.1109/MIC.2012.71. 262

J. Wache, R. Subramanian, M. K. Abadi, R. L. Vieriu, S. Winkler, and N. Sebe. 2015. Implicit Personality Profiling Based on Psycho–Physiological Responses to Emotional Videos. In *Int'l Conference on Multimodal Interaction*, pp. 239–246. DOI: 10.1145/2818346.2820736. 221, 237, 238, 239, 241, 245, 247, 249

A. Wächter and L. T. Biegler. 2006. On the implementation of a primal-dual interior point filter line search algorithm for large-scale nonlinear programming. *Mathematical Programming*, 106(1): 25–57. DOI: 10.1007/s10107-004-0559-y. 211

K. Walby. 2005. Open-street camera surveillance and governance in Canada. *Canadian Journal of Criminology and Criminal Justice/La Revue canadienne de criminologie et de justice pénale*, 47(4): 655–684. DOI: 10.3138/cjccj.47.4.655. 91

O. Walter, R. Haeb-Umbach, S. Chaudhuri, and B. Raj. 2013. Unsupervised word discovery from phonetic input using nested Pitman-Yor language modeling. In *Proceedings of IEEE International Conference on Robotics and Automation (ICRA) Workshop on Autonomous Learning*. 47

L. Wan, M. Zeiler, S. Zhang, Y. L. Cun, and R. Fergus. 2013. Regularization of neural networks using DropConnect. In *ICML*, 28(3): 1058–1066. 6

C. Wang, K. Ren, W. Lou, and J. Li. 2010. Toward publicly auditable secure cloud data storage services. *IEEE Network*, 24(4): 19–24. DOI: 10.1109/MNET.2010.5510914. 260

C. Wang, S. Chow, Q. Wang, K. Ren, and W. Lou. 2013. Privacy-preserving public auditing for secure cloud storage. *IEEE Transactions on Computers*, 62(2): 362–375. 260

D. Wang and G. J. Brown (eds.). 2006. *Computational Auditory Scene Analysis: Principles, Algorithms, and Applications*, volume 147. Wiley Interscience. 36

H. Wang and C. Schmid. 2013. Action recognition with improved trajectories. In *ICCV*, pp. 3551–3558. DOI: 10.1109/ICCV.2013.441. 10, 11

J. Wang, S. Kumar, and S.-F. Chang. 2012a. Semi-supervised hashing for large-scale search. *IEEE Trans. PAMI*, 34(12): 2393–2406. DOI: 10.1109/TPAMI.2012.48. 113

J. Wang, H. T. Shen, J. Song, and J. Ji. 2014. Hashing for similarity search: A survey. *arXiv:1408.2927*. 120

J. Wang, W. Liu, S. Kumar, and S.-F. Chang. Jan. 2016a. Learning to hash for indexing big data—A survey. *Proceedings of the IEEE*, 104(1): 34–57. DOI: 10.1109/JPROC.2015.2487976. 120

L. Wang, Y. Qiao, and X. Tang. 2015. Action recognition with trajectory-pooled deep-convolutional descriptors. In *CVPR*, pp. 4305–4314. DOI: 10.1109/CVPR.2015.7299059. 11

L. Wang, Y. Xiong, Z. Wang, Y. Qiao, D. Lin, X. Tang, and L. Van Gool. 2016b. Temporal segment networks: Towards good practices for deep action recognition. In *ECCV*, pp. 20–36. DOI: 10.1007/978-3-319-46484-8_2. 11, 24

S. Wang and S. Dey. Nov. 2009. Modeling and characterizing user experience in a cloud server based mobile gaming approach. In *Global Telecommunications Conference, 2009. GLOBECOM 2009. IEEE*, pp. 1–7. http://ieeexplore.ieee.org/lpdocs/epic03/wrapper.htm?arnumber=5425784. DOI: 10.1109/GLOCOM.2009.5425784. 220, 309

S. Wang and S. Dey. Apr. 2010a. Addressing response time and video quality in remote server based internet mobile gaming. *2010 IEEE Wireless Communication and Networking Conference*, 5: 1–6. http://ieeexplore.ieee.org/lpdocs/epic03/wrapper.htm?arnumber=5506572. DOI: 10.1109/WCNC.2010.5506572. 309

S. Wang and S. Dey. Dec. 2010b. Rendering adaptation to address communication and computation constraints in cloud mobile gaming. In *Global Telecommunications Conference (GLOBECOM 2010), IEEE*, pp. 1–6. http://ieeexplore.ieee.org/lpdocs/epic03/wrapper.htm?arnumber=5684144. DOI: 10.1109/GLOCOM.2010.5684144. 309

S. Wang and S. Dey. 2013. Adaptive mobile cloud computing to enable rich mobile multimedia applications. *IEEE Transactions on Multimedia*, 15(4): 870–883. DOI: 10.1109/TMM.2013.2240674. 311

X. Wang, S. Chen, and S. Jajodia. 2005. Tracking anonymous peer-to-peer VoiP calls on the internet. In *Proceedings of the ACM* $12^{th}$ *Conference on Computer and Communications Security*, pp. 81–91. 97, 98

X. Wang, A. Farhadi, and A. Gupta. 2016c. Actions $\sim$ transformations. In *CVPR*, pp. 2658–2667. 11, 24

Y. Wang. 2004. An FSM model for situation-aware mobile application software systems. In *ACM-SE 42: Proceedings of the 42nd Annual Southeast Regional Conference*, pp. 52–57. ISBN 1-58113-870-9. DOI: 10.1145/1102120.1102133. 165

Y. Wang and M. S. Kankanhalli. 2015. Tweeting cameras for event detection. In *Proceedings of the 24th International Conference on World Wide Web*, pp. 1231–1241. ACM. DOI: 10.1145/2736277.2741634. 185

Y. Wang and G. Mori. 2009. Human action recognition by semilatent topic models. *IEEE Transactions on Pattern Analysis and Machine Intelligence*, 31(10): 1762–1774. DOI: 10.1109/TPAMI.2009.43. 187

Y. Wang, C. von der Weth, Y. Zhang, K. H. Low, V. K. Singh, and M. Kankanhalli. 2016d. Concept based hybrid fusion of multimodal event signals. In *2016 IEEE International Symposium on Multimedia (ISM)*, pp. 14–19. IEEE. DOI: 10.1109/TPAMI.2009.43. 185

Z. Wang, L. Sun, X. Chen, W. Zhu, J. Liu, M. Chen, and S. Yang. 2012b. Propagation-based social-aware replication for social video contents. In *Proceedings of the 20th ACM International Conference on Multimedia*, pp. 29–38. ACM. DOI: 10.1145/2393347.2393359. 153

R. Weber and H. Blott. 1997. An approximation based data structure for similarity search. Technical report, ESPRIT Project HERMES. 109, 110

R. Weber, H.-J. Schek, and S. Blott. 1998. A quantitative analysis and performance study for similarity-search methods in high-dimensional spaces. In *Proceedings of the International Conference on Very Large DataBases*, pp. 194–205. 107, 109, 110

WebM. April 2013. The WebM project web page. http://www.webmproject.org. 292

K. Q. Weinberger and L. K. Saul. 2009. Distance metric learning for large margin nearest neighbor classification. *Journal of Machine Learning Research*, 10(Feb): 207–244. 149

C. Weinhardt, A. Anandasivam, B. Blau, N. Borissov, T. Meinl, W. Michalk, and J. Stober. 2009. Cloud computing—A classification, business models, and research directions. *Business and Information Systems Engineering*, 1(5): 391–399. DOI: 10.1007/s12599-009-0071-2. 259

M. Weintraub. 1985. *A Theory and Computational Model of Auditory Monoaural Sound Separation*. PhD thesis, Department of Electrical Engineering, Stanford University. 37

Y. Weiss, A. Torralba, and R. Fergus. Dec. 2009. Spectral hashing. In *NIPS*, pp. 1753–1760. 110, 112, 122, 124

J.Wen, M. Severa,W. Zeng, M. Luttrell, and W. Jin. 2001. A format-compliant configurable encryption framework for access control of multimedia. In *2001 IEEE Fourth Workshop*

*on Multimedia Signal Processing*, pp. 435–440. DOI: 10.1109/MMSP.2001.962772. 92, 93

J. Wen, M. Severa, W. Zeng, M. H. Luttrell, and W. Jin. 2002. A format-compliant configurable encryption framework for access control of video. *IEEE Transactions on Circuits and Systems for Video Technology*, 12(6): 545–557. DOI: 10.1109/TCSVT.2002.800321. 92, 93

P. Wieschollek, O. Wang, and A. Sorkine-Hornung. June 2016. Efficient large-scale approximate nearest neighbor search on the GPU. In *CVPR*, pp. 2027–2035. DOI: 10.1109/CVPR.2016.223. 111

D. Willis, A. Dasgupta, and S. Banerjee. 2014. ParaDrop: A multi-tenant platform for dynamically installed third party services on home gateways. In *Proceedings of ACM SIGCOMM Workshop on Distributed Cloud Computing (DCC)*, pp. 43–48. DOI: 10.1145/2645892.2645901. 259

J. G. Wilpon, L. R. Rabiner, C.-H. Lee, and E. R. Goldman. 1990. Automatic recognition of keywords in unconstrained speech using hidden Markov models. *Proceedings of IEEE Transactions on Acoustics, Speech and Signal Processing*, 38(11): 1870–1878. DOI: 10.1109/29.103088. 95, 98

K. W. Wilson and B. Raj. 2010. Spectrogram dimensionality reduction with independence constraints. In *ICASSP*, pp. 1938–1941. DOI: 10.1109/ICASSP.2010.5495308. 37

K. W. Wilson, B. Raj, P. Smaragdis, and A. Divakaran. 2008. Speech denoising using nonnegative matrix factorization with priors. In *ICASSP*, pp. 4029–4032. DOI: 10.1109/ICASSP.2008.4518538. 36

C. V. Wright, L. Ballard, F. Monrose, and G. M. Masson. 2007. Language identification of encrypted VoiP traffic: Alejandra y Roberto or Alice and Bob. In *Proceedings of the* $16^{th}$ *USENIX Security Symposium*, pp. 1–12. 97, 98

C. V. Wright, L. Ballard, S. E. Coull, F. Monrose, and G. M. Masson. 2008. Spot me if you can: Uncovering spoken phrases in encrypted VoiP conversations. In *IEEE Symposium on Security and Privacy*, pp. 35–49. DOI: 10.1109/SP.2008.21. 97, 98

B. Wu, S. Lyu, B.-G. Hu, and Q. Ji. 2015a. Multi-label learning with missing labels for image annotation and facial action unit recognition. *Pattern Recognition*, 48(7): 2279–2289. DOI: 10.1016/j.patcog.2015.01.022. 58

F. Wu and B. A. Huberman. Nov. 2007. Novelty and collective attention. *PNAS '07*, 104(45): 17599–17601. http://www.pnas.org/content/104/45/17599.abstract. DOI: 10.1073/pnas.0704916104. 210

J. Wu, C. Yuen, N. Cheung, J. Chen, and C. W. Chen. Dec. 2015b. Enabling adaptive high-frame-rate video streaming in mobile cloud gaming applications. *IEEE Transactions on Circuits and Systems for Video Technology*, 25(12): 1988–2001. DOI: 10.1109/TCSVT.2015.2441412. 309

L. Wu, S. Garg, and R. Buyya. 2011. SLA-based resource allocation for Software as a Service Provider (SaaS) in cloud computing environments. In *Proceedings of IEEE/ACM*

*International Symposium on Cluster, Cloud and Grid Computing (CCGRID)*, pp.195–204. DOI: 10.1109/CCGrid.2011.51. 260

L. Wu, R. Jin, and A. K. Jain. 2013. Tag completion for image retrieval. *IEEE TPAMI*, 35(3): 716–727. DOI: 10.1109/TPAMI.2012.124. 58

Z. Wu. 2014. Gaming in the cloud: One of the future entertainment. In *Interactive Multimedia Conference*, pp. 1–6. 290, 291

Z. Wu, X. Wang, Y.-G. Jiang, H. Ye, and X. Xue. 2015c. Modeling spatial-temporal clues in a hybrid deep learning framework for video classification. In *ACM Multimedia*, pp. 461–470. DOI: 10.1145/2733373.2806222. 11, 12

Z. Wu, Y.-G. Jiang, X. Wang, H. Ye, and X. Xue. 2016. Multi-stream multi-class fusion of deep networks for video classification. In *ACM Multimedia*, pp. 791–800. DOI: 10.1145/2964284.2964328. 7, 12, 24

x264. July 2012. x264 web page. http://www.videolan.org/developers/x264.html. 292

L. Xie, A. Natsev, J. R. Kender, M. Hill, and J. R. Smith. 2011. Visual memes in social media: Tracking real-world news in YouTube videos. In *Proceedings of the 19th ACM International Conference on Multimedia*, pp. 53–62. ACM. DOI: 10.1145/2072298.2072307. 156

X. Xie, H. Liu, S. Goumaz, and W.-Y. Ma. 2005. Learning user interest for image browsing on small-form-factor devices. In *Proceedings of the SIGCHI Conference on Human Factors in Computing Systems*, pp. 671–680. ACM. DOI: 10.1145/1054972.1055065. 148

X. Xiong and F. De La Torre. 2013. Supervised descent method and its applications to face alignment. In *CVPR*, pp. 532–539. DOI: 10.1109/CVPR.2013.75. 225

Z. Xiong, R. Radhakrishnan, and A. Divakaran. 2003. Generation of sports highlights using motion activity in combination with a common audio feature extraction framework. In *Proceedings of the 2003 International Conference on Image Processing*, volume 1, pp. I–5. IEEE. DOI: 10.1109/ICIP.2003.1246884. 38

H. Xu, J. Wang, Z. Li, G. Zeng, S. Li, and N. Yu. November 2011. Complementary hashing for approximate nearest neighbor search. In *ICCV*, pp. 1631–1638. DOI: 10.1109/ICCV.2011.6126424. 118

H. Xu, S. Venugopalan, V. Ramanishka, M. Rohrbach, and K. Saenko. 2015a. A multi-scale multiple instance video description network. *arXiv:1505.05914*. 28

J. Xu, E.-C. Chang, and J. Zhou. 2013. Weak leakage-resilient client-side deduplication of encrypted data in cloud storage. In *Proceedings of the 8th ACM SIGSAC Symposium on Information, Computer and Communications Security*, pp. 195–206. ACM. DOI: 10.1109/ICCV.2011.6126424. 95

J. Xu, T. Mei, T. Yao, and Y. Rui. 2016. MSR-VTT: A large video description dataset for bridging video and language. In *CVPR*, pp. 5288–5296. 15, 16, 27

K. Xu, M. Song, X. Zhang, and J. Song. 2009. A cloud computing platform based on P2P. In *Proceedings of IEEE International Symposium on IT in Medicine and Education (ITIME)*, pp. 1–4. DOI: 10.1109/ITIME.2009.5236386. 259

L. Xu, X. Guo, Y. Lu, S. Li, O. C. Au, and L. Fang. Jul. 2014. A low latency cloud gaming system using edge preserved image homography. In *2014 IEEE International Conference on Multimedia and Expo (ICME)*, pp. 1–6. DOI: 10.1109/ICME.2014.6890279. 307

M. Xu, N. Maddage, C. Xu, M. Kankanhalli, and Q. Tian. 2003. Creating audio keywords for event detection in soccer video. In *Proceedings of the 2003 International Conference on Multimedia and Expo*, volume 2, pp. II–281. IEEE. DOI: 10.1109/ICME.2003.1221608. 38

R. Xu, C. Xiong, W. Chen, and J. J. Corso. 2015b. Jointly modeling deep video and compositional text to bridge vision and language in a unified framework. In *AAAI*, pp. 2346–2352. 16, 17

Z. Xu, Y. Yang, and A. G. Hauptmann. 2015c. A discriminative CNN video representation for event detection. In *CVPR*, pp. 1798–1807. DOI: 10.1109/CVPR.2015.7298789. 9

K. Yadati, H. Katti, and M. Kankanhalli. 2014. CAVVA: Computational affective video-in-video advertising. *IEEE Transactions on Multimedia*, 16(1): 15–23. DOI: 10.1109/TMM.2013.2282128. 237

M. A. Yakubu, P. K. Atrey, and N. C. Maddage. 2015. Secure audio reverberation over cloud. In *10th Annual Symposium on Information Assurance (ASIA '15)*, p. 39. 99, 100

Y. Yan, E. Ricci, R. Subramanian, O. Lanz, and N. Sebe. 2013. No matter where you are: Flexible graph-guided multi-task learning for multi-view head pose classification under target motion. In *IEEE ICCV*, pp. 1177–1184. DOI: 10.1109/ICCV.2013.150. 57, 71

Y. Yan, E. Ricci, R. Subramanian, G. Liu, and N. Sebe. 2014. Multitask linear discriminant analysis for view invariant action recognition. *IEEE Transactions on Image Processing (TIP)*, 23(12): 5599–5611. DOI: 10.1109/TIP.2014.2365699. 56

J. Yang, K. Yu, Y. Gong, and T. Huang. 2009. Linear spatial pyramid matching using sparse coding for image classification. In *CVPR*, 1794–1801. DOI: 10.1109/CVPR.2009.5206757. 233, 234, 235

L. Yang and R. Jin. 2006. Distance metric learning: A comprehensive survey. Michigan State University, 2(2): 78. 149

Q. Yang, M. J. Wooldridge, and H. Zha. 2015. Trailer generation via a point process-based visual attractiveness model. In *Proceedings of the 24th International Conference on Artificial Intelligence*, pp. 2198–2204. AAAI Press. ISBN 9781577357384. 193, 200

Y. Yang, P. Cui, W. Zhu, and S. Yang. 2013. User interest and social influence based emotion prediction for individuals. In *Proceedings of the 21st ACM International Conference on Multimedia*, pp. 785–788. ACM. DOI: 10.1145/2502081.2502204. 148, 150

A. C.-C. Yao. July 1981. Should tables be sorted? *Journal of the ACM*, 28(3): 615–628. DOI: 10.1145/322261.322274. 108, 113

L. Yao, A. Torabi, K. Cho, N. Ballas, C. Pal, H. Larochelle, and A. Courville. 2015a. Describing videos by exploiting temporal structure. In *ICCV*, pp. 4507–4515. DOI: 10.1109/ICCV.2015.512. 12, 16, 17, 28

T. Yao, C.-W. Ngo, and S. Zhu. 2012. Predicting domain adaptivity: Redo or recycle? In *ACM Multimedia*, pp. 821–824. DOI: 10.1109/ICCV.2015.512. 17

T. Yao, T. Mei, C.-W. Ngo, and S. Li. 2013. Annotation for free: Video tagging by mining user search behavior. In *ACM Multimedia*, pp. 977–986. DOI: 10.1145/2502081.2502085. 15

T. Yao, T. Mei, and C.-W. Ngo. 2015b. Learning query and image similarities with ranking canonical correlation analysis. In *ICCV*, pp. 28–36. DOI: 10.1109/ICCV.2015.12. 17

T. Yao, Y. Pan, C.-W. Ngo, H. Li, and T. Mei. 2015c. Semi-supervised domain adaptation with subspace learning for visual recognition. In *CVPR*, pp. 2142–2150. DOI: 10.1109/CVPR.2015.7298826. 17

T. Yao, Y. Pan, Y. Li, Z. Qiu, and T. Mei. 2016. Boosting image captioning with attributes. *arXiv:1611.01646*. 17

K. Yatani and K. N. Truong. 2012. Bodyscope: A wearable acoustic sensor for activity recognition. In *Proceedings of the ACM Conference on Ubiquitous Computing*, pp. 341–350. ACM. DOI: 10.1145/2370216.2370269. 52

S. Yau and J. Liu. 2006. Hierarchical situation modeling and reasoning for pervasive computing. In *The Fourth IEEE Workshop on Software Technologies for Future Embedded and Ubiquitous Systems, 2006, and the 2006 Second International Workshop on Collaborative Computing, Integration, and Assurance*, pp. 5–10. IEEE. DOI: 10.1109/SEUS-WCCIA.2006.25. 165, 167

G. Ye, Y. Li, H. Xu, D. Liu, and S.-F. Chang. 2015a. Eventnet: A large scale structured concept library for complex event detection in video. In *ACM Multimedia*, pp. 471–480. DOI: 10.1145/2733373.2806221. 22

H. Ye, Z. Wu, R.-W. Zhao, X. Wang, Y.-G. Jiang, and X. Xue. 2015b. Evaluating two-stream CNN for video classification. In *ACM ICMR*, pp. 435–442. DOI: 10.1145/2671188.2749406. 10

S. Yi, Z. Hao, Z. Qin, and Q. Li. 2015. Fog computing: Platform and applications. In *Proceedings of IEEE Workshop on Hot Topics in Web Systems and Technologies (HotWeb)*, pp. 73–78. DOI: 10.1109/HotWeb.2015.22. 257

R. Yogachandran, R. Phan, J. Chambers, and D. Parish. 2012. Facial expression recognition in the encrypted domain based on local Fisher discriminant analysis. In *Proceedings of the IEEE Transactions on Affective Computing*, pp. 83–92. DOI: 10.1109/T-AFFC.2012.33. 88, 90

H. Yu, J. Wang, Z. Huang, Y. Yang, and W. Xu. 2016. Video paragraph captioning using hierarchical recurrent neural networks. In *CVPR*, pp. 4584–4593. 15, 16, 28, 29

Z. Yuan, J. Sang, Y. Liu, and C. Xu. 2013. Latent feature learning in social media network. In *Proceedings of the 21st ACM International Conference on Multimedia*, pp. 253–262. ACM. DOI: 10.1145/2502081.2502284. 148

M. Yuen, I. King, and K. Leung. 2011. A survey of crowdsourcing systems. In *Proceedings of IEEE International Conference on Social Computing (SocialCom)*, pp. 739–773. DOI: 10.1109/PASSAT/SocialCom.2011.203. 257

K. Yun, Y. Peng, D. Samaras, G. J. Zelinsky, and T. L. Berg. 2013. Studying relationships between human gaze, description, and computer vision. In *CVPR*, pp. 739–746. DOI: 10.1109/CVPR.2013.101. 220, 226

G. Zen, B. Lepri, E. Ricci, and O. Lanz. 2010. Space speaks: Towards socially and personality aware visual surveillance. In *ACM International Workshop on Multimodal Pervasive Video Analysis*, pp. 37–42. DOI: 10.1145/1878039.1878048. 238, 240

W. Zeng and S. Lei. 1999. Efficient frequency domain video scrambling for content access control. In *Proceedings of the Seventh ACM International Conference on Multimedia (Part 1)*, pp. 285–294. DOI: 10.1145/319463.319627. 92, 93

W. Zeng and S. Lei. 2003. Efficient frequency domain selective scrambling of digital video. *IEEE Transactions on Multimedia*, 5(1): 118–129. DOI: 10.1109/TMM.2003.808817. 92, 93

S. Zha, F. Luisier, W. Andrews, N. Srivastava, and R. Salakhutdinov. 2015. Exploiting image-trained CNN architectures for unconstrained video classification. In *BMVC*, 60: 1–13. DOI: 10.5244/C.29.60. 9, 24

B. Zhang, L. Wang, Z. Wang, Y. Qiao, and H. Wang. 2016. Real-time action recognition with enhanced motion vector CNNs. In *CVPR*, pp. 2718–2726. 11

H. Zhang, Z.-J. Zha, S. Yan, J. Bian, and T.-S. Chua. 2012. Attribute feedback. In *Proceedings of the 20th ACM International Conference on Multimedia*, pp. 79–88. ACM. DOI: 10.1145/2393347.2393365. 148

T. Zhang and C.-C. J. Kuo. 2001. Audio content analysis for online audiovisual data segmentation and classification. *IEEE Transactions on Speech and Audio Processing*, 9(4): 441–457. DOI: 10.1109/89.917689. 38

T. Zhang, C. Du, and J. Wang. June 2014. Composite quantization for approximate nearest neighbor search. In *ICML*, pp. 838–846. DOI: 10.1109/ICCV.2011.6126424. 131

T. Zhang, G.-J. Qi, J. Tang, and J. Wang. June 2015a. Sparse composite quantization. In *CVPR*, pp. 4548–4556. DOI: 10.1109/CVPR.2015.7299085. 131

Y.-Q. Zhang, W.-L. Zheng, and B.-L. Lu. 2015b. *Transfer Components Between Subjects for EEG-based Driving Fatigue Detection*, pp. 61–68. Springer. DOI: 10.1007/978-3-319-26561-2_8. 250

Z. Zhao, K. Hwang, and J. Villeta. 2012. Game cloud design with virtualized CPU/GPU servers and initial performance results. In *Proceedings of the 3rd Workshop on Scientific Cloud Computing*, ScienceCloud ’12, pp. 23–30. ACM, New York. DOI: 10.1145/2287036.2287042. 301

Y. Zheng, X. Yuan, X. Wang, J. Jiang, C. Wang, and X. Gui. 2015. Enabling encrypted cloud media center with secure deduplication. In *Proceedings of the 10th ACM Symposium on*

*Information, Computer and Communications Security*, pp. 63–72. ACM. DOI: 10.1145/2714576.2714628. 94, 96

E. Zhong, B. Tan, K. Mo, and Q. Yang. 2013. User demographics prediction based on mobile data. *Pervasive and Mobile Computing*, pp. 823–837. DOI: 10.1016/j.pmcj.2013.07.009. 147

M. Zhou, R. Zhang, W. Xie, W. Qian, and A. Zhou. 2010. Security and privacy in cloud computing: A survey. In *Proceedings of IEEE International Conference on Semantics Knowledge and Grid (SKG)*, pp. 105–112. DOI: 10.1109/SKG.2010.19. 260

C. Zhu, R. H. Byrd, P. Lu, and J. Nocedal. 1997. Algorithm 778: L-BFGS-B: Fortran subroutines for large-scale bound-constrained optimization. *ACM Transactions on Mathematical Software (TOMS)*, 23(4): 550–560. DOI: 10.1145/279232.279236. 207

W. Zhu, J. Hu, G. Sun, X. Cao, and Y. Qiao. 2016. A key volume mining deep framework for action recognition. In *CVPR*, pp. 1991–1999. DOI: 10.1109/CVPR.2016.219. 12, 24

X. Zhu and A. B. Goldberg. 2009. Introduction to semi-supervised learning. *Synthesis Lectures on Artificial Intelligence and Machine Learning*. Morgan and Claypool Publishers. 72

X. Zhuang, S. Tsakalidis, S. Wu, P. Natarajan, R. Prasad, and P. Natarajan. 2011. Compact audio representation for event detection in consumer media. In *Proceedings of Interspeech*, pp. 2089–2092. 46

J. R. Zipkin, F. P. Schoenberg, K. Coronges, and A. L. Bertozzi. 2016. Point-process models of social network interactions: Parameter estimation and missing data recovery. *European Journal of Applied Mathematics*, 27: 502–529. DOI: 10.1017/S0956792515000492. 209

# Index

# Editor Biography

## Shih-Fu Chang

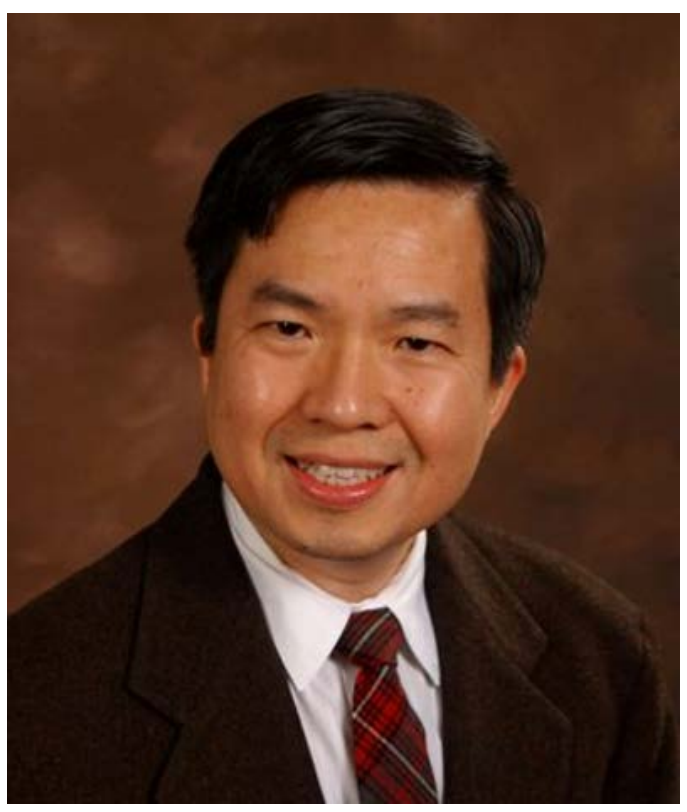

**Shih-Fu Chang** is the Richard Dicker Professor at Columbia University, with appointments in both Electrical Engineering Department and Computer Science Department. His research is focused on multimedia information retrieval, computer vision, machine learning, and signal processing. A primary goal of his work is to develop intelligent systems that can extract rich information from the vast amount of visual data such as those emerging on the Web, collected through pervasive sensing, or available in gigantic archives. His work on content-based visual search in the early 90s—VisualSEEk and VideoQ—set the foundation of this vibrant area. Over the years, he continued to develop innovative solutions for image/video recognition, multimodal analysis, visual content ontology, image authentication, and compact hashing for large-scale indexing. His work has had major impacts in various applications like image/video search engines, online crime prevention, mobile product search, AR/VR, and brain machine interfaces.

His scholarly work can be seen in more than 350 peer-reviewed publications, many best-paper awards, more than 30 issued patents, and technologies licensed to seven companies. He was listed as the Most Influential Scholar in the field of Multimedia by Aminer in 2016. For his long-term pioneering contributions, he has been awarded the IEEE Signal Processing Society Technical Achievement Award, ACM Multimedia Special Interest Group Technical Achievement Award, Honorary Doctorate from the University of Amsterdam, the IEEE Kiyo Tomiyasu Award, and IBM Faculty Award. For his contributions to education, he received the Great Teacher Award from the Society of Columbia Graduates. He served as Chair of ACM SIGMM (2013–2017), Chair of Columbia Electrical Engineering Department (2007–2010), the Editor-in-Chief of the IEEE Signal Processing Magazine (2006–2008), and advisor for several international research institutions and companies. In his current capacity as Senior Executive Vice Dean at Columbia Engineering, he plays a key role in the School's strategic planning, special research initiatives, international collaboration, and faculty development. He is a Fellow of the American Association for the Advancement of Science (AAAS), a Fellow of the IEEE, and a Fellow of the ACM.